\documentclass{article}

\usepackage{microtype}
\usepackage{graphicx}
\usepackage{subfigure}
\usepackage{booktabs} 
\usepackage{enumitem}

\usepackage{hyperref}

\usepackage{soul}

\usepackage[accepted]{icml2024}

\usepackage{amsmath}
\usepackage{amssymb}
\usepackage{mathtools}
\usepackage{amsthm}
\usepackage[T1]{fontenc}
\usepackage[utf8]{inputenc}
\usepackage[capitalize,noabbrev]{cleveref}

\theoremstyle{plain}

\theoremstyle{definition}

\theoremstyle{remark}

\usepackage[textsize=tiny]{todonotes}

\icmltitlerunning{Interacting Diffusion Processes for Event Sequence Forecasting}

\usepackage{tikz}
\usepackage{multirow}
\DeclareUnicodeCharacter{2212}{-}

\definecolor{darkgreen}{rgb}{0.0, 0.26, 0.15}
\newcommand{\real}{\mathbb{R}}
\newcommand{\data}{\mathcal{D}}

\newcommand{\ex}{\mathbb{E}}

\newcommand{\bx}{\mathbf{x}}

\newcommand{\bz}{\mathbf{z}}

\newcommand{\be}{\mathbf{e}}
\newcommand{\bs}{\mathbf{s}}
\newcommand{\bt}{\mathbf{t}}

\newcommand{\loss}{\mathcal{L}}
\newcommand{\norm}{\mathcal{N}}

\newcommand{\tbf}[1]{\textbf{#1}}
\newcommand{\undl}[1]{\underline{#1}}
\newcommand{\tts}[1]{\text{#1}^*}
\newcommand{\ttt}[1]{\text{#1}}

\definecolor{britishracinggreen}{rgb}{0.0, 0.26, 0.15}

\usepackage{natbib}

\begin{document}



\twocolumn[
\icmltitle{Interacting Diffusion Processes for Event Sequence Forecasting}



\icmlsetsymbol{equal}{*}
\begin{icmlauthorlist}
\icmlauthor{Mai Zeng}{equal,yyy,ills,mila}
\icmlauthor{Florence Regol}{equal,yyy,ills,mila}
\icmlauthor{Mark Coates}{yyy,ills,mila}
\end{icmlauthorlist}

\icmlaffiliation{yyy}{Department of Electrical and Computer Engineering, McGill University, Montreal QC, Canada}
\icmlaffiliation{ills}{International Laboratory on Learning Systems (ILLS), Montreal, QC, Canada}
\icmlaffiliation{mila}{Mila Qu{\'e}bec AI Institute, Montreal, QC, Canada}

\icmlcorrespondingauthor{Mai Zeng}{mai.zeng@mail.mcgill.ca}
\icmlcorrespondingauthor{Florence Regol}{florence.robert-regol@mail.mcgill.ca}
\icmlcorrespondingauthor{Mark Coates}{mark.coates@mcgill.ca}

\icmlkeywords{Machine Learning, Sequence Modelling, Generative model}

\vskip 0.3in
]



\printAffiliationsAndNotice{\icmlEqualContribution} 


\begin{abstract}
Neural Temporal Point Processes (TPPs) have emerged as the primary framework for predicting sequences of events that occur at irregular time intervals, but their sequential nature can hamper performance for long-horizon forecasts.
To address this, we introduce a novel approach that incorporates a diffusion generative model. The model facilitates sequence-to-sequence prediction, allowing multi-step predictions based on historical event sequences. In contrast to previous approaches, our model directly learns the joint probability distribution of types and inter-arrival times for multiple events. The model is composed of two diffusion processes, one for the time intervals and one for the event types. These processes interact through their respective denoising functions, which can take as input intermediate representations from both processes, allowing the model to learn complex interactions. We demonstrate that our proposal outperforms state-of-the-art baselines for long-horizon forecasting of TPPs.

\end{abstract}


\section{Introduction}

\begin{figure}[t!]
\begin{tikzpicture}
   \node[anchor=south west,inner sep=0,yshift=-0.4cm] (image1) at (0,0) {\includegraphics[width=1\columnwidth]{
   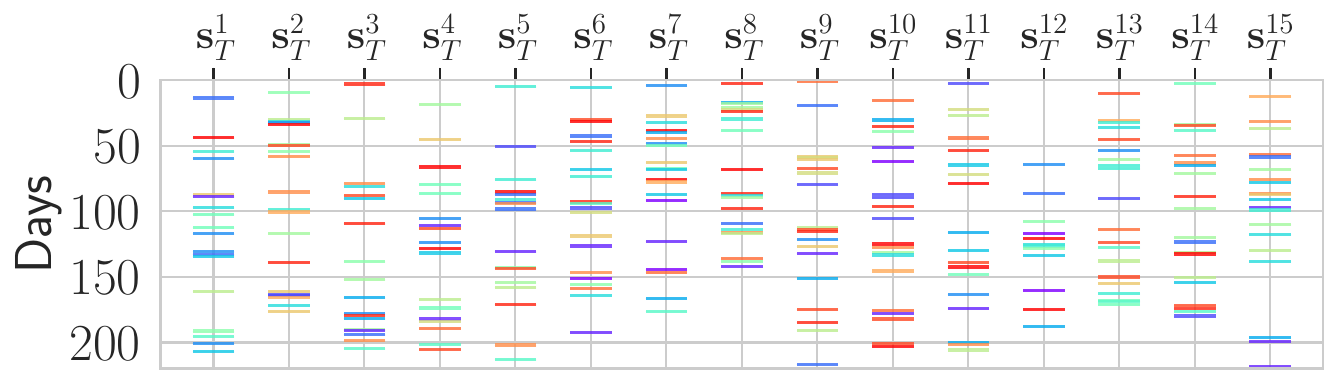}};
   
   \node[anchor=north west,inner sep=0,yshift=-0.4cm] (image2) at (image1.south west) {\includegraphics[width=1\columnwidth]{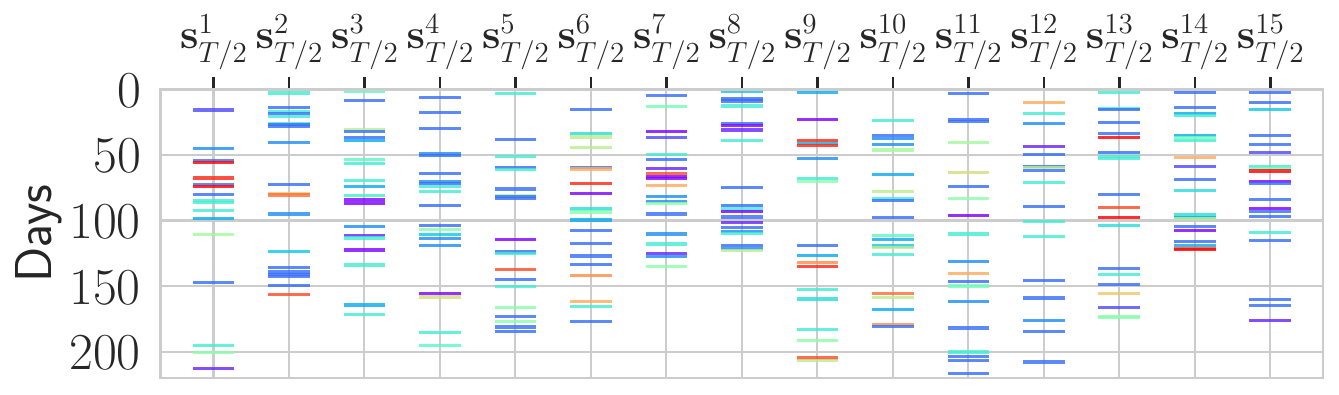}};

   \node[anchor=north west,inner sep=0,yshift=-0.4cm] (image3) at (image2.south west) {\includegraphics[width=1\columnwidth]{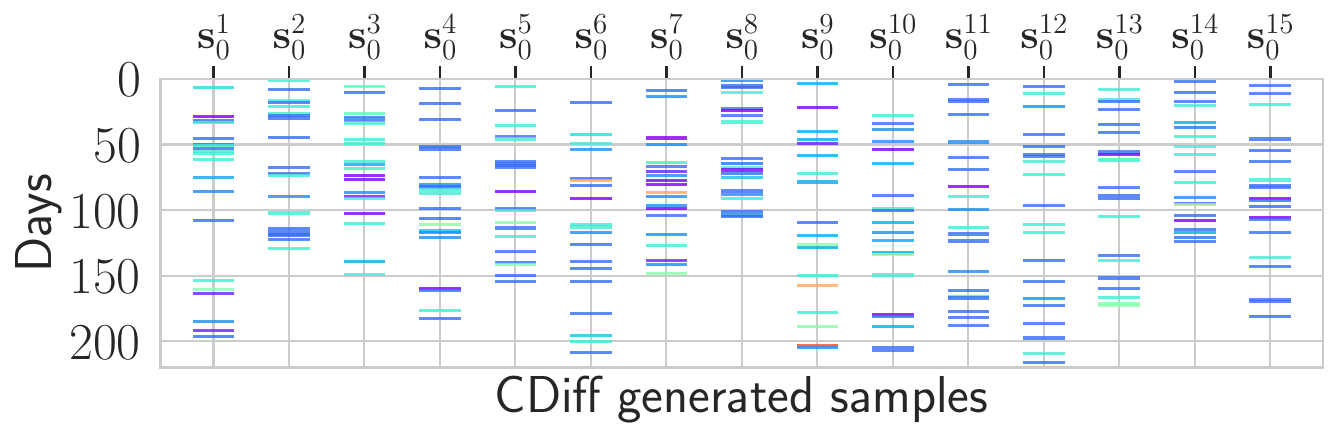}};

   \node[anchor=north west,inner sep=0,yshift=-0.1cm] (image4) at (image3.south west) {\includegraphics[width=1\columnwidth]{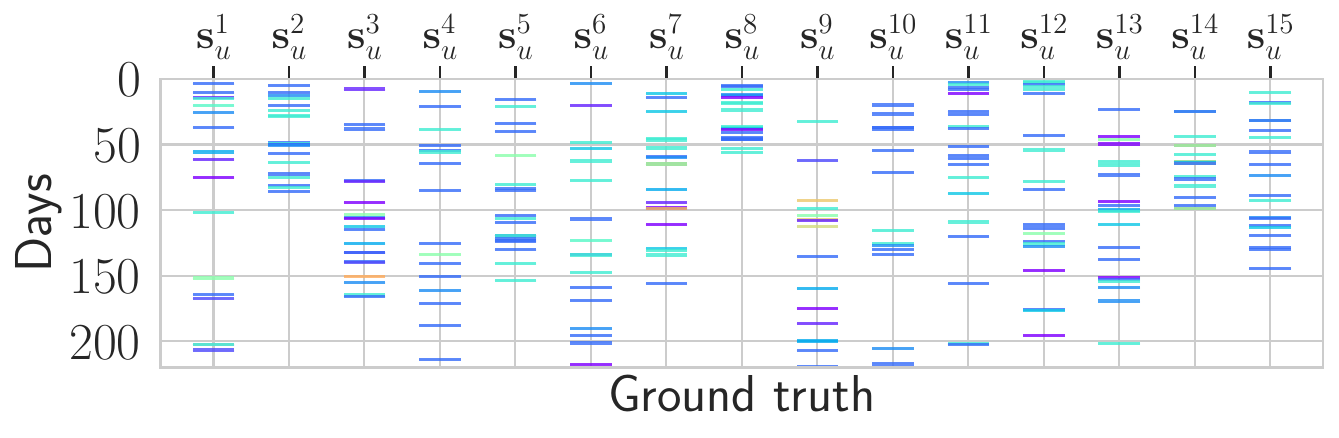}};

   \draw[-stealth, ultra thick]  (image1.south) to[out=-90,in=90] (image2.north);
   \draw[-stealth, ultra thick] (image2.south) to[out=-90,in=90]  (image3.north);
\end{tikzpicture}

   \caption
    {Visualization of the cross-diffusion generating process for 15 example Stackoverflow sequences. The colors indicates the different categories. We start by generating noisy sequences ($t=T$). Once we reach the end of the denoising process ($t=0$), we recover sequences similar to ground truth sequences.  } 
\label{fig:seq_visual_main_text}

\end{figure}

Predicting sequences of events has many practical applications, including forecasting purchase times and modeling transaction patterns or social media activity. The problem requires a dedicated model because it involves the complex task of jointly modeling two challenging data types: strictly positive continuous data for inter-arrival times and categorical data representing event types.

Early works employed intensity-based models such as the Hawkes process \cite{liniger2009multivariate}. This modelling choice has advantages, including interpretability --- it specifies the dynamics between events in the sequence explicitly. Subsequent efforts targeted integrating deep learning methods within the intensity framework~\cite{mei2017NeuralHawkesProcess,zuo2020TransformerHawkesProcess,yang2022AttNHP}.

Although they can fit complex distributions, intensity-based formulations have drawbacks~\cite{shchur2020intensityfree}. As generative modeling research has developed, TPP models have moved away from the intensity parameterization, with more flexible specifications allowing them to use the full potential of recent generative models~\citep{shchur2020intensityfree, Gupta2021,Lin2022}. 

Until recently, research has focused on next event forecasting. In~\citep{xue2022HYPROHybridlyNormalizeda,deshpande2021longhorizonforecastinga}, attention has turned to longer horizons, with the goal being forecasting multiple events. Recently proposed methods remain autoregressive, which can lead to a faster accumulation of error, as we illustrate in our experiments, but they are paired with additional modules that strive to mitigate this.

Our proposal goes a step further by directly generating a sequence of events. Consequently, our model can capture intricate interactions within the sequence of events between arrival times and event types.  {The crux of our proposed architecture is depicted in Fig. \ref{fig:seq_visual_main_text}. We introduce coupled denoising diffusion processes to learn the probability distribution of the
event sequences.} One is a categorical diffusion process; the other is real-valued. The interaction of the neural networks that model the reverse processes allows us to learn dependencies between event type and interarrival time. Fig.~\ref{fig:seq_visual_main_text} provides a visualization of the generation process\footnote{The code and implementation are available at our \href{https://github.com/networkslab/cdiff.git}{official repository}}.

Our approach significantly outperforms existing baselines for long-term forecasting, while also improving efficiency. Our experimental analysis provides insights into how the model achieves this: it can capture more complex correlation structures and is better at predicting distant events.

\section{Problem Statement}
Consider a sequence of events denoted by $ \bs^+ = \{ ({{x^+_i}}, e_i) \}_{1 \le i \le T}$, where ${{x^+_{i}}} \in (0,\infty) $ corresponds to the time interval between the events $e_i$ and $e_{i-1}$,  and the event $e_{i}$ belongs to one of $K$ categories: $e_{i} \in \mathcal{C}, | \mathcal{C}|=K $. The $+$-superscript is used to emphasize that the time-intervals are strictly positive. Given that we observe the start of a sequence (the context)  $ \bs^+_c = \{ ({{x^+_i}}, e_i) \}_{1 \le i \le I}$ (or ${{\bx^+_{c}}} = [{{x^+_{1}}},...,{{x^+_{I}}}] $ and $\be_{c} = [e_{1},...,e_{I}]$ 
 in vector form) with $I<T$, the goal is to forecast the remaining events. {The dataset consists of a set of sequences: $ \data = \{ \bs^{+,j} \}^M_{j=1}$} of potentially varying length.
\paragraph{Next $N$ events forecasting}\label{sec:problem_setting_1}
In this setting, the task is to predict the 
 following $N$ events in the sequence $\bs^+_u$: ${{\bx^+_u}} = [{{x^+_{I+1}}},...,{{x^+_{I+N}}}] $ and $\be_u = [e_{I+1},...,e_{I+N}]$. We also consider a slightly different setting: \textbf{interval forecasting}, where the task is to predict the events in a given time interval. We include the description of that setting with the metrics and methodology in Appendix~\ref{sec:interval_forecasting}.
 
\begin{figure}[h]
    \centering
    \includegraphics[scale=0.4]{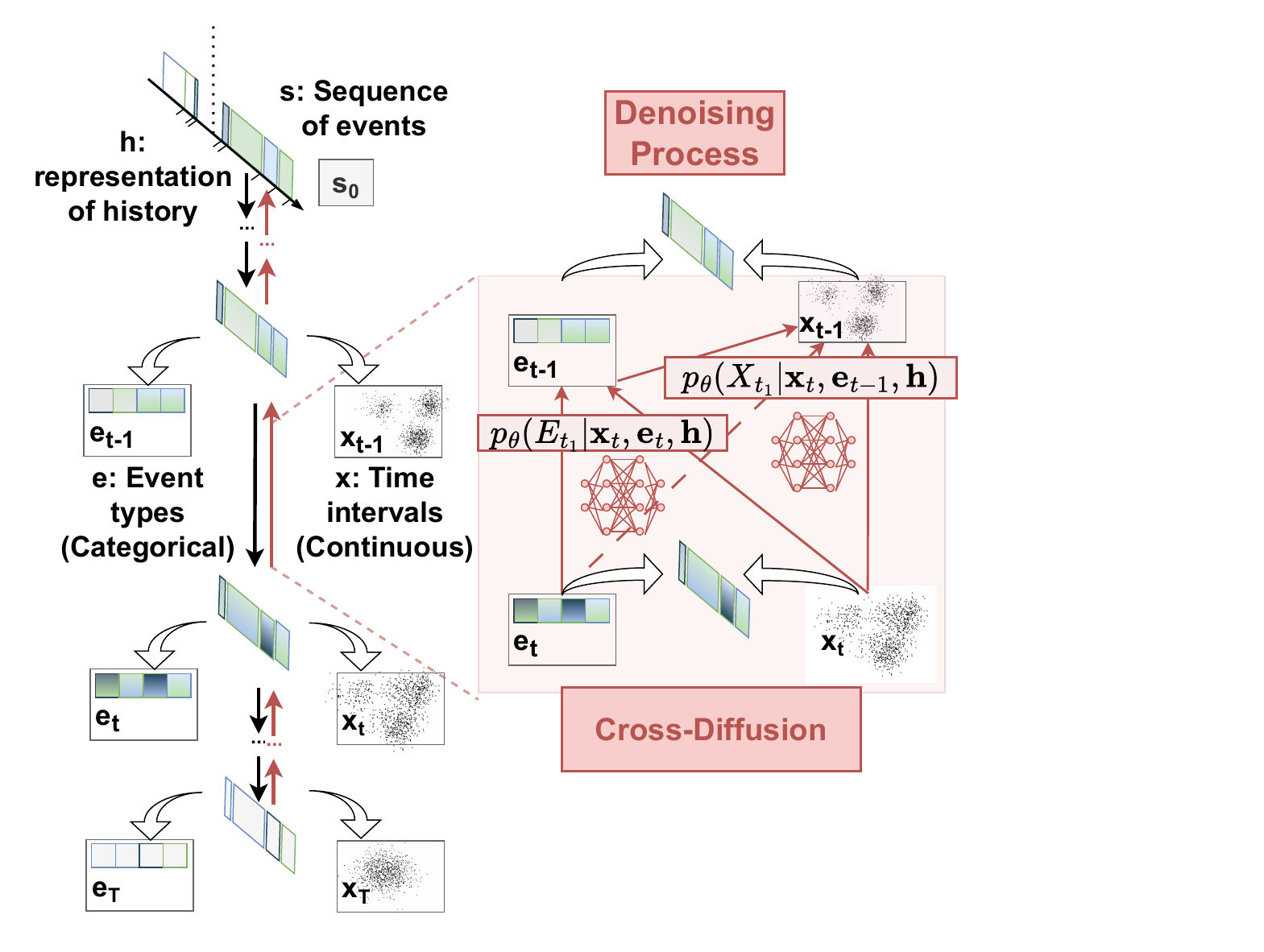}
    \caption{Architectural overview of our model CDiff. We employ two interacting denoising diffusion processes, one categorical and one real-valued, to model the high-dimensional event sequences. The neural networks modeling the reverse diffusion steps interact, allowing them to learn dependencies between event types and interarrival times. Generating an entire sequence at once avoids the error propagation that can plague autoregressive models.}
    \label{fig:overview}
\end{figure}


\section{Related Work}
We now briefly review and discuss relevant TPP modelling and forecasting literature. \citet{bosser2023} and \citet{Shchur2021} provide more comprehensive reviews. 

\paragraph{Hawkes-based methods.}
Early TPP forecasting approaches target single-event prediction (\textbf{Next $N{=}1$ event forecasting}) and adopt an intensity-based formulation~\citep{rasmussen2011tpplecture}. The multivariate Hawkes Process (MHP) \citep{liniger2009multivariate} is the basis for many models~\citep{du2016RecurrentMarkedTemporal, mei2017NeuralHawkesProcess, zuo2020TransformerHawkesProcess, yang2022AttNHP}. 
Some approaches retain the intensity function but deviate from the MHP, incorporating graph learning~\citep{Zhang2021}, non-parametric methods~\citep{Pan2021} or meta-learning~\citep{bae2023meta}. 
Other research addresses the efficiency~\citep{shchur2020triangular,nickel2020} and expressiveness~\citep{Omi2019}.

\paragraph{Non-Hawkes methods }
Striving to develop more effective models by departing from the intensity formulation, ~\citet{shchur2020intensityfree} use  a  log-normal  distribution paired with normalising flows. \citet{Lin2022} explore multiple conditional generative models for time forecasting including diffusion, variational inference, Generative Adversarial Networks (GANs), and normalizing flows. In all of these models, the types and interarrival times are modelled as conditionally independent given the history.  

These works limit themselves to modelling a single upcoming event ($N{=}1$). As a result, they do not exploit the models' impressive ability to represent complex high dimensional data. In addition, modelling type and interarrival time independently is undesirable given that different event types can often be associated with very different arrival patterns.
 
\paragraph{Long horizon forecasting}
\citet{xue2022HYPROHybridlyNormalizeda} and \citet{deshpande2021longhorizonforecastinga} consider long horizon forecasting. \citet{xue2022HYPROHybridlyNormalizeda} generate multiple candidate prediction sequences and introduce a selection module that aims to learn to select the best candidate. \citet{deshpande2021longhorizonforecastinga} introduce a hierarchical architecture and a ranking objective to improve prediction of the number of events in each interval.

Although these works explicitly target long horizon forecasting, their generation mechanisms remain sequential. The techniques try to mitigate the error propagation in sequential models, but fundamentally they still only learn a model for $p(\{e_{h+1},x^+_{h+1}\}| \{e_{i},x^+_{i}\}_{i\leq h} )$. As a result, the algorithms retain the core limitations of one-step ahead autoregressive forecasting. The approach of directly modeling a sequence of events has been explored in the non-marked setting~\cite{luedke2023add}. When there are no marks (which indicate different event types), the observed sequence of time intervals can be treated as (iid) samples from a conditional intensity distribution. Consequently,~\citet{luedke2023add} can directly parameterize the diffusion with a Poisson distribution in their Add-and-thin model. This differs from our approach -- we model the interaction between event types and time intervals by learning the joint distribution.
{For completeness, we include a comparison in the Appendix with a modified version of Add-and-thin augmented with a naive event type predictor module. }

\section{Methodology}

\paragraph{Model Overview}

Our proposal is to tackle the multi-event forecasting problem by directly modelling a complete sequence of $N$ events. We therefore frame our problem as learning the conditional distribution $P( S^+_u= \bs_u | \bs_c)$, where $\bs^+_u=(\be_u, \bx^+_u)$ is the sequence of event types and interval to forecast and $\bs^+_c=(\be_c, \bx^+_c)$ is the historical (context) sequence. We introduce our Cross-Diffusion (CDiff) model, which comprises two interacting diffusion processes.

In a nutshell, we diffuse simultaneously both the time intervals and the event types of the target sequence: {We first apply a Box-Cox transformation to the inter-arrival time values to transition from the strictly positive continuous domain ($X^+ \in (0, +\infty)$) to the more convenient unrestricted real space ($X \in (-\infty, +\infty)$). We use $S=(X,E)$ instead of $S^+=(X^+,E)$ to indicate the event sequence with $X$ in the unrestricted real space.} We gradually add Gaussian noise to the transformed time intervals and uniform categorical noise to the types ${S_0, S_1, \dots, S_T }$ until only noise remains in $S_T$. $S_0$ denotes the target sequence $S_u$. During training, we learn denoising distributions $p_{\theta}(S_{t-1} | S_t, \bs_c)$ that can undo each of the noise-adding steps. Our denoising functions are split in two, but interact with each other, which is why we call our model ``cross-diffusion.'' After training, we sample from $P( S_u | \bs_c)$ by sampling noise $S_T$, then gradually reversing the chain by sampling from $p_{\theta}(S_{t-1} | S_t, \bs_c)$ until we recover $S_0$. A high-level summary of our approach is illustrated in Fig.~\ref{fig:overview}. The specifics of the model and its training are provided in the subsequent sections.

\subsection{Model Details}

A TPP model can be divided into two components~\citep{Lin2022}: 1) the encoder of the variable length context $\bs_c$; and 2) the generative model of the future events. We focus on the latter and adopt the transformer-based context encoder proposed by \citet{xue2022HYPROHybridlyNormalizeda}  in order to generate a fixed-dimensional context representation denoted as $\boldsymbol{h} = f_{\theta}(\bs^+_c)$. 

{Again, we first apply a Box-Cox transformation to the inter-arrival time values to transition from the strictly positive continuous domain ($X^+ \in (0, +\infty)$) to the more convenient unrestricted real space ($X \in (-\infty, +\infty)$). This allows us to model the variables with Gaussian distributions in the diffusion process. Appendix \ref{sec:boxcox_transform} provides more detail.}

Although the target distribution consists of a combination of categorical and continuous variables, we can define a single diffusion process for it. To achieve this, we begin by defining a forward/noisy process that introduces $T$ new random variables, which are noisier versions of the sequence, represented by $S_0 = (X_0,E_0)$:
\begin{align}
    q(X_{1:T}, E_{1:T}| X_0, E_0) = \prod^T_{t=1} q(X_{t}, E_{t}| X_{t-1}, E_{t-1}).
\end{align}
For a diffusion model we strive to learn the \textit{inverse denoising} process by learning the intermediate distributions $p_{\theta}(S_{t-1}|S_{t},\bs_c)$. The log likelihood of the target distribution $ \log q(S_0|\bs_c)$ is obtained by marginalizing over the denoising process. Following the diffusion model setup of~\citep{ho2020ddpms}, this marginalization can be approximated as:
\begin{align}
    &\log q(S_0|\bs_c) \geq  \ex_{q(S_0|\bs_c)}\Big[\log p_{\theta}(S_0|S_1,\bs_c) \nonumber \\
    &\quad - KL(q(S_T|S_0,\bs_c) || q(S_T|\bs_c))  \nonumber \\ & - \sum^T_{t=2}KL\Big(q(S_{t-1}|S_{t},S_0,\bs_c) || p_{\theta}(S_{t-1}|S_{t},\bs_c)\Big) \Big] \label{eq:optim}.
\end{align}
Hence, we can summarize the generative diffusion model approach as follows: by minimizing the KL-divergences between the learned distributions  $p_{\theta}(S_{t-1}|S_{t},\bs_c)$ and the noisy distributions $q(S_{t-1}|S_t, S_0,\bs_c)$ at each $t$, we maximize the log likelihood of our target $ \log q(S_0|\bs_c)$.

\paragraph{Cross-diffusion for modeling event sequences}
As $X_u$ and $E_u$ are in different domains, we cannot apply a standard noise function to $q(X_{t}, E_{t}| X_{t-1}, E_{t-1})$. Instead, we factorize the noise-inducing distribution $ q(S_{t}| S_{t-1}) = q(X_{t}| X_{t-1})  q(E_{t}|  E_{t-1})$. It is important to stress that this independence is only imposed on the forward (noise-adding) process. We do not assume independence in $q(S_0|\bs_c)$ and our reverse diffusion process, described below, allows us to learn the dependencies.
Denoting by $Cat(;p)$ a categorical distribution with parameter $p$, and given an increasing variance schedule $\{\beta_1, \dots, \beta_T\}$, the forward process is:
\begin{align}
    q(S_{t}| S_{t-1}) =& q(X_{t}| X_{t-1})  q(E_{t}|  E_{t-1}), \\
    q(X_{t}| X_{t-1}) =& \norm(X_t; \sqrt{1-\beta_t}X_{t-1}, \beta_t \mathbf{I}),\\
    q(E_{t}|  E_{t-1}) = &Cat(E_t; (1-\beta_t)E_{t-1}+\beta_t/ K),\\
    q(X_T) =& \norm(X_T; 0,\mathbf{I}),\\
    q(E_{T}) =& Cat(E_T ; 1/K),
\end{align}

Next, we have to define the denoising process $p_{\theta}(S_{t-1}|S_t )$. We can express the joint distribution as:
{\small
\begin{align}
    p_{\theta}(S_{t-1}|S_t,\bs_c ) =& p_{\theta}(X_{t-1}|S_t ,E_{t-1},\bs_c)p_{\theta}(E_{t-1}|S_t,\bs_c ), \label{eq:facto}\\
     p_{\theta}(E_{t-1}|S_t, \bs_c  ) =& Cat(E_{t-1}| \pi_{\theta}(X_t, E_t, t, \bs_c )), \label{eq:cat_denoise}\\
   p_{\theta}(X_{t-1}|S_t, E_{t-1}, \bs_c  ) =&  \norm(X_{t-1}; \mu_{\theta}(X_t, E_{t-1}, t, \bs_c), \sigma_t).
\end{align}
}
Here we choose to fix  $\sigma_t = \beta_t$ and $\mu_{\theta}$ and $\pi_{\theta}$ are learnable.
With the presented approach, during denoising, we first sample event types, and then conditioned on the sampled event types, we sample inter-arrival times. We can also choose to do the reverse. A sensitivity study in Appendix~\ref{sec:model_e_t_order} shows that this choice has a negligible effect on performance.

This can be viewed as two denoising processes that interact through the learnable functions $\mu_\theta$ and $\pi_\theta$. One models the inter-arrival times (Gaussian) and one models the event types (Categorical). 
{{The denoising processes are conditioned on the historical event sequence $\bs_c$ and they are interacting with each other (through conditioning on $\be_{t-1}$ and $\bx_t$). Therefore, we modify the standard parametrization  of $\mu_{\theta}(\bx_t, t)$ and $\pi_{\theta}(\be_t, t)$ from~\citep{ho2020ddpms, hoogeboom2021ArgmaxFlowsMultinomial} to include these additional inputs. Rather than directly learning $\mu$ and $\pi$, we express them in terms of two other functions $\epsilon$ and $\phi$ to facilitate learning.

Introducing $\alpha_t \triangleq 1-\beta_t$ and $\bar{\alpha}_t \triangleq \prod_{i \leq t} \alpha_i$, the time denoising process is parameterized as:}}
\begin{align}
\small
    \label{eq:dt_parameterization}
    \mu_{\theta}(\bx_t, \be_{t-1}, t,\bs_c) =  \frac{\bx_t}{\sqrt{\alpha_t}}  - \frac{\beta_t \epsilon_{\theta}(\bx_t,\be_{t-1},t, \bs_c)}{\sqrt{\alpha_t}\sqrt{1-\bar{\alpha}_t}}.
    \end{align}
{{The denoising step of the event type is parameterized differently. Its learnable component is parameterized to directly predict, at step $t$, the targeted distribution of the data  $E_0$, modeled as $\hat{\be}_0$, from the event type, $e_t$, the (transformed) time interval, $x_t$, the denoising step $t$, and the context $s_c$:
\begin{align}
\hat{\be}_0 &= \phi_{\theta}(\be_{t}, \bx_t, t,\bs_c).
    \label{eq:type_parameterization}.
\end{align}
This prediction $\hat{\be}_0$ is then combined with the current $\be_{t}$ through a weighted sum, and subsequently normalized to obtain the parameters of the categorical distribution for $E_{t-1}$ (parameterized as $\pi_{\theta}(\bx_t,\be_t,t,\bs_c)$ in Eqn~\ref{eq:cat_denoise}):
\begin{align}
\small
\boldsymbol{\theta}(\be_{t}, \hat{\be}_0) =
[\alpha_t \be_t + \frac{1-\alpha_t}{K}] &\odot  [\bar{\alpha}_{t-1}\hat{\be}_0 + \frac{1-\bar{\alpha}_{t-1}}{K}], \nonumber\\
\tilde{\boldsymbol{\theta}} &\triangleq \boldsymbol{\theta}(\be_{t}, \hat{\be}_0), \label{eq:multinomial_bold_theta}\\
\pi_{\theta}(\bx_t,\be_t,t,\bs_c)&= \tilde{\boldsymbol{\theta}} / \sum^{\mathrm{K}}_{k=1} \tilde{\boldsymbol{\theta}}_k.
\end{align}}}
Here $\odot$ denotes the Hadamard product. 
{This concludes our description of the parameterization of the denoising process.} 
The learnable components of CDiff  are $\epsilon_{\theta}, \phi_{\theta}$ and $ f_{\theta}$. In our experiments, we use transformer-based networks that we describe in Section~\ref{sec:details}.

With a trained model $p_{\theta}(S^0 | \bs_c)$, given a context sequence $\bs_c$, we can generate samples of the next $N$ events, $\hat{\bs}^0 \sim p_{\theta}(S^0 | \bs_c)$. To form the final predicted forecasting sequence $\hat{\bs}_u$, we generate multiple samples, calculate the average time intervals, and set the event types to the majority types. With an abuse of notation, we denote this averaging of sequences as $\hat{\bs}_u \triangleq \frac{1}{A}\sum^A_{a=1} \hat{\bs}_a^0, \quad \hat{\bs}_a^0 \sim p_{\theta}(S^0 | \bs_c)$.

\subsection{Optimization}
The log-likelihood objective is provided in Equation~\eqref{eq:optim}. We can separate the objective for the joint $q(S_0)$ into standard optimization terms of either continuous or categorical diffusion using Equation~\eqref{eq:facto}.

Starting with the first log term, we separate it as:
\begin{align}
    & \ex_{q(S_0| s_c)}\Big[\log p_{\theta}(S_0|S_1,\bs_c) \Big] \nonumber \\   
  &\hspace{-0.5em}\approx    \sum^M_{j=1}  \log p_{\theta}(\bx^j_0|\bx^j_1 ,\hat{\be}^j_0,\bs^j_c) +  \log p_{\theta}(\be^j_0|\bx^j_1,\be^j_1,\bs^j_c ), 
\end{align}
with $\hat{\be}^j_0{\sim}p_{\theta}(E_0|\bx^j_1,\be^j_1,\bs^j_c )$, $\be^j_1\sim q(E_1| \be_0^j,\bs^j_c)$, and $\bx^j_1 \sim q(X_1 | \bx_0^j,\bs^j_c)$.
Next, we split the individual KL terms from~\eqref{eq:optim} similarly:
{\small
\begin{align}
 &\ex_{q(S_0| s_c)} \Big[ KL\Big(q(S_{t-1}|S_{t,0} ,\bs_c) || p_{\theta}(S_{t-1}|S_{t},\bs_c)\Big) \Big]  = 
   \nonumber\\   &\quad\ex_{q(S_0|\bs_c)}\Big[ KL\Big(q(X_{t-1}|X_{t,0},\bs_c) || p_{\theta}(X_{t-1}|S_t ,E_{t-1},\bs_c) \Big) \Big] \nonumber\\
  &\quad + \ex_{q(S_0|\bs_c)}\Big[ KL\Big(q(E_{t-1}|E_{t,0},\bs_c) || p_{\theta}(E_{t-1}|S_t,\bs_c )\Big) \Big] .
\end{align}}
 The target distribution of the event type can be expressed compactly by substituting the true $\be_0$ in Eqn~\eqref{eq:multinomial_bold_theta}:
\begin{align}
\bar{\boldsymbol{\theta}}_{\text{post}}(\be_t, \be_0) &\triangleq \boldsymbol{\theta}(\be_{t}, \be_0) / \sum^{\mathrm{K}}_{k=1} \boldsymbol{\theta}(\be_{t}, \be_0)_k,\\
    q(\be_{t-1}|\be_t, \be_0) &= Cat(\be_{t-1} |  \bar{\boldsymbol{\theta}}_{\text{post}}(\be_t, \be_0))\,.\label{eq:type_sampling}
\end{align}

We can therefore apply the typical optimization techniques of either continuous and categorical diffusion on each term:
\begin{align}
\ex_{q(S_0| s_c)}&\Big[ KL\Big(q(E_{t-1}|E_{t,0},\bs_c) || p_{\theta}(E_{t-1}|S_t,\bs_c )\Big) \Big] \nonumber \\
 &\hspace{-2em}\approx   - \sum^M_{j=1} \sum_k \bar{\boldsymbol{\theta}}_{\text{post}}(\be^j_t, \be^j_0)_k \cdot \log \frac{\bar{\boldsymbol{\theta}}_{\text{post}}(\be^j_t, \be^j_0)_k}{\pi_{\theta}(\bx^j_t,\be^j_t,t,\bs^j_c)_k} 
    \label{eq:type_loss}
\end{align}
with $\be^j_t \sim q(E_t| \be_0^j,\bs^j_c)$, $ \bx^j_t \sim q(X_t | \bx_0^j,\bs^j_c)$ for the event variables, and:
\begin{align}
&\ex_{q(S_0| s_c)} \Big[ KL\Big(q(X_{t-1}|X_{t,0}, s_c) || p_{\theta}(X_{t-1}|S_t ,E_{t-1},\bs_c) \Big) \Big]  \nonumber \\
&\approx  - \sum^M_{j=1} \Vert \epsilon - \epsilon_{\theta}(\sqrt{\bar{\alpha}_t} \bx^j_0 + \sqrt{1-\bar{\alpha}_t}\epsilon, t, \hat{\be}^j_{t-1}, \bs^j_c )\Vert^2    \label{eq:dt_loss}
\end{align}
with $\be^j_t \sim q(E_t| \be_0^j)$, $ \bx^j_t \sim q(X_t | \bx_0^j)$, $\hat{\be}^j_{t-1} \sim p_{\theta}(E_{t-1}|\bx^j_t,\be^j_t,\bs^j_c ) $ and $\epsilon \sim \norm(0,1)$ for the continuous interarrival time variables.

Our final objective is hence given by:
\begin{align}
  \loss &=  \sum^M_{j=1} \Big( \log p_{\theta}(\bx^j_0|\bx^j_1 ,\hat{\be}^j_0,\bs^j_c) \log p_{\theta}(\be^j_0|\bx^j_1,\be^j_1,\bs^j_c ) \nonumber\\
     &\hspace{1em}- \sum^T_{t=2} \Big(   \Vert \epsilon - \epsilon_{\theta}(\sqrt{\bar{\alpha}_t} \bx^j_0 + \sqrt{1-\bar{\alpha}_t}\epsilon, t, \hat{\be}^j_{t-1}, \bs^j_c )\Vert^2 \nonumber\\
     &\hspace{1em}+   \sum^K_{k=1} \bar{\boldsymbol{\theta}}_{\text{post}}(\be^j_t, \be^j_0)_k \cdot \log \frac{\bar{\boldsymbol{\theta}}_{\text{post}}(\be^j_t, \be^j_0)_k}{\pi_{\theta}(\bx^j_t,\be^j_t,t,\bs^j_c)_k}  \Big) \Big).
\end{align}
Finally, we adhere to the common optimization approach used in diffusion models and 
optimize only one diffusion timestep term per sample instead of the entire sum. The timestep is selected by uniformly sampling $t \sim U(0,T)$. 
We employ the algorithm from~\citep{song2020ddim} to accelerate the sampling. Appendix~\ref{sec:sampling} provides further details.


\section{Experiments}
In our experiments, we set $N=20$ (but include results for $N=5, 10$). For each sequence in the dataset $\data = \{ \bs^j \}^M_{j=1}$, we set the last $N$ events as $\bs_u$ and set all earlier events as the context $\bs_c$.
Means and standard deviations are computed over 10 trials. We train for a  maximum of 500 epochs and report the best trained model based on the validation set. Hyperparameter selection uses the Tree-Structured Parzen Estimator  hyperparameter search algorithm  from~\citet{bergstra2011algorithmshpo}. To avoid numerical error when applying the Box-Cox transformation to the $x^+$ values, we first add 1e-7 to all time values and then scale by $100$. We transform back to $x^+$ after we estimate $x$ using the inverse Box-Cox transformation, with the same parameter obtained from the train set, and downscale by $100$. The detailed model description is in Appendix~\ref{appendix:detailed model description}, along with sensitivity studies for some of the hyperparameters.

\subsection{Datasets}
We use six real-world datasets. Taobao~\citep{alibaba2018tbdataset} tracks user clicks made on a website; Taxi~\citep{whong2014taxidataset} contains trips to neighborhoods by taxi drivers; StackOverflow~\citep{leskovec2014sodataset} tracks the history of posts on stackoverflow; Retweet~\citep{zhou2013retweetdata} tracks user interactions on social media; MOOC \cite{kumar2019moocdataset} tracks user interactions within an online course system; and Amazon \cite{ni2019amazondataset} tracks the sequence of product categories reviewed by a group of users. We focus on datasets containing sequences with multiple events as our goal is multi-event prediction. Our synthetic dataset is generated from a Hawkes model. We follow \citet{xue2022HYPROHybridlyNormalizeda} for the train/val/test splits, which we report in Appendix~\ref{sec:dataset}, together with additional dataset details.

\subsection{Baselines}
\label{sec:baselines_main}
We compare our CDiff model with one naive and $6$ state-of-the-art baselines for event sequence modeling. When available, we use reported hyperparameters, and otherwise we employ a tuning procedure (see Appendix~\ref{sec:hyper-parameters}).
\begin{itemize}[leftmargin=*]

\setlength\itemsep{0.1em}
    \item \textbf{Homogeneous Poisson Process (naive)} is a constant intensity function. For the type prediction, we compute the marginal categorical distribution over the training set.
    
    \item \textbf{Neural Hawkes Process (NHP)}~\cite{mei2017NeuralHawkesProcess} is 
 a Hawkes-based model that uses a continuous LSTM. 
 
    \item \textbf{Attentive Neural Hawkes Process (AttNHP)}~\cite{yang2022AttNHP} is a Hawkes-based model that integrates attention. It is the SOTA for single event forecasting.

    \item \textbf{Log-Normal Mixture Model (LNM)} \cite{shchur2020intensityfree} is an intensity-free temporal point process with the feature of fast sampling.

    \item \textbf{Temporal Conditional Diffusion Denoising Model (TCDDM)} \cite{Lin2022} is a diffusion based generative model that relies on the assumption of conditional independence between inter-arrival time and event type.
    
    \item \textbf{Dual-TPP}~\cite{deshpande2021longhorizonforecastinga}: Dual-TPP  targets long horizon forecasting by jointly learning a distribution of the count of events in segmented time intervals.
    
    \item \textbf{HYPRO} \cite{xue2022HYPROHybridlyNormalizeda}  is the SOTA for multi-event/long horizon forecasting. It uses AttNHP as a base model, but includes a sequence selection module. 
\end{itemize}

\subsection{Evaluation Metrics}
Assessing long-horizon performance is challenging as we must compare mixed-type vectors. There is no existing proper scoring rule. Therefore, we report multiple metrics. 

\textbf{Optimal Transport Distance ($\textbf{OTD}$):} We use the OTD to compare event sequences, following~\citet{mei2019ImputingMissingEvents}. $L(\hat{\bs}_u , \bs_u)$ is the minimum cost of editing a predicted event sequence $\hat{\bs}_u$ into the ground truth $\bs_u$. To accomplish this edit, we must identify the best {\em alignment} -- a one-to-one partial matching $\mathbf{a}$ -- of the events in the two sequences. We use the algorithm from~\citep{mei2019ImputingMissingEvents} to find this alignment, and report the average $\textbf{OTD}$ values when using various deletion/insertion cost constants $C = \{0.05,0.5,1,1.5, 2,3,4\}$. Appendix \ref{sec:otd_and_more_otd_results} presents more details about this metric.

$\textbf{RMSE}_e$
assesses how well the event type distribution in the predicted sequence matches ground truth. For each type $k$, we count the number of type-$k$ events in $\bx^+_u$, denoted $C_k$, as well as that in $\hat{\bx}^+_u$, denoted $\hat{C}_k$.  We report the root mean square error $\textbf{RMSE}_e=  \sqrt{\frac{1}{M}\sum_{j=1}^{M} \frac{1}{K} \sum^K_{k=1} (C^j_k - \hat{C}^j_k)^2}$. We also report time-series forecasting metrics: $\textbf{RMSE}_{x^+}$, $\textbf{MAPE}$, and $\textbf{sMAPE}$ (a normalized $MAPE$).  Appendix~\ref{sec:otd_and_more_otd_results} provides metric details.

\begin{table*}[h]
    \caption{$\textbf{OTD}$, $\textbf{RMSE}_{e}$, $\textbf{RMSE}_{x^+}$ and \textbf{sMAPE} of real-world datasets reported in mean $\pm$ s.d. Best are in bold, the next best is underlined. *indicates stat. significance w.r.t to the best method.}
    \vskip 0.1in
    \centering\resizebox{\linewidth}{!}{\begin{tabular}{lcccc|cccc} \toprule
    &\multicolumn{4}{c}{\textbf{Taxi}}&\multicolumn{4}{c}{\textbf{Taobao}}\\ 
        & $\textbf{OTD}$ & $\textbf{RMSE}_{e}$ &$\textbf{RMSE}_{x^+}$ & \textbf{sMAPE} & $\textbf{OTD}$ & $\textbf{RMSE}_{e}$&$\textbf{RMSE}_{x^+}$ & \textbf{sMAPE}\\ \midrule
    \textbf{HYPRO}      & \underline{21.653 $\pm$ 0.163} & $\text{\underline{1.231 $\pm$ 0.015}}^*$ & $\text{\underline{0.372 $\pm$ 0.004}}^*$ & $\text{\underline{93.803 $\pm$ 0.454}}^*$ & \textbf{44.336 $\pm$ 0.127}  &$\text{\underline{2.710 $\pm$ 0.021}}^*$ & $\text{\underline{{0.594 $\pm$ 0.030}}}^*$ & $\text{{134.922 $\pm$ 0.473}}^*$ \\
     \textbf{Dual-TPP}      & ${24.483 \pm 0.383}^*$ & $\text{1.353 $\pm$ 0.037}^*$ & $\text{0.402 $\pm$ 0.006}^*$ & $\text{95.211 $\pm$ 0.187}^*$ & $\text{47.324 $\pm$ 0.541}^*$ & $\text{3.237 $\pm$ 0.049}^*$ & $\text{0.871 $\pm$ 0.005}^*$ & $\text{141.687 $\pm$ 0.431}^*$\\ 
      \textbf{Attnhp}     &  ${24.762 \pm 0.217}^*$  &$\text{1.276  $\pm $ 0.015}^*$ & ${0.430 \pm 0.003}^*$& $\text{97.388 $\pm$  0.381}^*$ &  $\text{45.555 $\pm$ 0.345}^*$  & $\text{2.737 $\pm$ 0.021}$ & $\text{0.708 $\pm$ 0.010}^*$ & $\text{\underline{134.582 $\pm$ 0.920}}^*$ \\ 
       \textbf{NHP}      & $\text{25.114 $\pm$ 0.268}^*$ &$ {1.297 \pm 0.019}^*$ & ${0.399 \pm 0.040}^*$ & $\text{96.459 $\pm$ 0.521}^*$ &    $\text{48.131 $\pm$ 0.297}^*$   &$\text{3.355  $\pm$ 0.030}^*$ & $\text{0.837 $\pm$ 0.009}^*$ & $\text{137.644 $\pm$ 0.764}^*$\\  
      \textbf{LNM}     &  $\tts{24.053 $\pm$ 0.609}$  &$\tts{1.364 $\pm$ 0.032}$ & $\tts{0.384 $\pm$ 0.005}$& $\tts{95.719 $\pm$ 0.779}$ &  $\tts{45.757 $\pm$ 0.287}$  & $\tts{3.193 $\pm$ 0.043}$ & $\tts{0.575 $\pm$ 0.012}$ & $\ttt{{ 127.436 $\pm$ 0.606}}$ \\ 
       \textbf{TCDDM}      & $\ttt{22.148 $\pm$ 0.529}$ & $\tts{  1.309 $\pm$ 0.030 }$ & $\ttt{  0.382 $\pm$ 0.019 }$  & $\ttt{ 90.596 $\pm$ 0.574  }$  &    $\tts{45.563 $\pm$ 0.889 }$    & $\ttt{ 2.850 $\pm$ 0.058 }$  & $\ttt{ 0.569 $\pm$ 0.015 }$  & $\ttt{  126.512 $\pm$ 0.491 }$ \\      
       \midrule
         \textbf{CDiff}    & \textbf{21.013  $\pm$  0.158}   & \textbf{{1.131 $\pm$ 0.017}} & \textbf{{0.351 $ \pm$ 0.004}}& \textbf{87.993 $\pm$ 0.178} &   \underline{44.621 $\pm$ 0.139}  &\textbf{{2.653 $\pm$ 0.022}} & \textbf{{0.551 $\pm$ 0.002}} & \textbf{125.685 $\pm$ 0.151} \\ \bottomrule
         &\multicolumn{4}{c}{\textbf{StackOverflow}}&\multicolumn{4}{c}{\textbf{Retweet}} \\
         & $\textbf{OTD}$ & $\textbf{RMSE}_{e}$&$\textbf{RMSE}_{x^+}$ & \textbf{sMAPE} & $\textbf{OTD}$ & $\textbf{RMSE}_{e}$&  $\textbf{RMSE}_{x^+}$ & \textbf{sMAPE}\\ \midrule
         \textbf{HYPRO} & $\text{{42.359$\pm$0.170}}$ & \undl{1.140 $\pm$ 0.014} & $\text{1.554 $\pm$ 0.010}^*$ &  $\text{ 110.988 $\pm$  0.559 }^*$ & $\text{61.031$\pm$0.092}^*$ &$\text{2.623 $\pm$ 0.036}^*$ & $\text{30.100 $\pm$ 0.413}^*$  & \undl{106.110$\pm$ 1.505}\\
         \textbf{Dual-TPP} & $\text{\underline{41.752$\pm$0.200}}$ &\textbf{1.134 $\pm$ 0.019} & $\text{1.514 $\pm$ 0.017}^*$ &   $\text{ 117.582  $\pm$ 0.420 }^*$ &    $\text{ 61.095$\pm$0.101 }^*$  &$\text{2.679 $\pm$ 0.026}^*$ & 28.914 $\pm$ 0.300  & 106.900$\pm$ 1.293\\
         \textbf{AttNHP} & $\text{42.591 $\pm$ 0.408}^*$  &1.142 $\pm$ 0.011 & $\text{{1.340 $\pm$ 0.006}}$ & {108.542 $\pm$  0.531}       &     \undl{ 60.634 $\pm$ 0.097 }         & {2.561 $\pm$ 0.054} & $\text{28.812 $\pm$ 0.272}^*$ & $\text{107.234$\pm$ 1.293}^*$ \\
         \textbf{NHP} & $\text{43.791  $\pm$ 0.147}^*$ &$\text{1.244  $\pm$ 0.030}^*$ & $\text{1.487 $\pm$ 0.004}^*$ &      $\text{ 116.952 $\pm$ 0.404}^*$       &      $\tts{60.953 $\pm$ 0.079} $          &$\text{2.651  $\pm$ 0.045}^*$ & \underline{27.130 $\pm$ 0.224 } &$\text{ 107.075 $\pm$ 1.398}^*$\\ 
          \textbf{LNM} & $\tts{46.280 $\pm$ 0.892}$   & $\tts{  1.447 $\pm$ 0.057 }$  & $\tts{ 1.669 $\pm$ 0.005 }$  & $\tts{  115.122 $\pm$ 0.627 }$        &    $\tts{61.715 $\pm$ 0.152}$         & $\tts{2.776 $\pm$ 0.043 }$  & $\ttt{27.582 $\pm$ 0.191}$  & $\tts{106.711 $\pm$ 1.615 }$  \\
         \textbf{TCDDM} & $\ttt{42.128 $\pm$ 0.591}$  & $\tts{ 1.467 $\pm$ 0.014}$  & \undl{ 1.315 $\pm$ 0.004}  & \undl{107.659 ± 0.934}     &   \textbf{60.501 ± 0.087}      & \undl{2.387 ± 0.050}  & $\ttt{27.303 ± 0.152 }$  &$\textbf{106.048 ± 0.610}$ \\ 
         \midrule
         \textbf{CDiff} & \textbf{41.245 $\pm$ 1.400 } & {{1.141 $\pm$ 0.007}} & \textbf{{1.199 $\pm$ 0.006}}&      \textbf{106.175$\pm$ 0.340}   &  {60.661 ± 0.101}   & \textbf{{2.293 $\pm$ 0.034}} & \textbf{{27.101 $\pm$ 0.113} }  & {106.184 $\pm$ 1.121}\\\bottomrule

          &\multicolumn{4}{c}{\textbf{MOOC}}&\multicolumn{4}{c}{\textbf{Amazon}} \\
         & $\textbf{OTD}$ & $\textbf{RMSE}_{e}$&$\textbf{RMSE}_{x^+}$ & \textbf{sMAPE} & $\textbf{OTD}$ & $\textbf{RMSE}_{e}$&  $\textbf{RMSE}_{x^+}$ & \textbf{sMAPE}\\ \midrule
         \textbf{HYPRO} & $\ttt{48.621 $\pm$ 0.352}$ & $\ttt{\undl{1.169 $\pm$ 0.094}}$ & $\textbf{0.410 $\pm$ 0.005}$ & $\textbf{143.045 $\pm$ 7.992}$ & $\tts{\undl{38.613 $\pm$ 0.536}}$ & $\textbf{2.007 $\pm$ 0.054}$ & $\tts{0.477 $\pm$ 0.010}$ & \undl{82.506 $\pm$ 0.84}  \\
         \textbf{Dual-TPP} & $\ttt{50.184 $\pm$ 1.127}$ & $\tts{1.312 $\pm$ 0.019}$ & $\tts{0.435 $\pm$ 0.006}$ & $\tts{147.003 $\pm$ 2.908}$ & $\tts{42.646 $\pm$ 0.752}$ & $\ttt{2.562 $\pm$ 0.202}$ & $\tts{0.482 $\pm$ 0.012}$ & $\ttt{86.453 $\pm$ 2.044}$  \\
         \textbf{AttNHP} & $\tts{49.121 $\pm$ 0.720}$ & $\ttt{1.297 $\pm$ 0.049}$ & $\ttt{0.420 $\pm$ 0.009}$ & $\ttt{147.756 $\pm$ 4.812}$  & $\ttt{39.480 $\pm$ 0.326}$ & $\tts{2.166 $\pm$ 0.026}$ & {\undl{0.476 $\pm$ 0.033}} & $\tts{84.323 $\pm$ 1.815}$ \\
         \textbf{NHP} &$\tts{51.277 $\pm$ 1.768}$ & $\tts{1.458 $\pm$ 0.063}$ & $\tts{0.442 $\pm$ 0.007}$ & $\tts{148.913 $\pm$ 11.628}$  & $\tts{42.571 $\pm$ 0.293}$ & $\ttt{2.561 $\pm$ 0.060}$ & $\tts{0.519 $\pm$ 0.023}$ & $\tts{92.053 $\pm$ 1.553}$ \\ 
         \textbf{LNM} & $\tts{52.890 $\pm$ 1.151}$ & $\tts{1.428 $\pm$ 0.061}$ & $\tts{0.454 $\pm$ 0.008}$ & $\tts{149.987 $\pm$ 16.581}$   & $\tts{43.820 $\pm$ 0.232}$ & $\tts{3.050 $\pm$ 0.286}$ & $\tts{0.481 $\pm$ 0.145}$ & $\tts{90.910 $\pm$ 1.611}$  \\
         \textbf{TCDDM} & $\tts{50.739 $\pm$ 0.765}$ & $\tts{1.407 $\pm$ 0.112}$ & $\ttt{0.429 $\pm$ 0.015}$ & \undl{145.745 $\pm$ 11.835} & $\tts{42.245 $\pm$ 0.174}$ & $\tts{2.998 $\pm$ 0.115}$ & $\ttt{\undl{0.476 $\pm$ 0.111}}$ & $\ttt{83.826 $\pm$ 1.508}$  \\ 
         \midrule
         \textbf{CDiff} & $\textbf{47.214 $\pm$ 0.628}$   & $\textbf{{1.095 $\pm$ 0.048}}$ & \undl{{0.411 $\pm$ 0.009}} & $\tts{146.361 $\pm$ 14.837}$ & \textbf{37.728  $\pm$  0.199}   & \undl{{2.091 $\pm$ 0.163}} & \textbf{{0.464 $\pm$ 0.086}} & \textbf{81.987 $\pm$ 1.905} \\
         \bottomrule
    \end{tabular}}

    \label{tab:condensed_table}
\end{table*}

\subsection{Implementation details}\label{sec:details}
Since all methods are generative, we can generate several samples to form the final predictions $\hat{\be}_u$ and $\hat{\bx}^+_u$. We generate 5 samples from $ P(S^+_u| S^+_c = \bs^+_c)$ and average the time vectors to form $\hat{\bx}^+_u$, and use majority voting over the event vectors to form  $\hat{\be}_u$. For the history encoder $f_{\theta}$, we adopt the architecture in AttNHP \citep{yang2022AttNHP}, which is a continuous-time Transformer module. For the two diffusion denoising functions $\epsilon_{\theta}(\cdot), \phi_{\theta}(\cdot)$, we use the PyTorch built-in transformer block~\citep{paszke2019pytorch}. We use the following positional encoding from~\citep{zuo2020TransformerHawkesProcess} for the sequence index $i$ in the $f_{\theta}(\cdot)$ transformer:
\begin{align}
[\mathbf{m}(y_j, D)]_i=\left\{
\begin{aligned}
\cos(y_j/10000^{\frac{i-1}{D}}) \quad \text{if $i$ is odd\,,}\\
\sin(y_j/10000^{\frac{i}{D}}) \quad \text{if $i$ is even.}
\end{aligned}
\right.
\label{eq:temporal_enc}
\end{align}

Appendix \ref{sec:positional_enc_diff} provides more details about the positional encoding implementation and its use for the diffusion timestep $t$. For the diffusion process, we use a cosine ${\beta}$ schedule, as proposed by \citet{nichol2021improvedddpm}. Appendix~\ref{sec:hyper-parameters} provides provides more detail concerning hyperparameters. 

\section{Results}
\label{sec:results_main_text}

\begin{figure*}[h]
    \centering
    \includegraphics[width=0.42\linewidth]{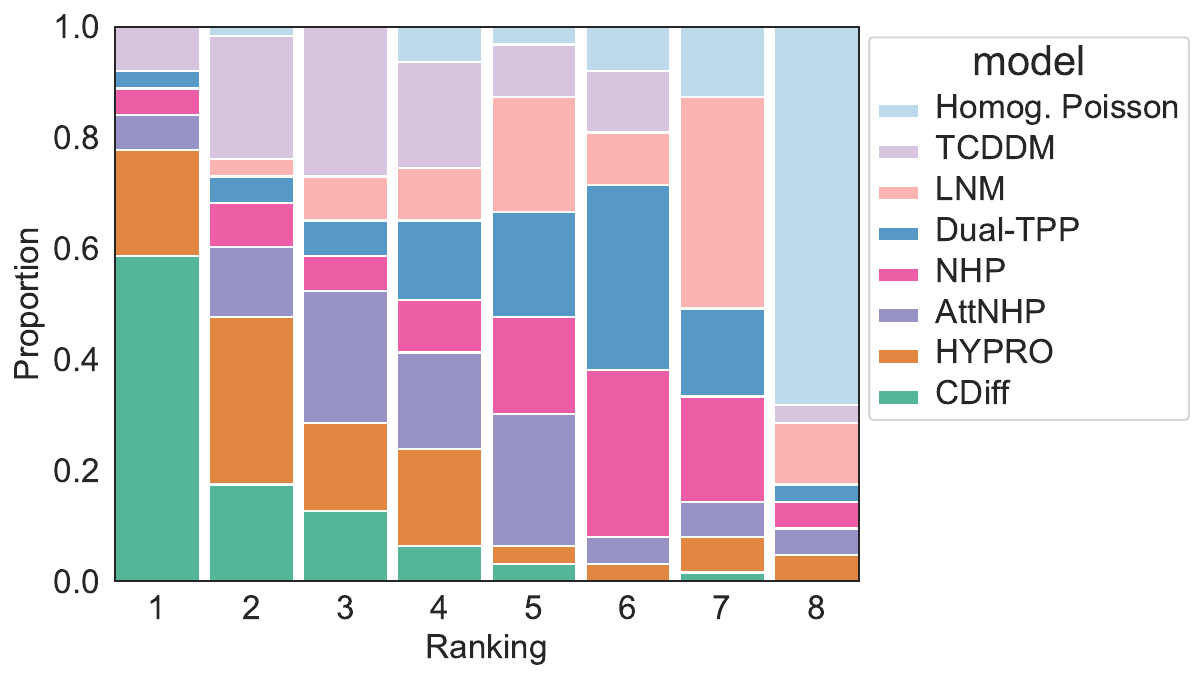}
    \includegraphics[width=0.28\textwidth   ,trim={50 0 162 0},clip]{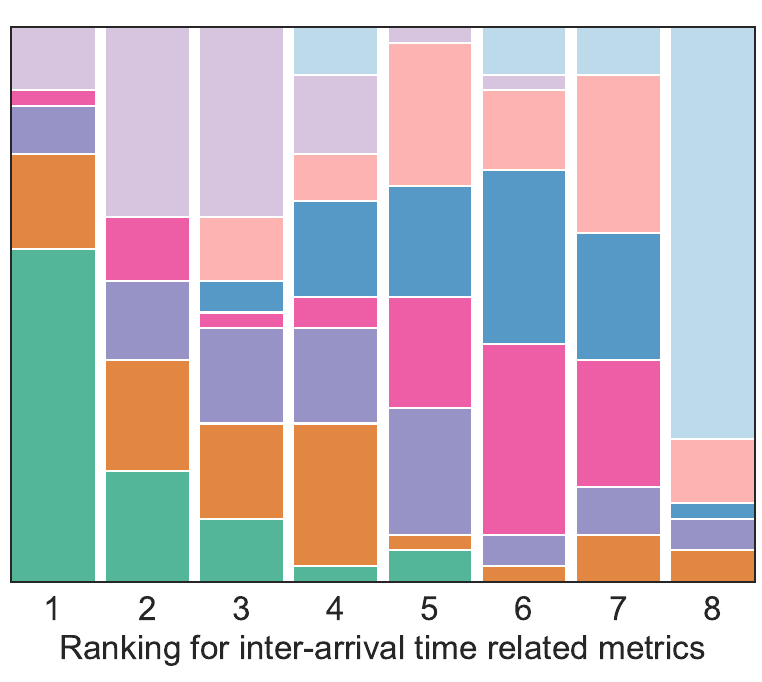}
    \includegraphics[width=0.28\textwidth  ,trim={50 0 162 0},clip]{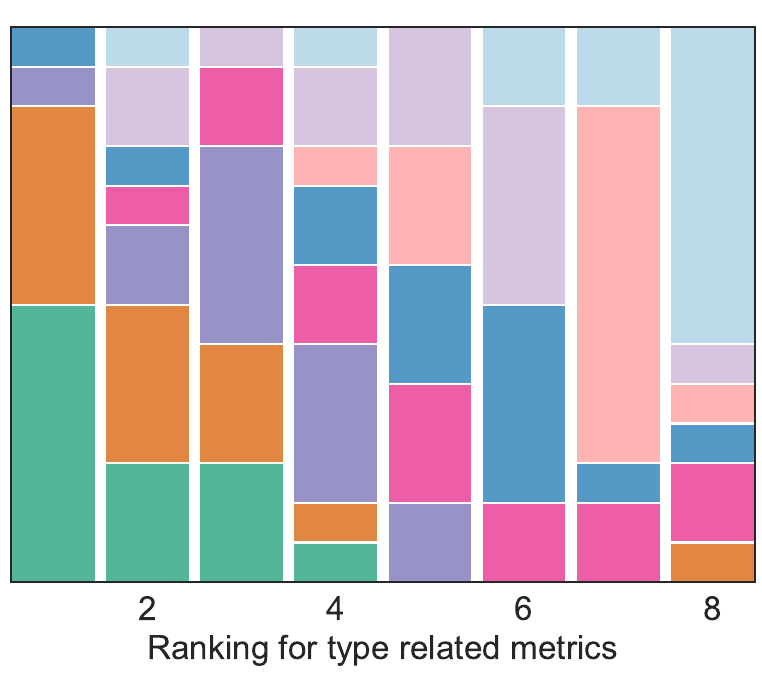}
    \caption{\textbf{Left)} Stacked column chart of ranks of the algorithms across the 5 datasets for all the metrics. We collect the rank for each metric ($9$ metrics in total, as we include additional metrics from the interval forecasting experiment described in the Appendix~\ref{sec:interval_forecasting}). The x-axis is the rank, and the y-axis is the proportion adding up to 1.\textbf{ Middle)} Stacked column chart of ranks only for time-related metrics ($\textbf{RMSE}_{x^+}$, $\textbf{MAPE}$, $\textbf{sMAPE}$, $\textbf{RMSE}_{|\bs^+|}$,  $\textbf{MAE}_{|\bs^+|}$). \textbf{ Right)}  Stacked column chart of ranks only for type-related metric ($\textbf{RMSE}_{e}$).}
    \label{fig:ranking_time_type}
\end{figure*}

Table~\ref{tab:condensed_table} presents results of a subset of experiments for four selected metrics on real-world datasets. Complete results are in Appendix \ref{sec:complete_result}. We test for significance using a paired Wilcoxon signed-rank test at the $5\%$ significance level. 

In alignment with previous findings, AttNHP consistently outperforms NHP, reaffirming its position as the SOTA single event forecasting method. HYPRO ranks as the second-best baseline since it leverages AttNHP as its base model and is designed for multi-event forecasting.  
Attention-based TCDDM and AttNHP show comparable results, while RNN models like NHP and others fall behind. The basic Homogeneous Poisson model ranks lowest. Our CDiff method consistently surpasses all baselines, often with a statistically significant margin, a trend that holds across various experiments, datasets, and metrics, as shown in Figure~\ref{fig:ranking_time_type}.

Figure~\ref{fig:ranking_time_type}(left) demonstrates CDiff's consistent top ranking. The middle and right panels show its outperformance for event type and time interval metrics. RNN-based models like LNM, NHP, and Dual-TPP fall short in long-term forecasting compared to attention-based models. TCDDM and LNM, while better at timing predictions, struggle with event type forecasting. This limitation is likely due to their assumption of conditional independence for event type prediction, which may impair their ability to capture complex relationships between event types and inter-arrival times.

\subsection{CDiff can model complex inter-arrival times}

\begin{figure}[h]
    \centering
    \includegraphics[width=\linewidth]{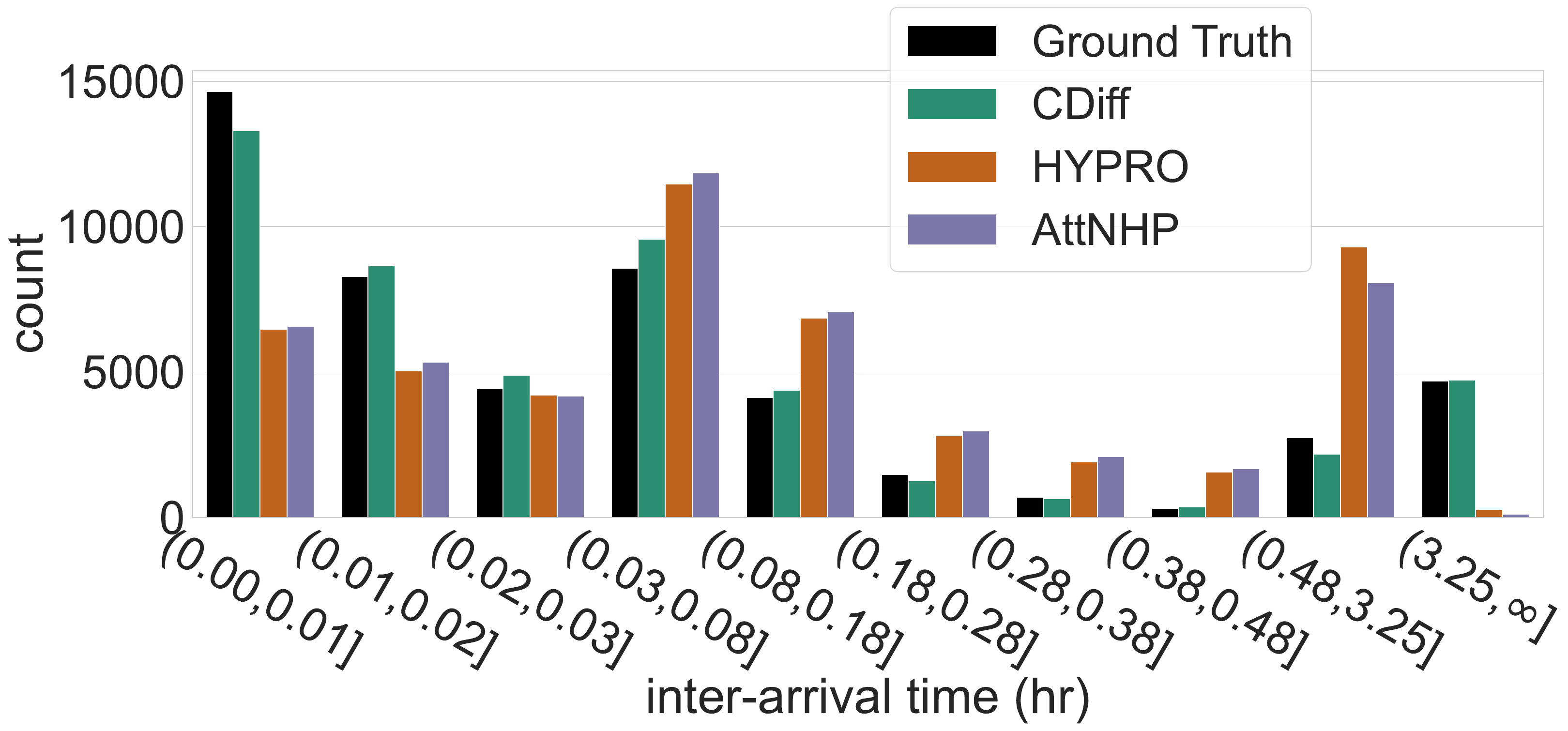}
    \caption{Histogram of true and predicted inter-arrival times for the Taobao dataset. Note that the bin widths gradually increase to make visual comparison easier.}
    \label{fig:hist_plot_taobao_dt_dist}
\end{figure}

We first examine the learned marginal distribution for time intervals. We use the Taobao dataset for our analysis because it is a relatively challenging dataset, with 17 event types and a marginal distribution of inter-arrival times that appears to be multi-modal. From the histograms of inter-arrival time prediction in Fig.~\ref{fig:hist_plot_taobao_dt_dist}, we see that CDiff is better at capturing the ground truth distribution. CDiff is effective at generating both longer intervals, falling within the range $(3h25,  \infty]$, and shorter intervals, within the range $(0,0.01h]$.

\begin{figure}[h!]
    \centering
    \includegraphics[width=0.4\textwidth]{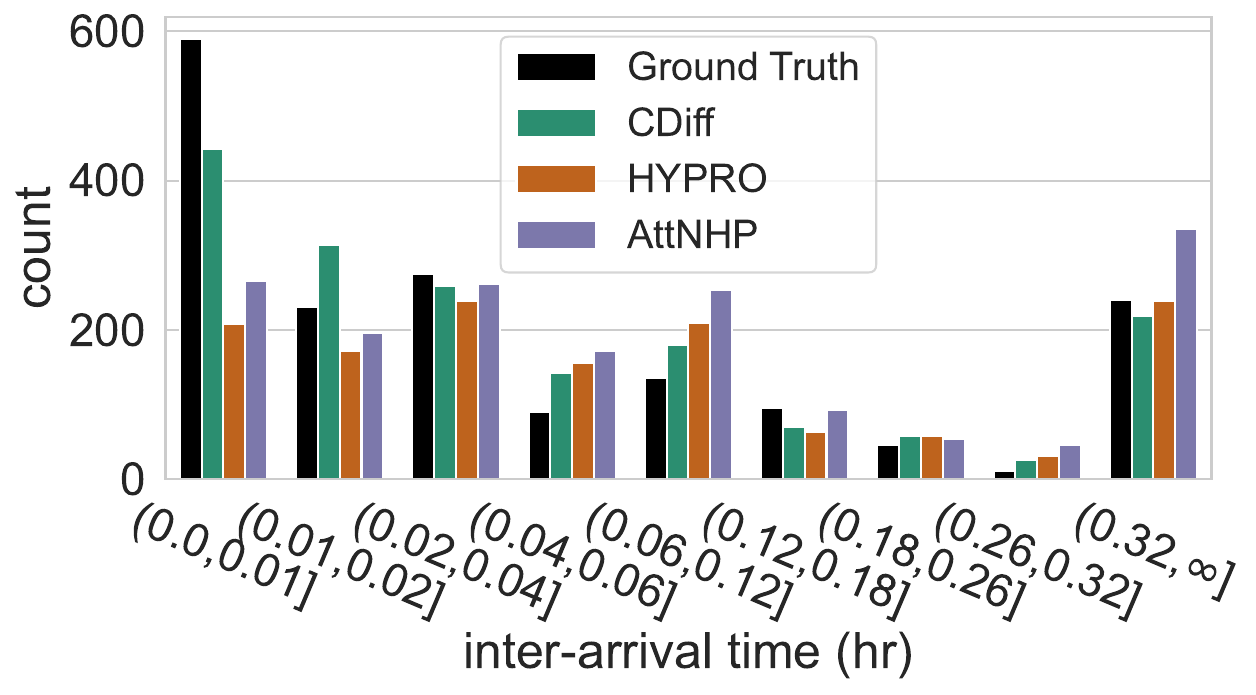}
    \includegraphics[width=0.4\textwidth]{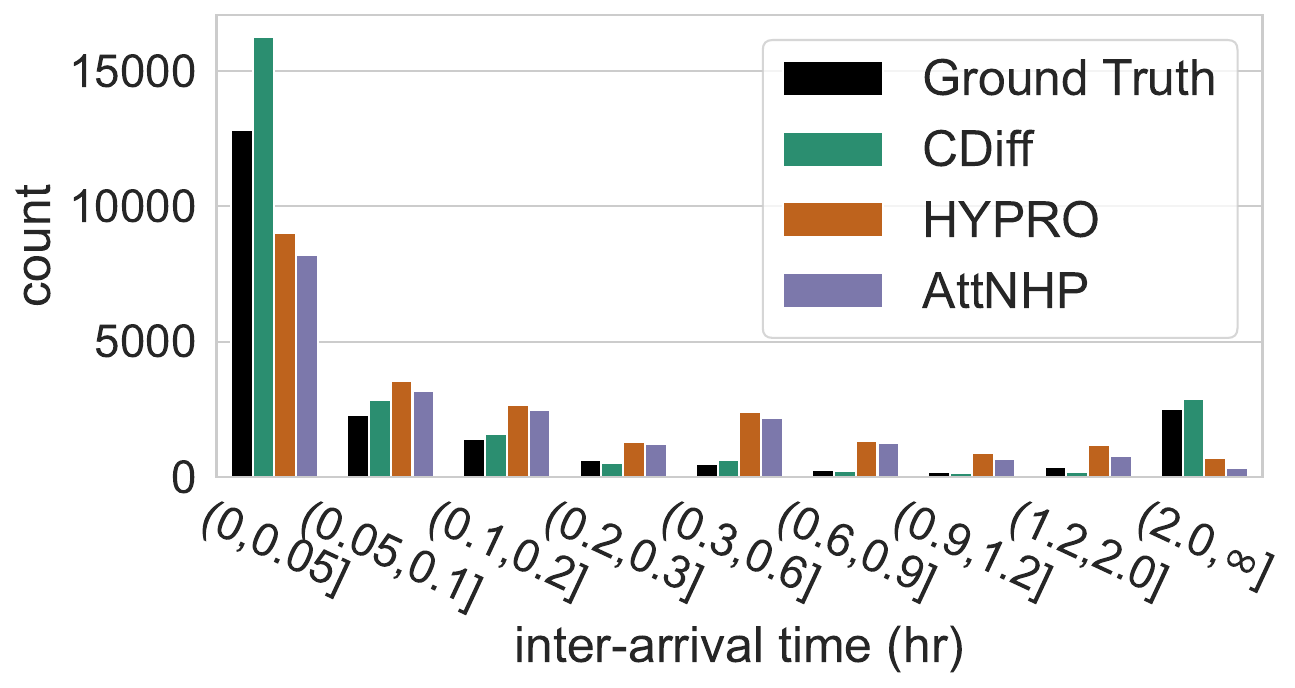}
    \caption{Histogram of true and predicted inter-arrival times for cases when the next event is type $e{=}7$ (top) and $e{=}16$ (bottom) for the Taobao dataset. Bin widths gradually increase so that counts are more comparable.}
    \label{fig:condition_on_time_dt_e}
\end{figure}

In contrast, HYPRO and AttNHP, the most competitive models, struggle to generate a sufficient number of values at the extremities of the marginal distribution. This also impacts the methods' ability to capture the joint relationship between time intervals and event types. To illustrate this, we consider two of the event categories for the Taobao dataset, and we plot the count histograms of the time intervals for categories 7 and 16 in Figure~\ref{fig:condition_on_time_dt_e}.  

First, it is noticeable that HYPRO and AttNHP fail to generate an adequate number of events for these specific categories, resulting in counts lower than the ground truth. In contrast, CDiff generates the appropriate quantity. This implies that CDiff is better at capturing the marginal categorical distribution of events. For both event types, the ground truth exhibits many very short intervals (the first bin) and then a rapid drop. CDiff manages to follow this pattern, while also accurately capturing the number of events in the tail (the final bin). HYPRO and AttNHP struggle to match the rapid decay. In the bottom panel, they also fail to produce many large inter-arrival times. These observations may be attributed to the fact that HYPRO and AttNHP rely on exponential distributions to model time intervals and are autoregressive whereas our architecture does not rely on a parametric TPP model and jointly models the distribution of the $N$ events in the sequence. 
\subsection{CDiff can forecast long horizon events}\label{sec:future}
{CDiff is explicitly designed to perform multi-event prediction, so we expect it to be better at predicting long horizon events, i.e., those near the end of the prediction horizon, such as events $N{-}1$ and $N$.
To verify this, we perform the following experiment.
We collect sequences of errors for time intervals. For each sequence to predict of length $N$, we compute a sequence of absolute errors made at each time interval:
$[\delta_1, \dots, \delta_N]$ where $\delta_i =  |x^+_i - \hat{x}^+_i| $. 
We can therefore construct a set of error sequences  $\{ [\delta^j_1, \dots, \delta^j_N] \}^M_{j=1}$ for each baseline for a given dataset of $M$ test sequences.

In general, a sequence of errors $[\delta_1, \dots, \delta_N]$ is expected to increase, as it is harder to predict events further in the future. Our goal is to compare how fast this error is growing for the various forecasting approaches.

To do so, we use a one-tailed Wilcoxon signed-paired test to test our method, CDiff, against each baseline for each error step $\delta_i$. We report the p-values for each different event index $i$. The tested null hypothesis is that the median of the population of differences between the paired data of CDiff error minus baseline error is equal or greater than zero. For later time intervals, rejection of the hypothesis implies that, with statistical significance, the median of the $\text{CDiff } \delta_i$  is smaller than the median of $\text{baseline } \delta_i$. In Table~\ref{tab:p value inter arrival time}, the p-values generally decrease as we move further into the future, showing an overall trend that the error of the competing baselines is increasing more rapidly than that of CDiff.}

\begin{table}[h]
    \caption{The p-values obtained from the Wilcoxon signed-paired tests, comparing CDiff vs HYPRO, AttNHP, NHP and Dual-TPP at various future steps $\delta_i$ (step $i=1, 5, 10, 20$). }
    \vskip 0.1in
    \centering\resizebox{\linewidth}{!}{\begin{tabular}{lcccc} \toprule
       \textbf{Taobao} & p-value $\delta_1$ &  p-value $\delta_5$ & p-value $\delta_{10}$ &  p-value $\delta_{20}$ \\ \midrule
    \textbf{HYPRO}      & $\ttt{5.10e-3}$ & $\ttt{1.704e-4}$ & $\ttt{1.855e-06}$ & $\ttt{2.099e-07}$ \\

      \textbf{AttNHP}     &  $\ttt{ 3.117e-1 }$  &$\ttt{6.157e-2}$ & $\ttt{1.149e-3}$& $\ttt{9.440e-4}$  \\ 

       \textbf{NHP}      & $\ttt{2.488e-09}$ & $\ttt{2.777e-09}$ & $\ttt{6.798e-11}$  & $\ttt{1.061e-13}$ \\    
       \textbf{Dual-TPP}      & $\ttt{2.030e-05}$ & $\ttt{1.511e-05}$ & $\ttt{2.368e-13}$  & $\ttt{3.725e-09}$  \\  
       \midrule
       \textbf{Stackoverflow} &  p-value $\delta_1$ &  p-value $\delta_5$ & p-value $\delta_{10}$ &  p-value $\delta_{20}$ \\ \midrule
    \textbf{HYPRO}      & $\ttt{1.396e-07}$  &$\ttt{1.913e-4}$ & $\ttt{1.585e-10}$ & $\ttt{1.427e-09}$ \\

      \textbf{AttNHP}     &  $\ttt{9.327e-4}$  & $\ttt{3.192e-4}$ & $\ttt{8.882e-06}$ & $\ttt{4.146e-07}$ \\ 

       \textbf{NHP}      &  $\ttt{4.671e-3}$    & $\ttt{8.490e-06}$  & $\ttt{1.769e-3}$  & $\ttt{3.127e-06}$ \\    
       \textbf{Dual-TPP}   &  $\ttt{3.816e-05}$    & $\ttt{8.887e-07}$  & $\ttt{1.194e-08}$  & $\ttt{3.542e-08}$ \\ 
        \bottomrule
    \end{tabular}}

    \label{tab:p value inter arrival time}
\end{table}

\subsection{Forecasting shorter horizons}
\begin{figure}
    \centering
    \begin{minipage}{0.5\linewidth}
        \centering
        \includegraphics[width=\linewidth]{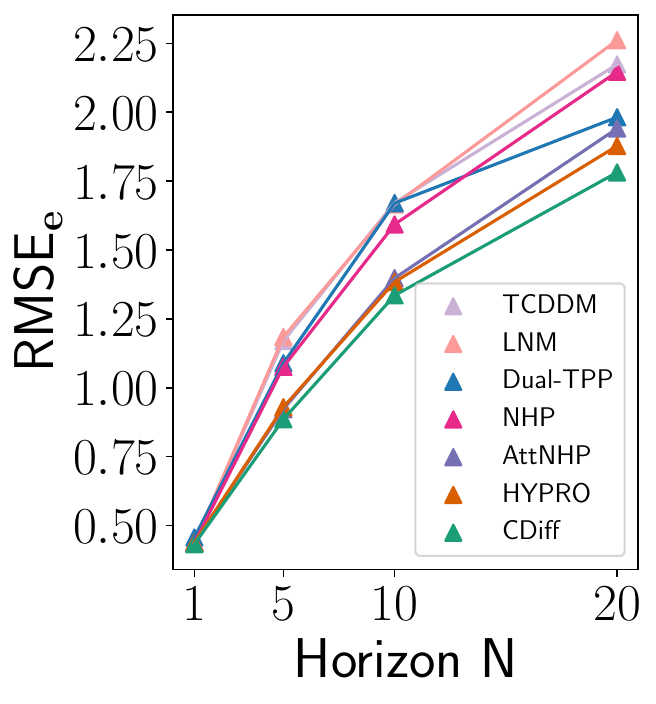} 
    \end{minipage}\hfill
    \begin{minipage}{0.48\linewidth}
        \centering
        \includegraphics[width=\linewidth]{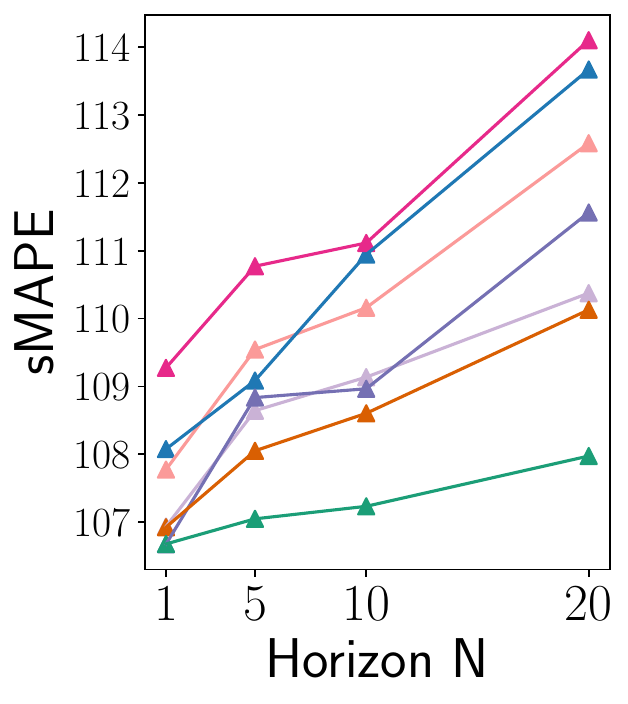} 
    \end{minipage}

    \caption{$RMSE_{e}$  and $sMAPE$, averaged over all datasets, for different horizons.}
    \label{fig:avg_rmse_e_smape_diff_horizon}
\end{figure}

\begin{table}[t]
  \caption{Complexity analysis on three datasets for $N=20$. The training time is for 500 epochs. The experiments were run on a GeForce RTX 2070 SUPER machine.}\label{tab:summary_sample_train_parameters}
      \vskip 0.1in
   \centering\resizebox{\linewidth}{!}{ \begin{tabular}{lllllllll} \toprule
                         &   & \textbf{HYPRO} & \textbf{DualTPP} & \textbf{Attnhp} & \textbf{LNM} & \textbf{TCDDM} &\textbf{CDiff} \\ \midrule
 \multirow{3}{*}{\rotatebox[origin=c]{90}{\textbf{Taxi}}} &\textbf{Sampling}  (sec/$\bs_u$ ) & 0.265          & 0.158            & 0.136           & \textbf{0.079} & 0.227     & \underline{0.104}          \\
&\textbf{num. param. }  (K)     & 40.5          & 40.1            & 19.3       & 19.1 & 20.3   & \textbf{17.1}          \\
&\textbf{Training}   (mins)   & 95             & 45               & \underline{35 }     & \textbf{20 } & {45 }       & \underline{35}             \\ \midrule
 \multirow{3}{*}{\rotatebox[origin=c]{90}{\textbf{Taobao}}} 
 &\textbf{Sampling}  (sec/$\bs_u$ ) & 0.325          & 0.240            & 0.209        & \textbf{0.094} & 0.285  & \underline{0.129 }         \\
&\textbf{num. param.} (K)       & 40.1          & 19.7            & { 19.6}         & \textbf{17.3} & {20.3} & 62.6          \\
&\textbf{Training  }  (mins)    & 105            & 60               & \underline{45}                   & \textbf{30}  & {60}  & \underline{45}             \\ \midrule
 \multirow{3}{*}{\rotatebox[origin=c]{90}{\textbf{Stack.}}} 
 &\textbf{Sampling}  (sec/$\bs_u$ )  & 0.294          & 0.207            & 0.191     & \textbf{0.111} & 0.233      & \underline{0.133}          \\
&\textbf{num. param.}  (K)      & 41.0          & 40.3            & 20.1         & \textbf{19.6} &  {20.3} & 63.9          \\
&\textbf{Training } (mins)      & 105            & 60               & \underline{45}                 & \textbf{35}  & {60}     & \underline{45}             \\ \bottomrule
    \end{tabular}}
\end{table}

Figure \ref{fig:avg_rmse_e_smape_diff_horizon} presents the results for shorter horizons: $N=1,5,10$. {See Appendix~\ref{sec:app_N1} for more detailed results on the single event forecasting case, i.e., $N=1$.} All methods improve as we reduce the forecasting horizon. For $\textbf{RMSE}_{\be}$, all models perform similarly to $N=20$. The performance difference grows as the prediction horizon increases. For $\textbf{sMAPE}$, CDiff outperforms the other models even for single event forecasting, and the outperformance increases rapidly with the prediction horizon. We attribute this to CDiff's ability to model more complex inter-arrival distributions.


\subsection{Sampling and training efficiency}\label{sec:complexity}

Table 4 summarizes the sampling time, number of trainable parameters, and training time for all methods across three datasets.  Starting with sampling time, CDiff outperforms most models, except for LNM, due to its non-autoregressive nature allowing for simultaneous generation of all events in the sequence. 
Regarding space complexity, CDiff naturally has the largest number of parameters (except for Taxi, which is a simpler task) as the dimension of the predicted vectors is $N$ times larger than all the other methods that generate one event at a time. {To account for the complexity difference and ensure that CDiff's superior performance is not due to additional parameters, we conduct further experiments with a fixed number of parameters in Appendix~\ref{sec:comparable_size}. }
In terms of training time, LNM is the most efficient, whereas CDiff, AttNHP, and NHP share similar training durations. Dual-TPP requires more time due to its count component, and HYPRO, which must also generate samples during training, demands the most training time.
\subsection{Ablation -- Joint vs Independent modeling of time and event type}
In the ablation study, we verify whether it is necessary to model the joint distribution in order to achieve better performance. In order to demonstrate this, we conduct a new experiment by introducing an independent model. In this model, we model the future sequence $P(S_0|\bs_c)$ using two independent processes (always conditioned on the same context $\bs_c$) $P(S_0|\bs_c) = P(X_0|\bs_c)P(E_0|\bs_c)$. In CDiff, we model the joint distribution by conditioning the time interval denoising distributions on the event type denoising distributions, and vice versa, as in Equations (9) and (10). In this ablation study we remove this interaction and strive to learn independent denoising distributions: 
\begin{align}
    Cat(E_{t-1}| \pi_{\theta}(E_t, t, \bs_c))\\
     \mathcal{N} (X_{t-1}| \mu_{\theta}(X_t, t, \bs_c),\sigma_t)
\label{eq:denoising functions removing}
\end{align}
The results are presented in Table \ref{tab:conditional independent}. For each metric and dataset we present, CDiff is always better than CDiff-indep, confirming that modeling the joint distribution is necessary. Moreover, we can see that CDiff-indep is not even the second best baseline; HYPRO is the second best or best method for half of the metrics, while CDiff-indep is the second best for 3/8 of the metrics. The ablation study thus highlights that: i) using a diffusion model to predict a sequence of multiple events is an effective strategy, even if the dependencies between event type and time interval are ignored (CDiff-indep is the second or third-best method); ii) modelling the dependencies via cross-diffusion leads to a significant performance improvement. 
\begin{table}[h]
    \caption{$\textbf{OTD}$, $\textbf{RMSE}_{e}$, $\textbf{RMSE}_{x^+}$ and \textbf{sMAPE} of Amazon and Stackoverflow reported in mean $\pm$ s.d. for comparison between conditional independent version of CDiff and the baselines. *indicates stat. significance w.r.t to the best method.}
          \vskip 0.1in
    \centering\resizebox{\linewidth}{!}{\begin{tabular}{lcccc} \toprule
        \textbf{Amazon} & $\textbf{OTD}$ & $\textbf{RMSE}_{e}$ &$\textbf{RMSE}_{x^+}$ & \textbf{sMAPE}\\ \midrule

         \textbf{HYPRO} & $\tts{\undl{38.61 $\pm$ 0.54}}$ & $\textbf{2.01 $\pm$ 0.05}$ & $\tts{0.48 $\pm$ 0.01}$ & \undl{82.51 $\pm$ 0.84} \\
         \textbf{Dual-TPP} & $\tts{42.65 $\pm$ 0.75}$ & $\ttt{2.56 $\pm$ 0.20}$ & $\tts{0.48 $\pm$ 0.01}$ & $\ttt{86.45 $\pm$ 2.04}$ \\
         \textbf{AttNHP} &     $\ttt{39.48 $\pm$ 0.33}$ & $\tts{2.17 $\pm$ 0.03}$ & {{0.48 $\pm$ 0.03}} & $\tts{84.32 $\pm$ 1.82}$ \\
         \textbf{NHP} &   $\tts{42.57 $\pm$ 0.29}$ & $\ttt{2.56 $\pm$ 0.06}$ & $\tts{0.52 $\pm$ 0.02}$ & $\tts{92.05 $\pm$ 1.55}$  \\

      \textbf{LNM}     & $\tts{43.82 $\pm$ 0.23}$ & $\tts{3.05 $\pm$ 0.29}$ & $\tts{0.48 $\pm$ 0.15}$ & $\tts{90.91 $\pm$ 1.61}$   \\ 
       \textbf{TCDDM}  & $\tts{42.25 $\pm$ 0.17}$ & $\tts{3.00 $\pm$ 0.12}$ & $\ttt{{0.48 $\pm$ 0.11}}$ & $\ttt{83.83 $\pm$ 1.51}$   \\      
  
       \midrule
       \textbf{CDiff-indep} &$ \tts{{40.49 $\pm$  0.60}} $& $\tts{{2.70 $\pm$ 0.24}}$ & $\ttt{\undl{0.47 $\pm$ 0.04}}$ &    $  \tts{84.77$\pm$ 1.32} $  \\
         \textbf{CDiff}    & $\textbf{37.73  $\pm$  0.20} $  & $\textbf{{2.09 $\pm$ 0.16 }}$ & $\textbf{{0.46 $ \pm$ 0.09}}$ & $\textbf{81.99 $\pm$ 1.91}$ \\ 
         \bottomrule \\

        \textbf{Stackoverflow} & $\textbf{OTD}$ & $\textbf{RMSE}_{e}$ &$\textbf{RMSE}_{x^+}$ & \textbf{sMAPE} \\ \midrule

         \textbf{HYPRO} &  $\text{{42.36$\pm$0.17}}$ & \undl{1.14 $\pm$ 0.01} & $\text{1.55 $\pm$ 0.01}^*$ &  $\text{ 110.99 $\pm$  0.56 }^*$ \\
         \textbf{Dual-TPP} & $\text{\underline{41.75$\pm$0.20}}$ &\textbf{1.13 $\pm$ 0.02} & $\text{1.51 $\pm$ 0.02}^*$ &   $\text{ 117.58  $\pm$ 0.42 }^*$ \\
         \textbf{AttNHP} &    $\text{42.59 $\pm$ 0.41}^*$  &1.14 $\pm$ 0.01 & $\text{{1.34 $\pm$ 0.01}}$ & {108.54 $\pm$  0.53}       \\
         \textbf{NHP} &   $\text{43.79  $\pm$ 0.15}^*$ &$\text{1.24  $\pm$ 0.03}^*$ & $\text{1.49 $\pm$ 0.00}^*$ &      $\text{ 116.95 $\pm$ 0.40}^*$       \\ 

      \textbf{LNM}   & $\tts{46.28 $\pm$ 0.89}$   & $\tts{  1.45 $\pm$ 0.06 }$  & $\tts{ 1.67 $\pm$ 0.01 }$  & $\tts{  115.12 $\pm$ 0.63 }$       \\ 
       \textbf{TCDDM}     & $\ttt{42.13 $\pm$ 0.59}$  & $\tts{ 1.47 $\pm$ 0.01}$  & { 1.32 $\pm$ 0.00}  & \undl{107.66 ± 0.93}     \\      
  
       \midrule
       \textbf{CDiff-indep} &  $\ttt{42.19 $\pm$ 0.14}$   & $\tts{{1.35 $\pm$ 0.12}}$ & $\ttt{\undl{1.26 $\pm$ 0.01} } $ & $\ttt{\undl{106.71 $\pm$ 0.44}}$\\
         \textbf{CDiff}   &   $\textbf{41.25 $\pm$ 1.40} $ &$\ttt{{1.14  $\pm$ 0.01}}$ & $\textbf{{1.20 $\pm$ 0.01}}$ & $\textbf{106.18 $\pm$ 0.34}$ \\ 
         \bottomrule

    \end{tabular}}

    \label{tab:conditional independent}
\end{table}

\section{Limitations}
Although offering impressive performance, there are limitations specific to our approach of modelling $N$ events at once. Unlike previous autoregressive approaches, our method requires the practitioner to select a fixed number of events $N$ to be modeled by the diffusion generative model. This can prove challenging when dealing with data that exhibits highly irregular time intervals ($x^+_i$). Essentially, if the length of time spanned by a fixed number of events varies significantly, then it will lead to a substantial variation in the nature and complexity of the forecasting task. This effect was not observed in the datasets we considered, as none displayed such high irregularities. 

\section{Conclusion}

We have proposed a diffusion-based generative model, CDiff, for event sequence forecasting. Extensive experiments demonstrate the superiority of our approach over existing baselines for long horizons. The approach also offers improved sampling efficiency. Our analysis sheds light on the mechanics behind the improvements, revealing that our model excels at capturing intricate correlation structure and at predicting distant events.

\section*{Acknowledgement}
We acknowledge the support of the Natural Sciences and Engineering Research Council of Canada (NSERC) [funding reference number 260250] and of the Fonds de recherche du Qu\'ebec.\\
Cette recherche a \'{e}t\'{e} financ\'{e}e par le Conseil de recherches en sciences naturelles et en g\'{e}nie du Canada (CRSNG), [num\'{e}ro de r\'{e}f\'{e}rence 260250] et par les Fonds de recherche du Qu\'{e}bec.
\section*{Impact Statement}
Forecasting methods for temporal point processes are impactful as they have many applications. In general, accurate forecasting has the typical positive impact of optimizing resource usage efficiency but also can raise privacy concerns. In particular for TPP models, the specific applications of these algorithms and even the datasets used to benchmark those models include monitoring consumer behavior. This poses a potential risk of malicious exploitation.
\bibliographystyle{icml2024}

\bibliography{main}

\clearpage
\newpage
\appendix
\onecolumn


\section{Appendix}
\subsection{Interval Forecasting}\label{sec:interval_forecasting}

In this time-based setting, the task is to predict the events that occur within a given subsequent time interval $t'$, i.e., $\bs^+_u$: ${{\bx^+_u}} = [{{x^+_{I+1}}},...] $ and $\be_u = [e_{I+1},...]$ such that $||{{\bx^+_u}}||_1 \leq t'$. 

This different setting also calls for different metrics, and the predicted $\hat{\bs}^+_u$ and ground truth $\bs^+_u$ can have a different number of events. We report both $\textbf{OTD}$ and the $\textbf{RMSE}_e$ metrics as they are robust to a varying number of events. We also report additional metrics that compare the number of events predicted:
\begin{enumerate}
 \item $  \textbf{MAE}_{|\bs^+|} = {\frac{1}{M} \sum_{j=1}^{M} \big||\bs_u^{+,j}| - |\hat{\bs}_u^{+,j}|\big|}$;
    \item   $\textbf{RMSE}_{|\bs^+|} = \sqrt{\frac{1}{M} \sum_{j=1}^{M} (|\bs_u^{+,j}| - |\hat{\bs}_u^{+,j}|)^2}$.   
\end{enumerate}

 For our experiment, we retain the same context sequences $\bs_c$ that were used for the \textbf{next $N$ events forecasting setting.}. Table~\ref{tab:time_interval_problem_setting_1} details the time interval values $t'$ of three experiments (long, medium and short horizon) for each dataset.

 \begin{table}[h]
    \small
    \caption{Time interval for interval forecasting problem.}


     \centering{%

    \begin{tabular}{lcccccc} \toprule 
       \textbf{Dataset} &  $t'$ long & $t'$ medium & $t'$ short & train/val/test & units\\ \midrule
       \textbf{Synthetic} &  2 & 1 & 0.5 & 1500/400/500 & second\\ 
       \textbf{Taxi} &  4.5 & 2.25 & 1.125 & 1300/200/400 & hour\\ 
       \textbf{Taobao} & 19.5 & 9.25 & 5.25 & 1300/200/500 & hour\\ 
      \textbf{Stack.} & 220 & 110 & 55 &  1400/400/400& day \\ 
       \textbf{Retweet} & 500 & 250 & 150 & 1400/600/800& second\\ 
       \textbf{MOOC} & 3.5 & 1.5 & 1 & 2400/717/1039 & hour\\ 
       \textbf{Amazon} & 20 & 10 & 5 & 3500/1000/1500 & hour\\ 
       \bottomrule
    \end{tabular}}

    \label{tab:time_interval_problem_setting_1}
\end{table}

\subsection{CDiff methodology for interval Forecasting}
To adapt our CDiff model to this setting, we select a number of events, denoted as $N$, and repeatedly generate $N$-length sequences until we reach the end of the forecasting window $t'$. That is, while $||\bx^+_u||_1 \leq t'$, we integrate the current $\bs^+_u$ into the context $\bs^+_c$ and regenerate $N$ additional events that we attach at the end of $\bs^+_u$. We set $N$ to be the maximum number of events observed within the given time interval in the training data.

\begin{figure*}

	\centering
	\subfigure[Synthetic dataset for $N=20$ forecasting]{
		\begin{minipage}{13.5cm}
            \includegraphics[width=\textwidth]{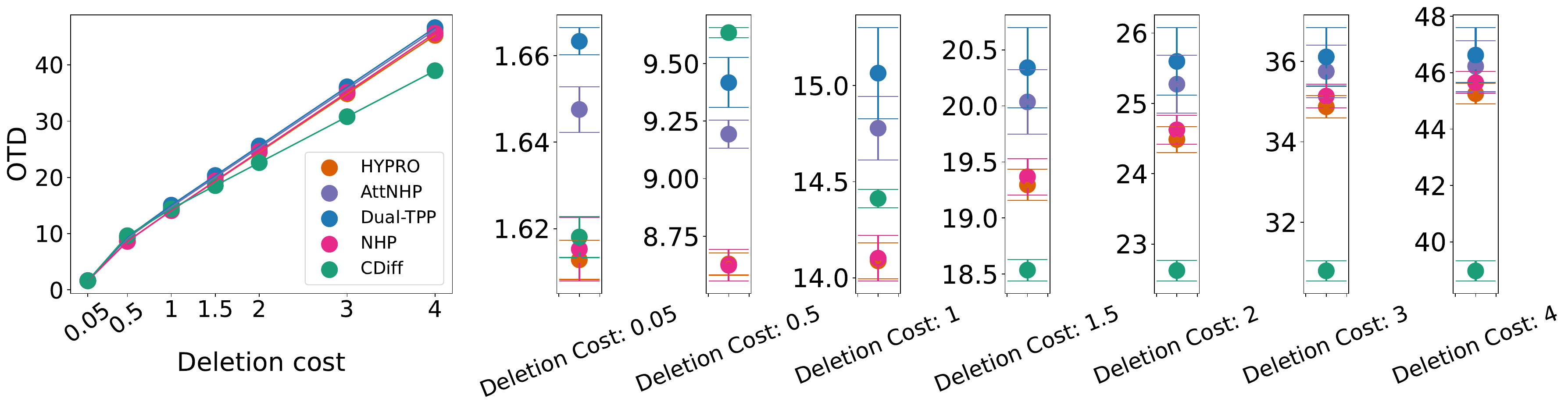} \\
		\end{minipage}
	}
	\subfigure[Taxi dataset without for $N=20$ forecasting]{
		\begin{minipage}{13.5cm}
			\includegraphics[width=\textwidth]{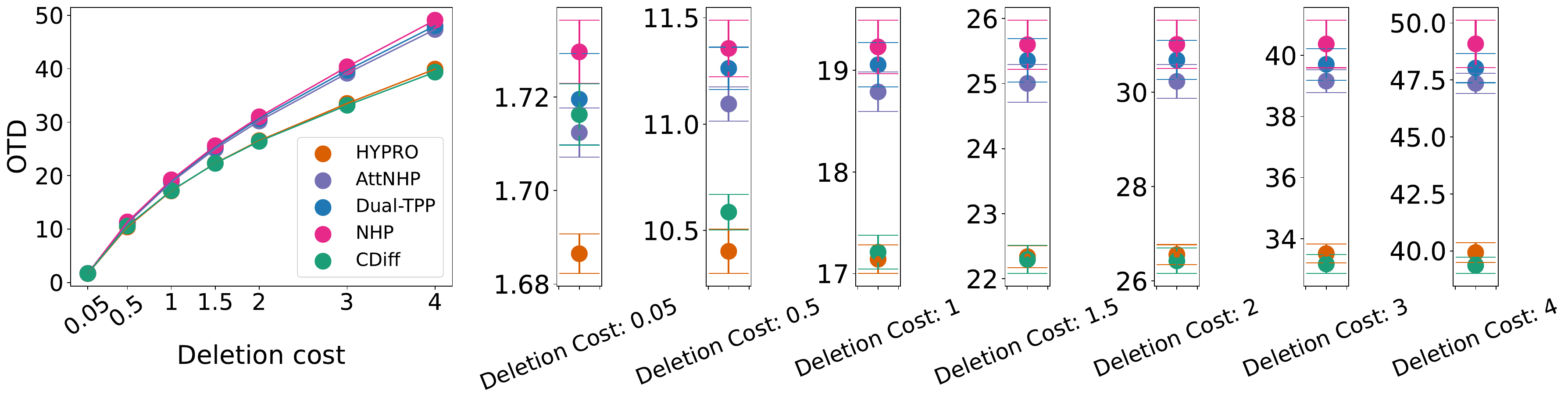} \\
		\end{minipage}
	}
    \subfigure[Stackoverflow dataset for $N=20$ forecasting]{
        \begin{minipage}{13.5cm}
            \includegraphics[width=\textwidth]{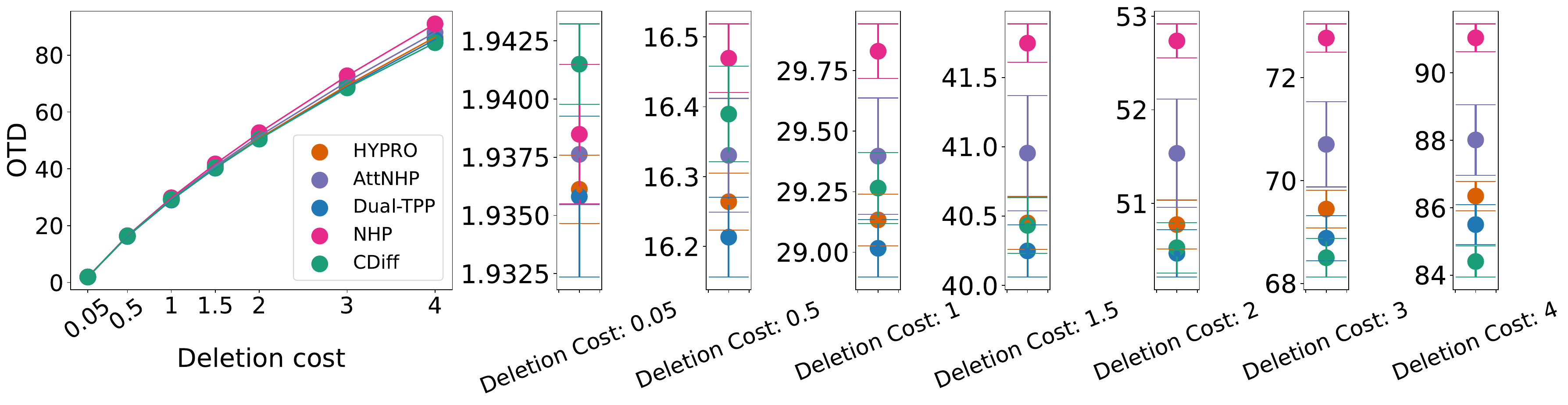}
        \end{minipage}
    }
    \subfigure[Taobao dataset $N=20$ forecasting]{
        \begin{minipage}{13.5cm}
            \includegraphics[width=\textwidth]{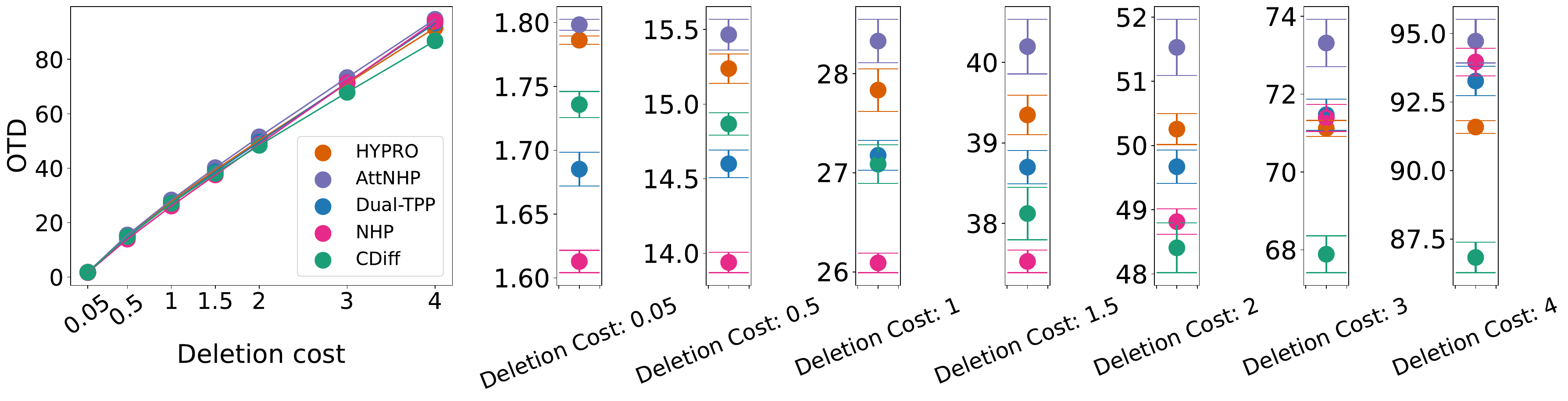}
        \end{minipage}
    }
    \subfigure[Retweet dataset $N=20$ forecasting]{
        \begin{minipage}{13.5cm}
            \includegraphics[width=\textwidth]{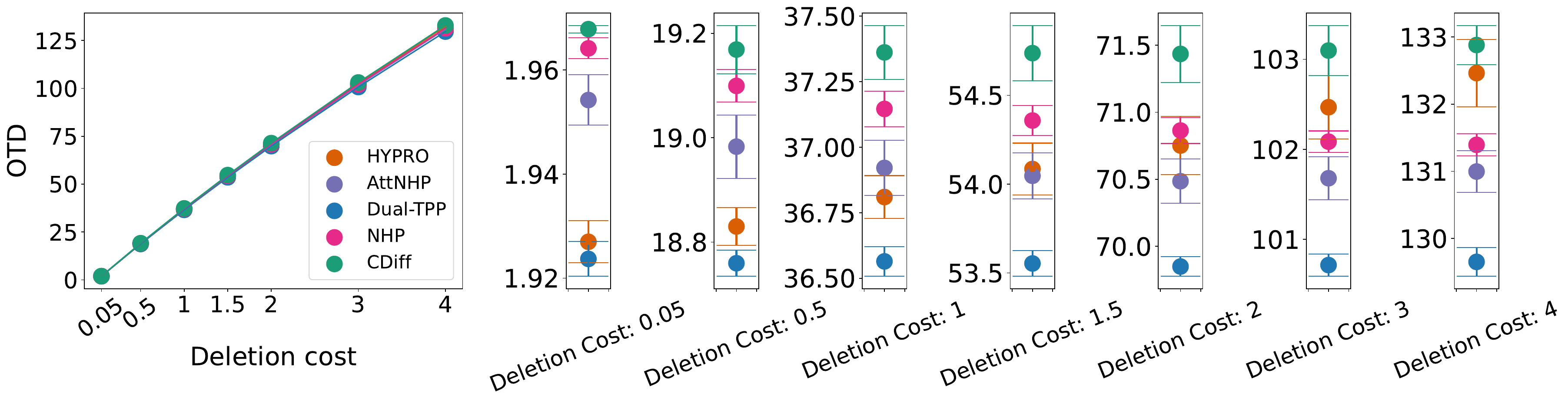}
        \end{minipage}
    }
	\caption{OTD for each specific deletion/addition cost for $N=20$ forecasting, we chose a variety of \(C_{\text{del}}\) values---\(0.05, 0.5, 1, 1.5, 2, 3, 4\)---based on the recommendations in \citep{xue2022HYPROHybridlyNormalizeda}. Subsequently, we calculated the mean and s.d. of OTD across all the datasets.} 
	\label{fig:delete_cost_vs_otd_1}
 
\end{figure*}

\begin{figure*}

\centering
    \subfigure[Synthetic dataset interval forecasting]{
        \begin{minipage}{13.5cm}
            \includegraphics[width=\textwidth]{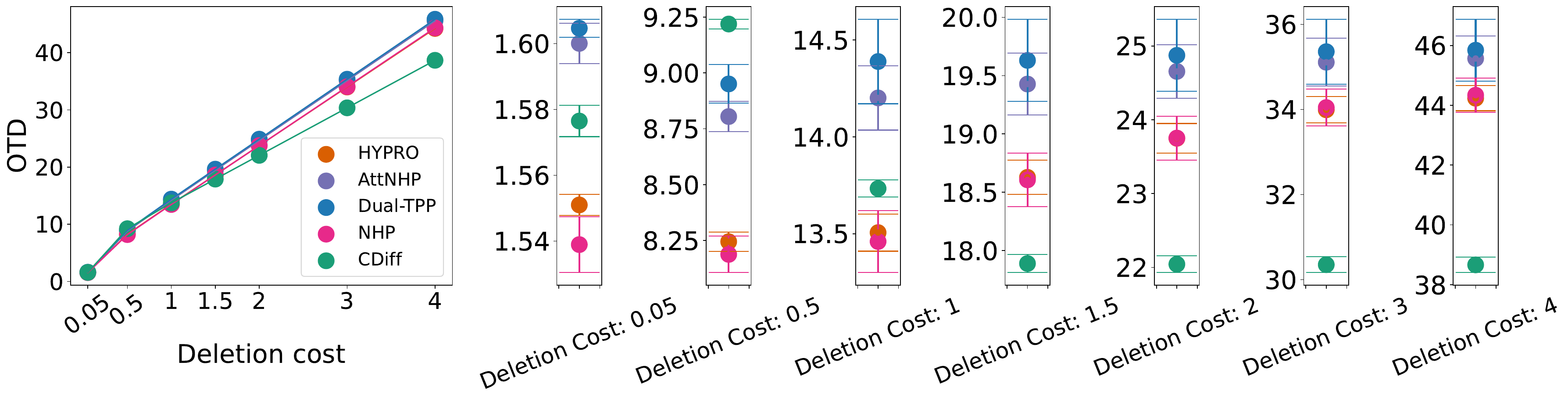}
        \end{minipage}
    }
    \subfigure[Taxi dataset interval forecasting]{
        \begin{minipage}{13.5cm}
            \includegraphics[width=\textwidth]{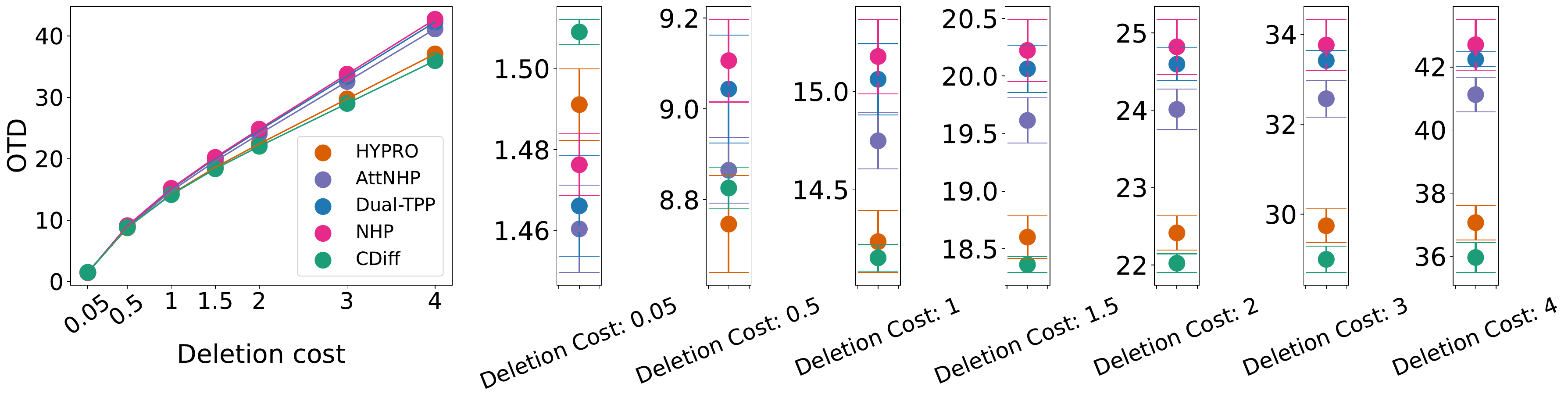}
        \end{minipage}
    }
    \subfigure[Stackoverflow dataset interval forecasting]{
        \begin{minipage}{13.5cm}
            \includegraphics[width=\textwidth]{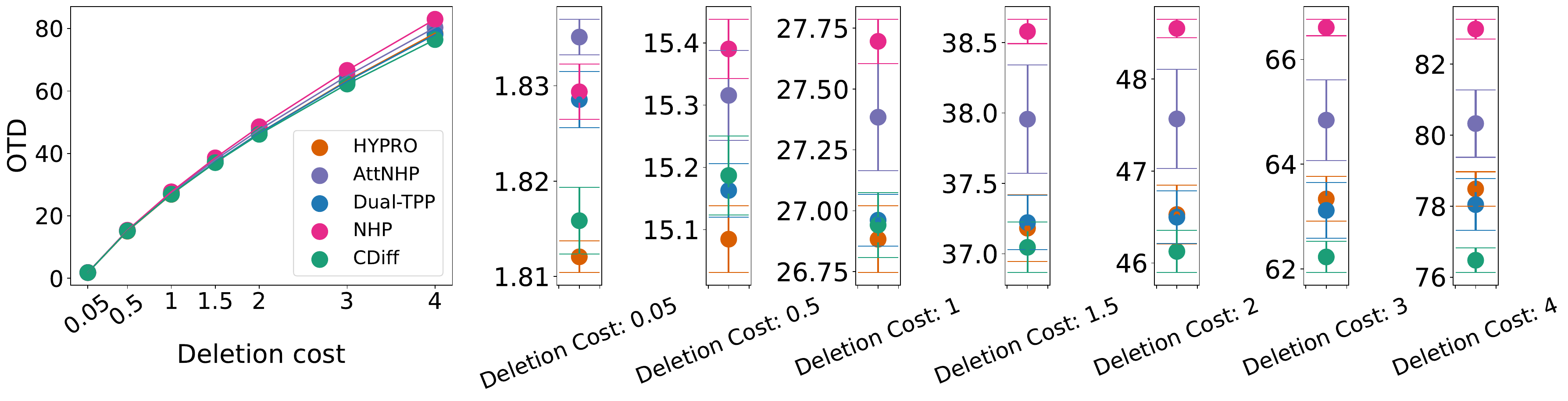}
        \end{minipage}
    }
    \subfigure[Taobao dataset interval forecasting]{
        \begin{minipage}{13.5cm}
            \includegraphics[width=\textwidth]{Figures/more_otd_results/delet_cost_vs_otd_with_filter_taobao_new.pdf}
        \end{minipage}
    }
    \subfigure[Retweet dataset interval forecasting]{
        \begin{minipage}{13.5cm}
            \includegraphics[width=\textwidth]{Figures/more_otd_results/delet_cost_vs_otd_with_filter_retweet_new.pdf}
        \end{minipage}
    }

	\caption{OTD for each specific deletion/addition cost for interval forecasting. We calculated the mean and s.d. of OTD across all the datasets for different \(C_{\text{del}}\) values.} 
	\label{fig:delete_cost_vs_otd_2}
 
\end{figure*}

\subsection{Metrics details and more OTD results}\label{sec:otd_and_more_otd_results}
The time-interval metrics are given by;
\begin{align}
RMSE_x &= \sqrt{\frac{1}{M}\sum_{j=1}^{M} ||\bx_u^{+,j} - \hat{\bx}_u^{+,j}}||_2^2,\\ 
    MAPE &= \frac{1}{M}\sum_{j=1}^{M} \frac{100}{N}
    \sum^N_{i=1}\frac{|x^{+,j}_{u,i} - \hat{x}^{+,j}_{u,i}|}{|x^{+,j}_{u,i}|}\\
    sMAPE &=\frac{1}{M}\sum_{j=1}^{M} \frac{100}{N} \sum^N_{i=1} \delta^j_i, \delta^j_i =  \frac{2|x^{+,j}_{u,i} - \hat{x}^{+,j}_{u,i}|}{|x^{+,j}_{u,i}|+|\hat{x}^{+,j}_{u,i}|}.
\end{align}

In the calculation of Optimal Transport Distance (OTD), the deletion cost hyperparameter, denoted by \(C_{\text{del}}\), plays a pivotal role. \citet{yang2022AttNHP} provided a full description and pseudo code for the dynamic algirthm to calculate the OTD. This parameter quantifies the expense associated with the removal or addition of an event token, irrespective of its category. For our experimentation, we chose a variety of \(C_{\text{del}}\) values---\(0.05, 0.5, 1, 1.5, 2, 3, 4\)---based on the recommendations provided by \cite{xue2022HYPROHybridlyNormalizeda}. Subsequently, we calculated the mean OTD. In the following section, the OTD metrics are delineated for each individual \(C_{\text{del}}\) value. As evidenced by Fig. \ref{fig:delete_cost_vs_otd_1} and \ref{fig:delete_cost_vs_otd_2}, our model outperforms across the board for the varying \(C_{\text{del}}\) settings overall. We also see that the OTD steadily increases overall, and that different C can permute the ordering of the competing baselines. For low \(C_{\text{del}}\), our method is outperformed by HYPRO and AttNHP sometimes, but this trend is reversed for larger \(C_{\text{del}}\) values for almost all datasets. This reflects the fact that the proposed CDiff method is better at predicting the number of events, so fewer deletions or additions are required. 

\subsection{Dataset details}\label{sec:dataset}
\begin{itemize}
    \item $\textbf{Taobao}$ \citep{alibaba2018tbdataset} This dataset captures user click events on Taobao's shopping websites between November 25 and December 03, 2017. Each user's interactions are recorded as a sequence of item clicks, detailing both the timestamp and the item's category. All item categories were ranked by frequency, with only the top 16 retained; the remaining were grouped into a single category. Thus, we have $K=17$ distinct event types, each corresponding to a category. The refined dataset features 2,000 of the most engaged users, with an average sequence length of 58.  The disjoint train, validation and test sets consist of 1300, 200, and 500 sequences (users), respectively, randomly sampled from the dataset. The time unit is $3$ hours; the average inter-arrival time is $0.06$ (i.e., $0.18$ hour). 

    \item $\textbf{Taxi}$ \citep{whong2014taxidataset} This dataset contains time-stamped taxi pickup and drop off events with zone location ids in New York city in 2013. Following the processing procedure of \cite{mei2019ImputingMissingEvents}, each event type is defined as a tuple of (location, action). The location is one of the $5$ boroughs (Manhattan, Brooklyn, Queens, The Bronx, Staten Island). The action can be either pick-up or drop-off. Thus, there are $K = 5 \times 2 = 10$ event types in total. The values $k=0,\ldots,4$ indicate pick-up events and $k=5,\ldots,9$ indicate drop-off events. A subset of 2000 sequences of taxi pickup events with average length 39 are retained. The average inter-arrival time is $0.22$ hour (time unit is $1$ hour.) The disjoint train, validation and test sets are randomly sampled and are of sise 1400, 200, and 400 sequences, respectively.

    \item $\textbf{StackOverflow}$ \citep{leskovec2014sodataset} This dataset contains two years of user awards from a question-answer platform. Each user was awarded a sequence of badges, with a total of $K=22$ unique badge types. The train, validation and test sets consist of $1400$, $400$ and $400$ sequences, resepctively, and are randomly sampled from the dataset. The time unit is $11$ days; the average inter-arrival time is $0.95$.

    \item $\textbf{Retweet}$ \citep{zhou2013retweetdata} This dataset contains sequences of user retweet events, each annotated with a timestamp. These events are segregated into three categories ($K=3$), denoted by: ``small'', ``medium'', and ``large'' users. Those with under 120 followers are labeled as small users; those with under 1363 followers are medium users, while the remaining users are designated as large users. Our studies focus on a subset of 9000 retweet event sequences. The disjoint train, validation and test sets consist of 6000, 1500, and 1500 sequences, respectively, randomly sampled from the dataset.

    \item {$\textbf{MOOC}$ \cite{kumar2019moocdataset} This datasets contains sequences of records of student interactions within an online course platform. Each interaction represents an event and can manifest in different forms (97 distinct types), such as viewing a video, completing a quiz, and other activities. We utilized the pre-processing approach described by \cite{bosser2023} in their extensive study on temporal point processes. This involved narrowing down the event types to a total of 50. Observing that a significant number of event sequences had less than or equal to 20 events, we chose to exclude these shorter sequences. Consequently, this process resulted in retaining 4,156 out of the initial 7,047 sequences, focusing on those with more than 20 events.}

    \item {$\textbf{Amazon}$ \cite{ni2019amazondataset} 
    The dataset contains time-stamped user product review behavior from January 2008 to October 2018. It consists of sequences of product review events for individual users. Each event in these sequences includes the timestamp and the category of the product reviewed, with every category corresponding to a distinct event type. The study is conducted on a subset comprising the 5200 most active users, each having an average sequence length of 70 events. This led to a refinement of the event types to a total of $K = 16$.}

    \item $\textbf{Synthetic Multivariate Hawkes Dataset}$ The synthetic dataset is generated using the \textbf{tick}\footnote{\textbf{tick} package can be found at \url{https://github.com/X-DataInitiative/tick}} package provided by \citet{bacry2018tickpackage}, using the Hawkes process generator. Our study uses the same equations proposed by \citet{Lin2022}. There are 5 event types. The impact function $g_{j,i}(y)$ measuring the relationship (impact) of type $i$ on type $j$ and is uniformly-randomly chosen from the following four functions: 
    \begin{align}
    \begin{split}
        g_a (y) &= 0.99 \exp (−0.4y) \\
        g_b (y) &= 0.01 \exp (-0.8y) + 0.03\exp(−0.6y) + 0.05 \exp(−0.4y) \\
        g_c (y) &= 0.25 | \cos 3y | \exp(−0.1y) \\
        g_d (y) &= 0.1(0.5 + y)^2
        \label{eq:syn_dataset}
    \end{split}
    \end{align}

\end{itemize}

\begin{figure}
    \centering
    \includegraphics[width=0.6\textwidth]{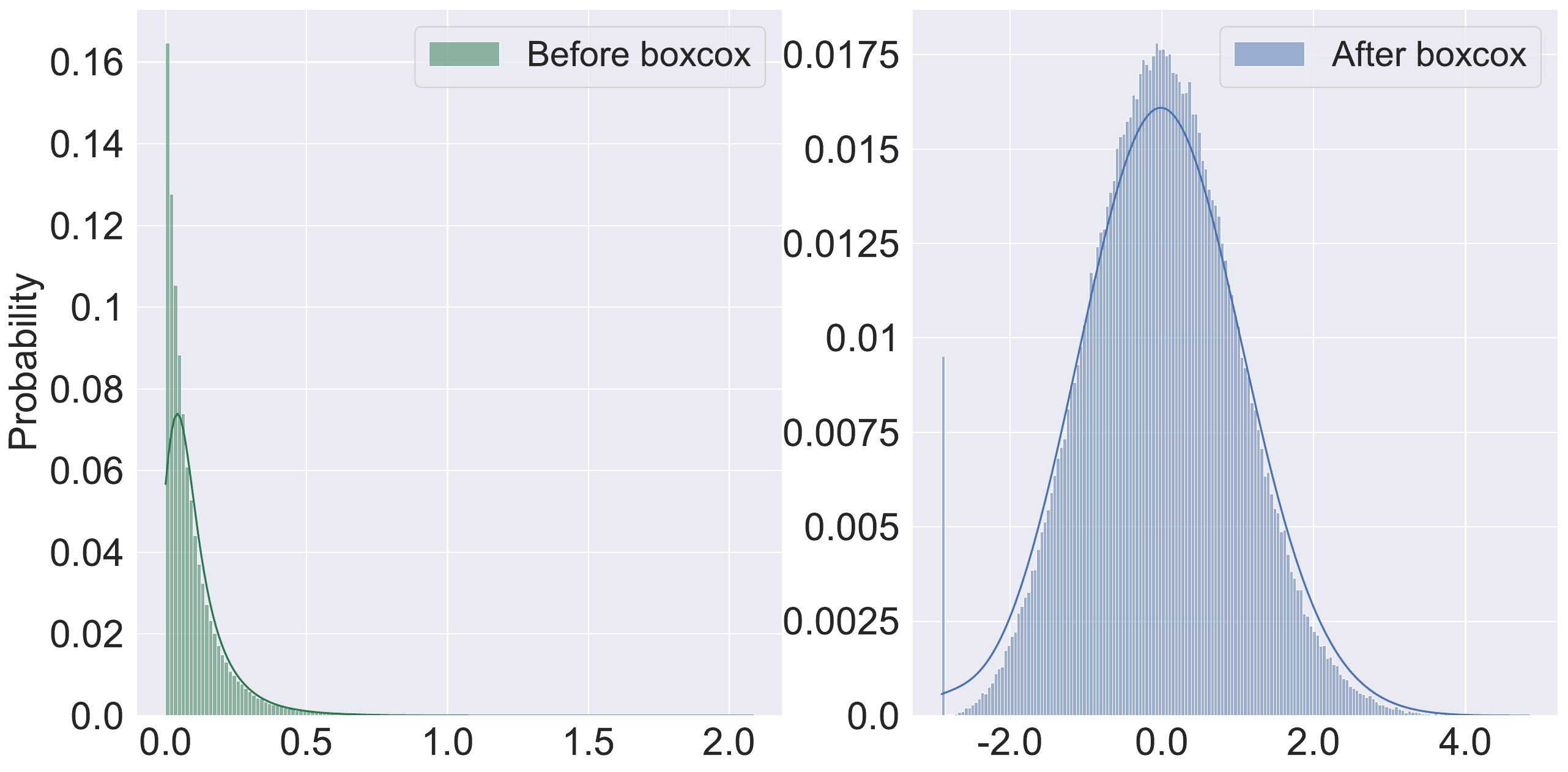}
    \caption{Inter-arrival time marginal histogram for synthetic dataset before (left) and after (right) boxcox transformation}
    \label{fig:box_cox_syn}
\end{figure}

\subsection{Box-Cox Transformation}
\label{sec:boxcox_transform}
For our study, the inter-arrival time marginal distribution shown in Fig.\ref{fig:box_cox_syn} (left) is clearly not a normal distribution. Since the diffusion probabilistic model we employ is a Gaussian-based generative model, we use the Box-Cox transformation to transform the inter-arrival time data, so that the transformed data approximately obeys a normal distribution.

The Box-Cox transformation \citep{tukey1957boxcoxtransformation} is a family of power transformations that are used to stabilize variance and make data more closely follow a normal distribution. The transformation is defined as:
\begin{align}
x(\lambda) = 
\begin{cases} 
    \frac{x^\lambda - 1}{\lambda} & \text{if } \lambda \neq 0, \\
    \log(x) & \text{if } \lambda = 0. 
\end{cases}
\label{eq:boxcox_transformation}
\end{align}
Here:
\begin{itemize}
    \item $x$ is the original data;
    \item $ x(\lambda)$  is the transformed data; and
    \item $\lambda$ is the transformation parameter.
\end{itemize}

The inter-arrival time is strictly larger than 0 but it can be extremely small because of the scale of the dataset. Therefore, in order to prevent numerical errors in tbe Box-Cox transformation we add $1\times 10^{-7}$ time units to all inter-arrival times. We then scale all values by $100$. We use the scaled inter-arrival time data from the train set to obtain the fitted $\lambda$ shown in Eq.\ref{eq:boxcox_transformation} and apply the transformation with the fitted $\lambda$ to the inter-arrival time data for both the validation dataset and test dataset. Fig.\ref{fig:box_cox_syn} shows an example of marginal histogram of inter-arrival time for the Synthetic train set before (left) and after (right) the Box-cox transformation. We transform back the predicted sequence inter-arrival times with the same fitted $\lambda$ obtained from the train set and undo the scaling by 100. We use the Box-cox transformation function from the \textbf{SciPy}\footnote{The SciPy package is available at \url{https://github.com/scipy/scipy}} package provided by \citet{virtanen2020scipy}.

\begin{table}[h]
    \caption{Sets of hyperparameters. Underlined values are those selected by the Tree-Structured Parzen Estimator~\citep{bergstra2011algorithmshpo}}
    \centering \resizebox{\columnwidth}{!}{%
    \begin{tabular}{lcccccccccc} \toprule 
       \textbf{Parameters}   & num. heads & num. layers & time embedding & Transformer feed-forward embedding & num. diffusion steps & LR \\ \midrule
       \textbf{Synthetic} & \{1,$\underline{2}$,4\} & \{\underline{1}, 2, 4\} & \{4, 8, \underline{16}, 32, 64, 128\}  & \{8, {16}, \underline{32}, {64}, 128, 256\} & \{50, {100}, \underline{200}, 300, 500\} & \{0.001, 0.0025, \underline{0.005}\} \\ 
       \textbf{Taxi} & \{1,$\underline{2}$,4\} & \{\underline{1}, 2, 4\} & \{4, \underline{8}, {16}, 32, 64, 128\} & \{8, \underline{16}, 32, 64, 128, 256\} & \{50, \underline{100}, 200, 300, 500\} & \{0.001, 0.0025, \underline{0.005}\} \\
       \textbf{Taobao} & \{1,$\underline{2}$,4\} & \{\underline{1}, 2, 4\}  & \{4, {8}, {16}, \underline{32}, 64, 128\} & \{8, {16}, 32, \underline{64}, 128, 256\} & \{50, {100}, \underline{200}, 300, 500\} & \{\underline{0.001}, 0.0025, {0.005}\} \\
       \textbf{Stackoverflow} & \{1,$\underline{2}$,4\} & \{\underline{1}, 2, 4\}  & \{4, {8}, {16}, \underline{32}, 64, 128\} & \{8, {16}, 32, \underline{64}, 128, 256\}  & \{50, {100}, \underline{200}, 300, 500\} & \{ {0.001}, \undl{0.0025}, {0.005}\} \\
       \textbf{Retweet} & \{1,$\underline{2}$,4\} & \{\underline{1}, 2, 4\} & \{4, {8}, {16}, \underline{32}, 64, 128\} & \{8, {16}, 32, \underline{64}, 128, 256\}  & \{50, {100}, \underline{200}, 300, 500\} & \{0.001, \underline{0.0025}, {0.005}\} \\
       \textbf{MOOC} & \{1,$\underline{2}$,4\} & \{\underline{1}, 2, 4\} & \{4, {8}, {16}, \underline{32}, 64, 128\} & \{8, {16}, 32, \underline{64}, 128, 256\}  & \{50, {100}, \underline{200}, 300, 500\} & \{0.001, \underline{0.0025}, {0.005}\} \\
       \textbf{Amazon} & \{1,$\underline{2}$,4\} & \{\underline{1}, 2, 4\} & \{4, {8}, {16}, \underline{32}, 64, 128\} & \{8, {16}, 32, \underline{64}, 128, 256\}  & \{50, {100}, \underline{200}, 300, 500\} & \{0.001, \underline{0.0025}, {0.005}\} \\
       \bottomrule
    \end{tabular} }

    \label{tab:hyper-param_sets}
\end{table}

\subsection{Hyper-parameters}
\label{sec:hyper-parameters}
Table \ref{tab:hyper-param_sets} specifies the hyperparameters that we use for our experiments and the candidate values. We train for a maximum of 500 epochs and we select the best hyperparameters using the Tree-Structured Parzen Estimator \citep{bergstra2011algorithmshpo}.
{
We have also performed a sensitivity study for the number of diffusion steps. As we can see in the following table, the method is not too sensitive to the number of diffusion steps. There is no sudden variation of performance as we gradually decreases the number of diffusion steps. A too low number of steps (25 and 50 steps in our case)  is worst overall for all metrics and datasets. Once we use more steps (100, 200 or 500 steps) the performance becomes similar and the hyperparameter search sill select the best number of steps for each dataset. 
\begin{table*}[h]
    \caption{Ablation study on the number of diffusion steps. *indicates stat. significance w.r.t to the best method.}
    \centering\resizebox{0.6\linewidth}{!}{\begin{tabular}{lcccc} \toprule
       $\textbf{Taxi (Diffusion Step 100)}$ & $\textbf{OTD}$ & $\textbf{RMSE}_{e}$ &$\textbf{RMSE}_{x^+}$ & \textbf{sMAPE} \\ \midrule
   
         \textbf{CDiff}    & \textbf{21.013  $\pm$  0.158}   & \textbf{{1.131 $\pm$ 0.017}} & \textbf{{0.351 $ \pm$ 0.004}}& $\ttt{87.993 $\pm$ 0.178}$ \\
         \textbf{CDiff-25} & $\ttt{22.083 $\pm$ 0.410 }$ & $\ttt{1.135 $\pm$ 0.022}$ & $\textbf{0.351 $\pm$ 0.004}$ &  $\ttt{\undl{87.963 $\pm$ 0.252}}$ \\
         \textbf{CDiff-50} & $\ttt{\undl{21.045 $\pm$ 0.228}}$ & $\textbf{1.131 $\pm$ 0.019}$ & $\ttt{0.352 $\pm$ 0.007}$ &  $\ttt{88.129 $\pm$ 0.193}$ \\
          \textbf{CDiff-100} & $\ttt{--}$ & $\ttt{{--}}$ & $\ttt{{--}}$ & $\ttt{--}$ \\
         \textbf{CDiff-200} & \ttt{21.545 $\pm$ 0.314} & $\ttt{\undl{1.133 $\pm$ 0.015 }}$ & $\textbf{{0.351 $\pm$ 0.010}}$ & $\textbf{87.839 $\pm$ 0.397}$  \\
         \textbf{CDiff-500} & \ttt{22.107 $\pm$ 0.244} & $\ttt{{1.138 $\pm$ 0.036}}$ & $\ttt{{0.353 $\pm$ 0.009}}$& $\ttt{88.053 $\pm$ 0.480}$  \\
         
         \bottomrule \\
         \textbf{StackOverflow (Diffusion Step 200)}& $\textbf{OTD}$ & $\textbf{RMSE}_{e}$&$\textbf{RMSE}_{x^+}$ & \textbf{sMAPE}\\ \midrule
         \textbf{CDiff} & \textbf{41.245 $\pm$ 1.400 } & $\ttt{\undl{1.141 $\pm$ 0.007}}$ & \textbf{{1.199 $\pm$ 0.006}}&      \textbf{106.175$\pm$ 0.340}  \\
         \textbf{CDiff-25} & $\tts{42.742 $\pm$ 0.146}$ & $\tts{{1.169 $\pm$ 0.030}}$ & $\tts{{1.331 $\pm$ 0.016}}$ &      $\tts{ 109.941$\pm$ 0.322} $\\
         \textbf{CDiff-50} & \ttt{42.094 $\pm$ 0.444} & $\tts{{1.172$\pm$ 0.042}}$ & \ttt{{1.306 $\pm$ 0.008}}&      \ttt{ \undl{107.055$\pm$ 0.400}} \\
         \textbf{CDiff-100} & \ttt{41.578 $\pm$ 0.261} & $\textbf{{1.139 $\pm$ 0.017}}$ & \ttt{{1.27 $\pm$ 0.022}}&      \ttt{105.365 $\pm$ 0.59} \\
         \textbf{CDiff-200} & $\ttt{--}$ & $\ttt{{--}}$ & $\ttt{{--}}$ & $\ttt{--}$  \\
         \textbf{CDiff-500} & \ttt{\undl{41.507 $\pm$ 0.16}} & $\ttt{{1.153 $\pm$ 0.019}}$ & $\ttt{\undl{1.181 $\pm$ 0.23}}$&      \ttt{107.842 $\pm$ 0.500} \\

         \bottomrule

    \end{tabular}}

    \label{tab:rebuttal whole table}
\end{table*}
}

{
\subsection{Comparison with fixed model size}\label{sec:comparable_size}

To ensure a fair comparison, we conducted an experiment to compare CDiff and AttNHP with a similar number of parameters. We employed two methods to increase the number of parameters for AttNHP.

In the first way, we attach additional inference heads to the autoregressive baselines, until the number of parameters matches the size of our model. We denote the number of additional heads that are predicting future events in the name (for example, \textbf{method} with 2 heads is denoted as \textbf{method-2}). If the baseline has 2 additional heads, the first head is trained to predict the next event of the sequence (as usual), and the second head is trained to predict the second future event of the sequence. At inference, the model predicts the next $2$ events, then integrates those $2$ events into the context sequence $s_c$ to predict the next 2 events (which would then be the 3rd and 4-th prediction) until $N$ events have been predicted. 

In the second, we increase the number of parameters for the baselines to match the number of parameters of our method. We denote it by \textbf{method-L}. Since we already used a hyperparameter search for each baseline, the increased number of parameters did not improve the results. We include it for completeness.

In the Table.\ref{tab:more heads exp}, we can see that neither the inclusion of additional heads nor the increase in the number of parameters is sufficient to reach CDiff's performance. Adding additional heads actually hurts the performance; for all datasets and all metrics, \textbf{AttNHP-3} is worse than the initial baseline \textbf{AttNHP}. 

Although it is true that our model has more parameters than the baselines (approximately 1.5x more), it is better than the baselines in terms of the other complexity metrics that we report (namely training time and sampling time). 

\begin{table*}[h]
    \caption{Comparison between multi-head AttNHP, AttNHP with more parameters and CDiff. $\textbf{OTD}$, $\textbf{RMSE}_{e}$, $\textbf{RMSE}_{x^+}$ and \textbf{sMAPE} of real-world datasets reported in mean $\pm$ s.d. Best are in bold, the next best is underlined.}
    \vskip 0.1in
    \centering\resizebox{0.5\linewidth}{!}{\begin{tabular}{lcccc} \toprule
       \textbf{Taxi} & $\textbf{OTD}$ & $\textbf{RMSE}_{e}$ &$\textbf{RMSE}_{x^+}$ & \textbf{sMAPE}\\ \midrule
      \textbf{AttNHP}     &  ${24.762 \pm 0.217}^*$  &$\text{1.276  $\pm $ 0.015}^*$ & ${\undl{0.430 \pm 0.003}}^*$& $\text{\underline{97.388 $\pm$  0.381}}^*$ \\ 

      \textbf{AttNHP-3}     &  $\tts{26.376$\pm$  0.229}$  &$\tts{{1.554 $\pm$ 0.022}}$ & $\tts{0.452$\pm$  0.005}$& $\tts{105.860$\pm$  0.504}$ \\ 
      
      \textbf{AttNHP-L}     &  $\tts{\undl{ 24.174$\pm$  0.245}}$  &$\tts{\undl{1.274 $\pm$ 0.022}}$ & $\tts{0.434 $\pm$ 0.002}$& $\tts{97.645$\pm$  0.693}$ \\ 

       \midrule
         \textbf{CDiff}    & \textbf{21.013  $\pm$  0.158}   & \textbf{{1.131 $\pm$ 0.017}} & \textbf{{0.351 $ \pm$ 0.004}}& \textbf{87.993 $\pm$ 0.178} \\ \bottomrule  \\
      \textbf{Taobao}  & $\textbf{OTD}$ & $\textbf{RMSE}_{e}$ &$\textbf{RMSE}_{x^+}$ & \textbf{sMAPE} \\ \midrule
      \textbf{AttNHP} &  $\ttt{\undl{45.555 $\pm$ 0.345}}^*$  & $\ttt{\undl{2.737 $\pm$ 0.021}}$ & $\ttt{0.708 $\pm$ 0.010}^*$ & $\ttt{{134.582 $\pm$ 0.920}}^*$ \\ 

      \textbf{AttNHP-3}     &  $\tts{48.967$\pm$  0.072}$  & $\tts{3.877$\pm$  0.012}$ & $\tts{0.933$\pm$  0.005}$ & $\tts{{136.130$\pm$  0.619}}$ \\ 
      
      \textbf{AttNHP-L}    &  $\tts{46.515$\pm$  0.191}$  & $\ttt{{2.897 $\pm$ 0.019}}$ & $\ttt{\undl{0.697$\pm$  0.005}}$ & $\tts{\undl{132.276$\pm$  0.993}}$ \\ 

       \midrule
         \textbf{CDiff}    & \textbf{44.621 $\pm$ 0.139}  &\textbf{{2.653 $\pm$ 0.022}} & \textbf{{0.551 $\pm$ 0.002}} & \textbf{125.685 $\pm$ 0.151} \\ \bottomrule  

    \end{tabular}}

    \label{tab:more heads exp}
\end{table*}
}

{
\subsection{Add-and-thin comparison}
\citet{luedke2023add} proposed Add-and-thin model to perform multi-step forecasting for time intervals only. There is no consideration or modeling of event type.
Since our work is focused on modeling the joint interaction between time intervals and event types, we cannot directly  comparison with this method. 
However, for completeness, we can include a modified version by augmenting the add-thin-add model with a simple event type predictor module. This event type predictor model is based on the marginal probabilities of the training set. 
As we can see in the Table.\ref{tab:rebuttal whole table}, this modified \textbf{Add-and-thin-augm.} is outperformed by CDiff for both the event type metrics and the time interval metrics, further demonstrating the importance of modeling the joint interaction of type and time. We would stress, however, that this modified version of Add-and-thin was not presented in the original paper.

\begin{table*}[h]
    \caption{$\textbf{OTD}$, $\textbf{RMSE}_{e}$, $\textbf{RMSE}_{x^+}$ and \textbf{sMAPE} of real-world datasets reported in mean $\pm$ s.d. Best are in bold, the next best is underlined. *indicates stat. significance w.r.t to the best method.}
    \vskip 0.1in
    \centering\resizebox{0.5\linewidth}{!}{\begin{tabular}{lcccc} \toprule \\
      $\textbf{Taxi}$  & $\textbf{OTD}$ & $\textbf{RMSE}_{e}$ &$\textbf{RMSE}_{x^+}$ & \textbf{sMAPE} \\ \midrule
    \textbf{HYPRO}      & \underline{21.653 $\pm$ 0.163} & $\text{\underline{1.231 $\pm$ 0.015}}^*$ & $\text{\underline{0.372 $\pm$ 0.004}}^*$ & $\text{\underline{93.803 $\pm$ 0.454}}^*$ \\
      \textbf{LNM}     &  $\tts{24.053 $\pm$ 0.609}$  &$\tts{1.364 $\pm$ 0.032}$ & $\tts{0.384 $\pm$ 0.005}$& $\tts{95.719 $\pm$ 0.779}$  \\      
       \midrule
      \textbf{Add-and-Thin-augm.}      & $\tts{ 24.929 $\pm$ 0.737}$ & $\ttt{   --  }$ & $\tts{  0.632 $\pm$ 0.018 }$  & $\tts{ 107.070 $\pm$ 0.590 }$ \\      
       \midrule
         \textbf{CDiff}    & \textbf{21.013  $\pm$  0.158}   & \textbf{{1.131 $\pm$ 0.017}} & \textbf{{0.351 $ \pm$ 0.004}}& \textbf{87.993 $\pm$ 0.178} \\ \midrule
     $\textbf{Taobao}$ & $\textbf{OTD}$ & $\textbf{RMSE}_{e}$ &$\textbf{RMSE}_{x^+}$ & \textbf{sMAPE} \\ \midrule
    \textbf{HYPRO}      & \textbf{44.336 $\pm$ 0.127}  &$\text{\underline{2.710 $\pm$ 0.021}}^*$ & $\text{\underline{{0.594 $\pm$ 0.030}}}^*$ & $\text{{134.922 $\pm$ 0.473}}^*$ \\
      \textbf{LNM}     & $\tts{45.757 $\pm$ 0.287}$  & $\tts{3.193 $\pm$ 0.043}$ & $\tts{0.575 $\pm$ 0.012}$ & $\ttt{{ 127.436 $\pm$ 0.606}}$ \\      
       \midrule
      \textbf{Add-and-Thin-augm.}  & $\tts{49.030 $\pm$ 0.943}$    & $\ttt{ -- }$  & $\tts{ 1.300 $\pm$ 0.032 }$  & $\tts{  144.597 $\pm$ 0.699 }$ \\      
       \midrule
         \textbf{CDiff}  &  \ttt{44.621 $\pm$ 0.139}  &\ttt{{2.653 $\pm$ 0.022}} & \ttt{{0.551 $\pm$ 0.002}} & \ttt{125.685 $\pm$ 0.151} \\ 

         \bottomrule
         $\textbf{StackOverflow}$& $\textbf{OTD}$ & $\textbf{RMSE}_{e}$&$\textbf{RMSE}_{x^+}$ & \textbf{sMAPE}\\ \midrule
         \textbf{HYPRO} & $\text{{42.359$\pm$0.170}}$ & \undl{1.140 $\pm$ 0.014} & $\text{1.554 $\pm$ 0.010}^*$ &  $\text{ 110.988 $\pm$  0.559 }^*$ \\
          \textbf{LNM} & $\tts{46.280 $\pm$ 0.892}$   & $\tts{  1.447 $\pm$ 0.057 }$  & $\tts{ 1.669 $\pm$ 0.005 }$  & $\tts{  115.122 $\pm$ 0.627 }$ \\
         \midrule
       \textbf{Add-and-Thin-augm.}      & $\tts{ 45.693 $\pm$ 0.368}$ & $\ttt{   --  }$ & $\tts{  1.620$\pm$ 0.090 }$  & $\tts{ 111.468 $\pm$ 0.702}$ \\
       \midrule
         \textbf{CDiff} & \textbf{41.245 $\pm$ 1.400 } & {{1.141 $\pm$ 0.007}} & \textbf{{1.199 $\pm$ 0.006}}&      \textbf{106.175$\pm$ 0.340} \\
         \bottomrule \\
         $\textbf{Retweet}$& $\textbf{OTD}$ & $\textbf{RMSE}_{e}$&$\textbf{RMSE}_{x^+}$ & \textbf{sMAPE} \\ \midrule
         \textbf{HYPRO} & $\text{61.031$\pm$0.092}^*$ &$\text{2.623 $\pm$ 0.036}^*$ & $\text{30.100 $\pm$ 0.413}^*$  & \undl{106.110$\pm$ 1.505}\\
          \textbf{LNM} &    $\tts{61.715 $\pm$ 0.152}$         & $\tts{2.776 $\pm$ 0.043 }$  & $\ttt{27.582 $\pm$ 0.191}$  & $\tts{106.711 $\pm$ 1.615 }$  \\
         \midrule
       \textbf{Add-and-Thin-augm.}      & $\tts{ 61.013 $\pm$ 0.190}$    & $\ttt{ -- }$  & $\tts{ 32.010 $\pm$ 0.046 }$  & $\tts{  116.895 $\pm$ 0.607 }$ \\
       \midrule
         \textbf{CDiff} & $\ttt{60.661 ± 0.101}$   & \textbf{{2.293 $\pm$ 0.034}} & \textbf{{27.101 $\pm$ 0.113} }  & {106.184 $\pm$ 1.121}\\
         \bottomrule 
    \end{tabular}}

    \label{tab:rebuttal whole table}
\end{table*}
}

{
\subsection{Next single event prediction}\label{sec:app_N1}
Figure \ref{fig:avg_rmse_e_smape_diff_horizon} in our paper illustrates how the sequence length for prediction affects the overall ranking of the baselines.  We include the specific results for $N=1$ in Table \ref{tab:single_event_table_top} format here to improve readability for Figure \ref{fig:avg_rmse_e_smape_diff_horizon}. 
\begin{table*}[h]
    \caption{$\textbf{RMSE}_{x^+}$, $\textbf{Accuracy}$, $\textbf{sMAPE}$ and $\textbf{Error Rate}$ for $N=1$ of real-world datasets reported in mean $\pm$ s.d. Since we only have one event, we can report the $\textbf{Error Rate}$  of our single event type prediction. Best are in bold, the next best is underlined. HYPRO and Dual-TPP with single event forecasting will become AttNHP and RMTPP. *indicates stat. significance w.r.t to the best method.}
    \vskip 0.1in
    \centering\resizebox{0.5\linewidth}{!}{\begin{tabular}{lccc} \toprule\\ 
      \textbf{Taxi}  & $\textbf{RMSE}_{x^+} \downarrow$ & $\textbf{Accuracy} \uparrow$ & $\textbf{sMAPE} \downarrow$ \\ \midrule

     \textbf{AttNHP} & $\textbf{0.321 $\pm$ 0.003}$ & $\ttt{0.905 $\pm$ 0.007}$ & $\textbf{85.132 $\pm$ 0.261}$ \\ 
      \textbf{RMTPP} &  $\ttt{\undl{0.335 $\pm$  0.006}}$  &$\ttt{0.907  $\pm$  0.010}$ & $\ttt{89.115 $\pm$ 0.753}$ \\ 
       \textbf{NHP}      & $\ttt{0.340 $\pm$ 0.007}$ & $\textbf{0.910 $\pm$ 0.007}$ & $\ttt{90.625 $\pm$ 0.608}$\\  
      \textbf{LNM}     &  $\tts{0.377 $\pm$ 0.009}$  &$\ttt{0.904 $\pm$ 0.007}$ & $\tts{90.032 $\pm$ 0.470}$\\ 
     
       \midrule
         \textbf{CDiff}    & $\ttt{0.337 $\pm$  0.009}$   & \ttt{\undl{0.909 $\pm$ 0.004}} & \ttt{\undl{87.124 $ \pm$ 0.608}}\\ 
         \bottomrule\\ 
        \textbf{Taobao} & $\textbf{RMSE}_{x^+} \downarrow$ & $\textbf{Accuracy} \uparrow$ & $\textbf{sMAPE} \downarrow$ \\ \midrule

     \textbf{AttNHP} &  $\ttt{\undl{0.527 $\pm$ 0.004}}$ & $\ttt{\undl{0.468 $\pm$ 0.011}}$ & $\ttt{129.133 $\pm$ 1.354}$\\ 
      \textbf{RMTPP} &  $\tts{{0.531 $\pm$ 0.007}}$  & $\ttt{\undl{0.468 $\pm$ 0.021}}$ & $\tts{131.432 $\pm$ 1.992}$\\ 
       \textbf{NHP}      &  $\tts{{0.531 $\pm$ 0.004}}$   & $\ttt{0.458 $\pm$ 0.009}$ & $\tts{133.693 $\pm$ 2.246}$\\  
      \textbf{LNM}     &  $\tts{0.532 $\pm$ 0.007}$  & $\tts{0.450 $\pm$ 0.007}$ & $\textbf{126.009 $\pm$ 1.482}$ \\ 
     
       \midrule
         \textbf{CDiff}    &  $\textbf{0.516 $\pm$ 0.009}$  &\textbf{{0.477 $\pm$ 0.003}} & \ttt{\undl{127.121 $\pm$ 1.356}}\\ 
         \bottomrule
    \end{tabular}}
    \label{tab:single_event_table_top}
\end{table*}
}

{
\subsection{Detailed Model details}
\label{appendix:detailed model description}
Here we explain the learnable functions in Eq.\ref{eq:dt_parameterization} and Eq.\ref{eq:type_parameterization}: 
$\phi_{\theta}(e_{t},x_{t},s,t)$ and $\epsilon_{\theta}(e_{t},x_{t-1},s,t)$ serve as denoising functions for the type diffusion process and the time diffusion process, respectively. 
\begin{itemize}
    \item The history embedding is derived from a history encoder that processes $\bs$, transforming it into an embedding with a dimension of $4 \times M$. This results in the overall sequence having dimensions of $R^{L \times (4 \times M)}$.

    \item For time values $t$ within the range of $0$ to $N_{step}$, where $N_{step}$ denotes the total number of diffusion steps, the time $t$ undergoes a transformation into a positional encoding as described in Equation 21. The positional encoding $\bt$ is then a vector in $\real^{M}$.
    \item $\bx_{t}$, $\in \real_+^{L}$, has its dimension expanded to $\real^{L\times M}$ for the function $\phi_{\theta}(e_{t},x_{t},s,t)$, $\real^{L\times (2\times M)}$ for $\epsilon_{\theta}(e_{t},x_{t-1},s,t)$ with the same transformation specified in Equation 21. 
    \item $\be_{t}$ denotes a sequence of one-hot vectors. Its dimension is augmented through a learnable event embedding matrix in $R^{M\times K}$, with the $k$-th column providing an $M$-dimensional embedding for the event type $k$. The resulting sequence embedding falls within $\real^{L\times M}$ for the function $\epsilon_{\theta}(e_{t},x_{t-1},s,t)$, $\real^{L\times (2\times M)}$ for $\phi_{\theta}(e_{t},x_{t},s,t)$
    \item To get the sequence order of the event tokens, an additional positional encoding specific to the order is computed. The event token order is converted into a positional encoding following Equation 21, resulting in dimensions of $\real^{L\times (4\times M)}$.
    \item By concatenating the embeddings for the diffusion time step, inter-arrival time, and event type, we obtain an embedding in $\real^{L\times (4\times M)}$. We incorporate the positional encoding 
    by summing the positional encoding to the concatenated embedding.
    \item This sequence of embeddings is then processed by a transformer block, facilitating cross-attention between the history embedding (in $R^{L \times (4\times M)}$) and the embedding of the forecasted event token sequences (in $\real^{L\times (4\times M)}$), yielding an output dimension of $4M$.
    \item Finally, a linear projection is applied to the inter-arrival time embedding to convert it into $\real$, and for the event type embedding, a linear projection converts it into $\real^K$, followed by a softmax function to get the logits of $\be_{t}$.

\end{itemize}
}

\subsection{Sampling Details}\label{sec:sampling}

In order to achieve a faster sampling time, we leverage the work of~\citet{song2020ddim}. We can re-express Eq.\ref{eq:dt_parameterization} as follows 
\begin{align}
\label{eq:dt_parameterization_faster_sampling}
    \bx_{t-1}&=\sqrt{\bar{\alpha}_{t-1}}(\frac{\bx_t - \sqrt{1-\bar{\alpha}_t}\epsilon_{\theta}(\bx_t, t, \be_t, \bs_c)}{\sqrt{\bar{\alpha}_t}})
     + \sqrt{1-\bar{\alpha}_{t-1}-\sigma_t^2} \cdot \epsilon_{\theta}(\bx_t, t, \be_t, \bs_c) 
     + \sigma_t \bz
\end{align}
Given a trained DDPM model, we can specify $\{\sigma_t \}^{\tau}_{t=1}$ and specify $\tau \subset \{1,2,..,T\}$ to accomplish the acceleration. In Eq.\ref{eq:dt_parameterization_faster_sampling}, if we set $\sigma_t = 0$ then we are performing DDIM (Denoising Diffusion Implicit Model) acceleration as in~\citep{song2020ddim}. For event type acceleration, we choose to directly jump steps, because for multinomial diffusion \citep{hoogeboom2021ArgmaxFlowsMultinomial}, instead of predicting noise, we predict $\be_0$. Therefore, our acceleration relies on decreasing the number of times we recalculate $\hat{\be}_0=\phi_{\theta}(\be_{t}, \bx_t, t,\bs_c)$. That is, given a sub-set $\tau \subset \{1,2,..,T\}$, we only recalculate $\hat{\be}_0$ $|\tau|$ times. In practice, we found it does not harm the prediction but it significantly accelerates the sampling due to $\phi_{\theta}(\cdot)$ requiring the majority of the computation effort.

\begin{figure}
    \centering
    \includegraphics[width=0.5\linewidth]{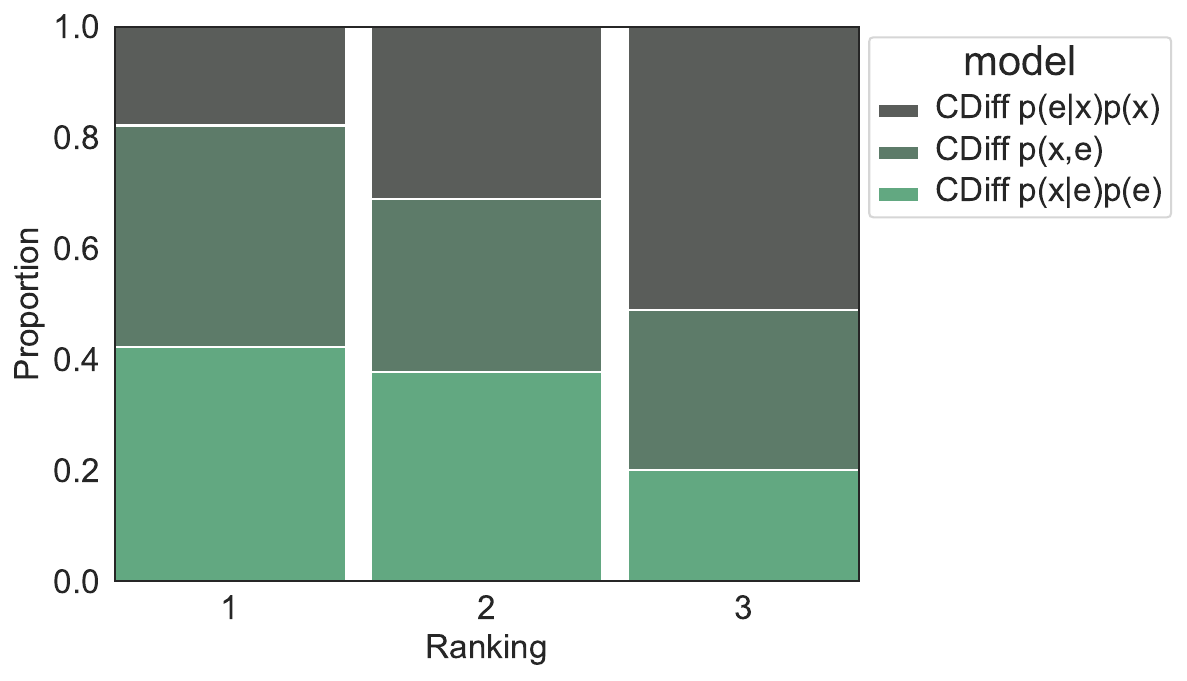}
    \caption{100\% Stacked column chart of ranks of different CDiff across the 5 datasets for all the metrics.}
    \label{fig:ablation_joint_vs_condition}
\end{figure}

\subsection{Comparison with $p(\bx, \be)$ and $p(\be|\bx)p(\bx)$}
\label{sec:model_e_t_order}
Mathematically, $p(\bx, \be) = p(\be | \bx)p(\be) = p(\bx | \be)p(\be)$, so there should not be any theoretical difference between sampling the event type and interarrival time jointly or sampling one first and then the other, conditioned on the first. We conducted an experiment to check that this was also observed in the practical implementation. 
Fig.\ref{fig:ablation_joint_vs_condition} shows that 
the order of sampling does not have a major effect, although there is a minor advantage to either jointly sampling from $p(\bx, \be)$ or sampling the event type first (i.e., from $p(\be | \bx)p(\be)$). This perhaps reflects that it is easier to learn the conditional inter-arrival time distributions, which may have slightly simpler structure.

\subsection{Positional Encoding for CDiff}
\label{sec:positional_enc_diff}
We use the transformer architecture as a denoising tool for reversing the diffusion processes. Therefore, we encode the position of both the diffusion step and the event token's order. 

It is important that our choice of encoding can differentiate between these two different types of position information. To achieve this, we use as input $(i + y_{N})$, where $i$ is the order of the event token in the noisy event sequence, and $y_{N}$ is the last timestamp of the historical event sequence. 

into Eq. \ref{eq:temporal_enc} (shown also below) for the order of the predicted sequence. This approach distinctly differentiates the positional information of the predicted event sequence from the diffusion time step's positional encoding. The positional encoding is then:
\begin{align}
[\mathbf{m}(y_j, D)]_i=\left\{
\begin{aligned}
\cos(y_j/10000^{\frac{i-1}{D}}) \quad \text{if $i$ is odd\,,}\\
\sin(y_j/10000^{\frac{i}{D}}) \quad \text{if $i$ is even.}
\end{aligned}
\right.
\label{eq:temporal_enc_2}
\end{align}

\subsection{More Diffusion Visualization}
\begin{figure}
    \centering
    \begin{minipage}[b]{0.45\linewidth}
    \centering
\begin{tikzpicture}
   \node[anchor=south west,inner sep=0,yshift=-0.4cm] (image1) at (0,0) {\includegraphics[width=\columnwidth]{
   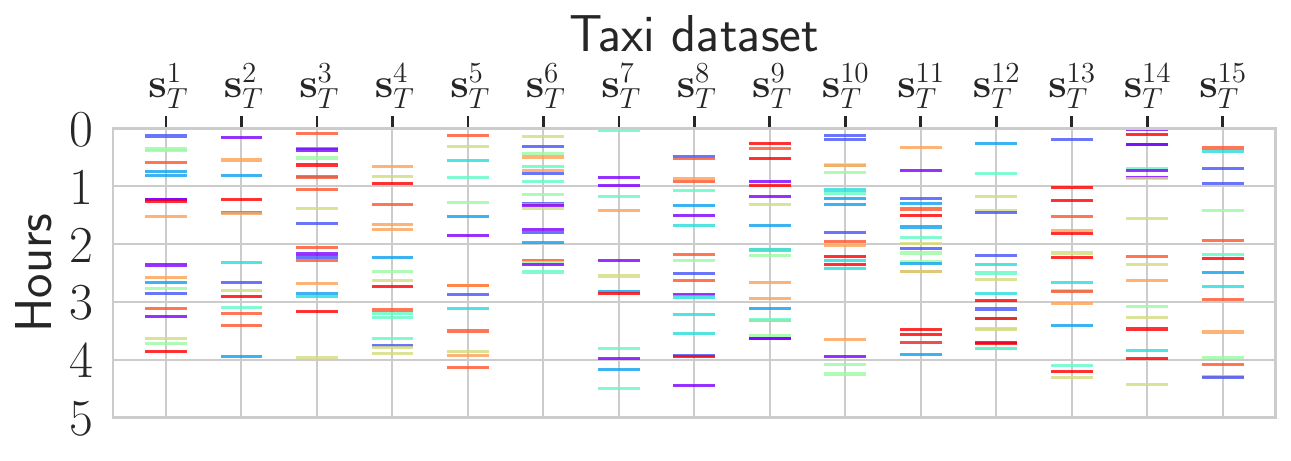}};
   
   \node[anchor=north west,inner sep=0,yshift=-0.4cm] (image2) at (image1.south west) {\includegraphics[width=\columnwidth]{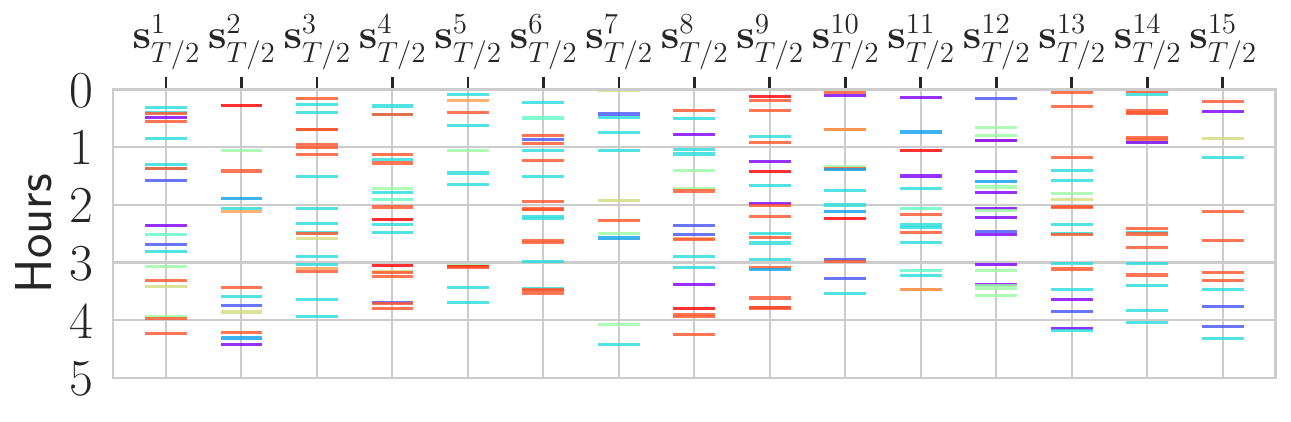}};

   \node[anchor=north west,inner sep=0,yshift=-0.4cm] (image3) at (image2.south west) {\includegraphics[width=\columnwidth]{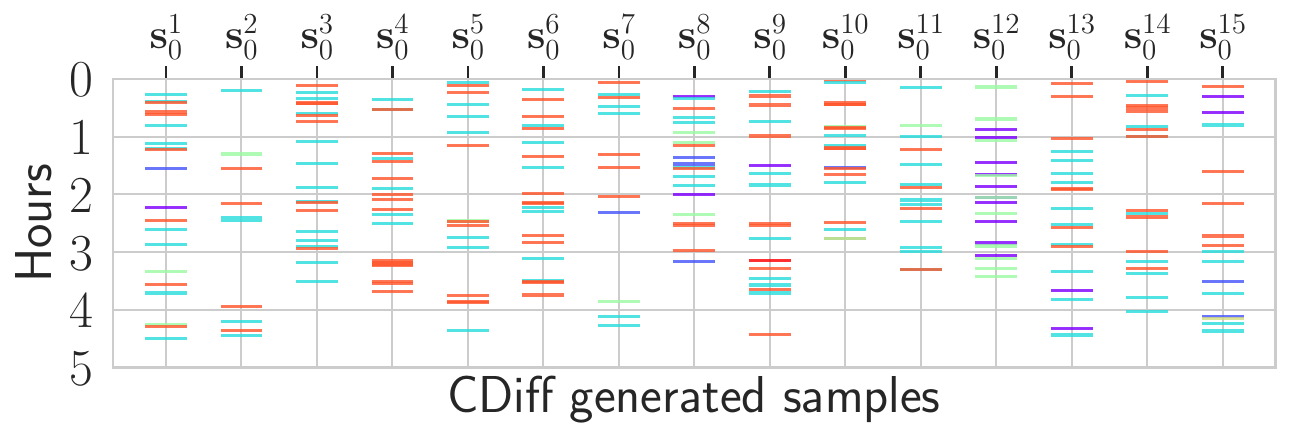}};

   \node[anchor=north west,inner sep=0,yshift=-0.1cm] (image4) at (image3.south west) {\includegraphics[width=\columnwidth]{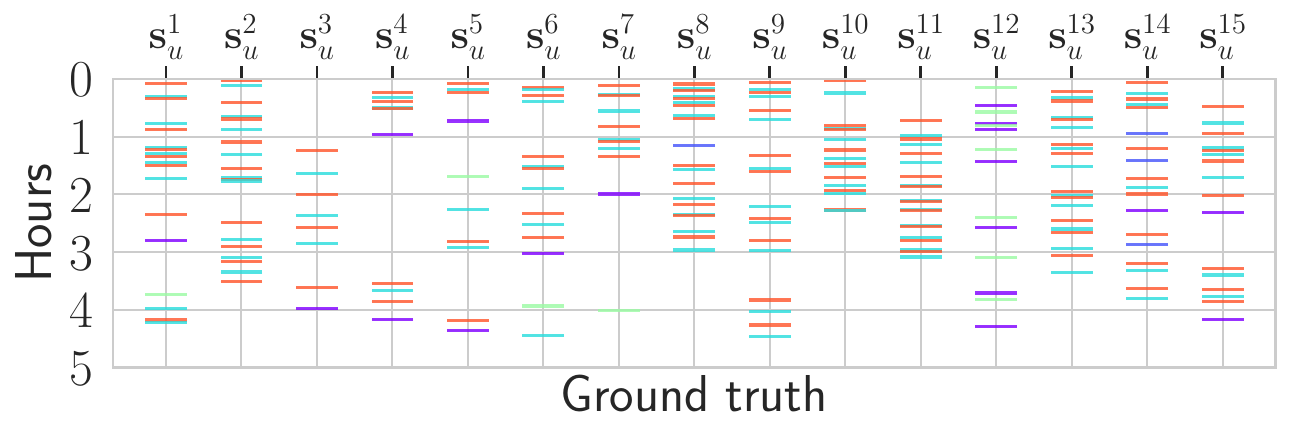}};

   \draw[-stealth, ultra thick]  (image1.south) to[out=-90,in=90] (image2.north);
   \draw[-stealth, ultra thick] (image2.south) to[out=-90,in=90]  (image3.north);
\end{tikzpicture}
    \end{minipage}
    \begin{minipage}[b]{0.45\linewidth}
    \centering
    \begin{tikzpicture}
   \node[anchor=south west,inner sep=0,yshift=-0.4cm] (image1) at (0,0) {\includegraphics[width=\columnwidth]{
   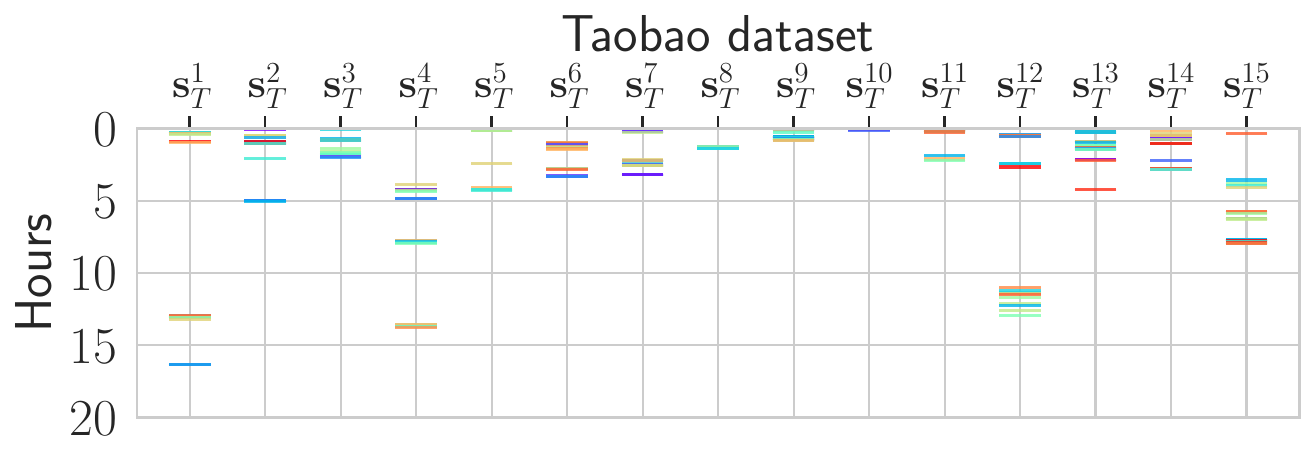}};
   
   \node[anchor=north west,inner sep=0,yshift=-0.4cm] (image2) at (image1.south west) {\includegraphics[width=\columnwidth]{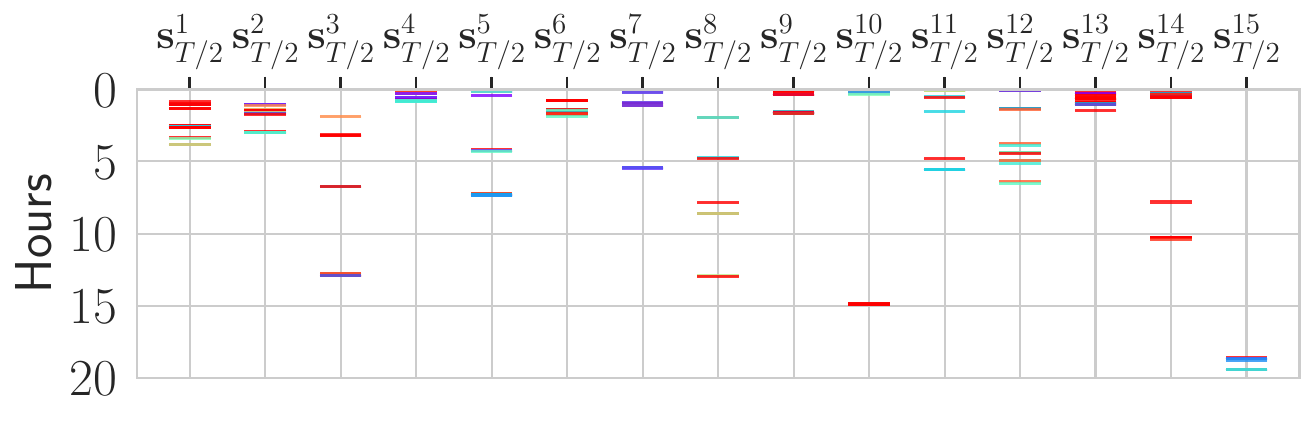}};

   \node[anchor=north west,inner sep=0,yshift=-0.4cm] (image3) at (image2.south west) {\includegraphics[width=\columnwidth]{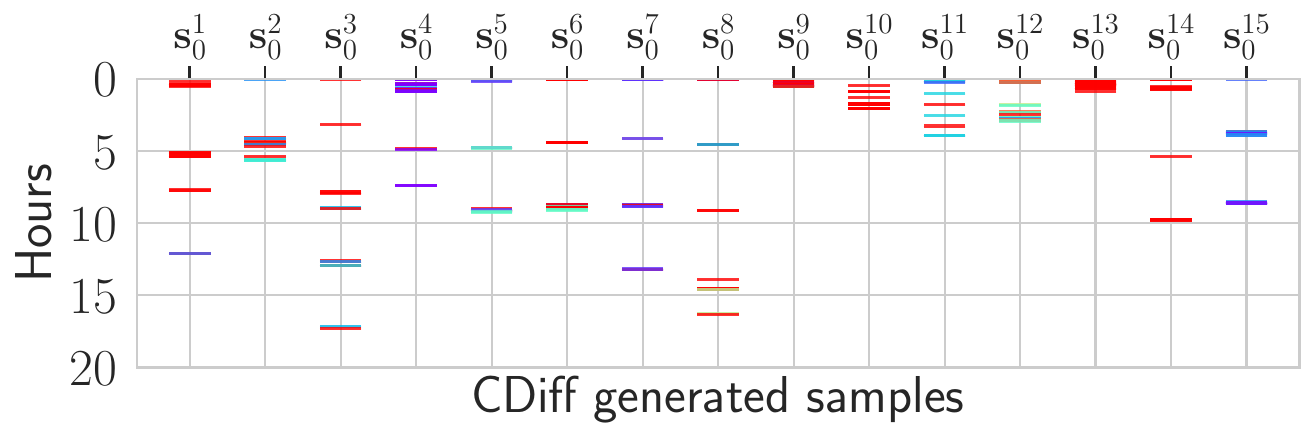}};

   \node[anchor=north west,inner sep=0,yshift=-0.1cm] (image4) at (image3.south west) {\includegraphics[width=\columnwidth]{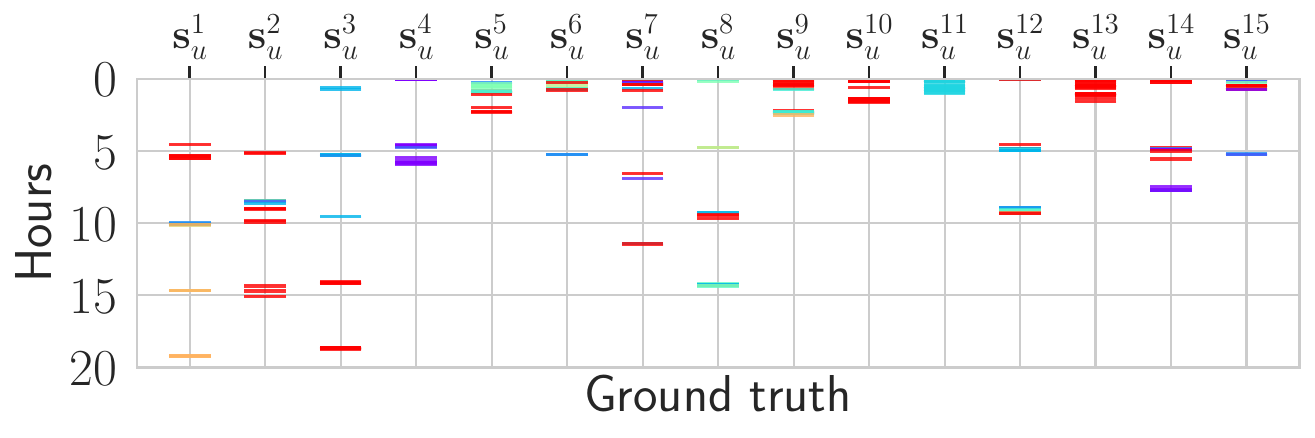}};

   \draw[-stealth, ultra thick]  (image1.south) to[out=-90,in=90] (image2.north);
   \draw[-stealth, ultra thick] (image2.south) to[out=-90,in=90]  (image3.north);
\end{tikzpicture}
    \end{minipage}
    \caption{Visualization of the cross-diffusion generating process for 15 examples sequences of the Taxi dataset (left) and the Taobao dataset (right). The colors indicates the different categories. We start by generating noisy sequences ($t=T$). Once we reach the end of the denoising process ($t=0$), we have recovered sequences similar to the ground truth sequences. We cut the sequence based on the time range so that every sequence can be aligned.}
    \label{fig:sample_chain_appendix}
\end{figure}
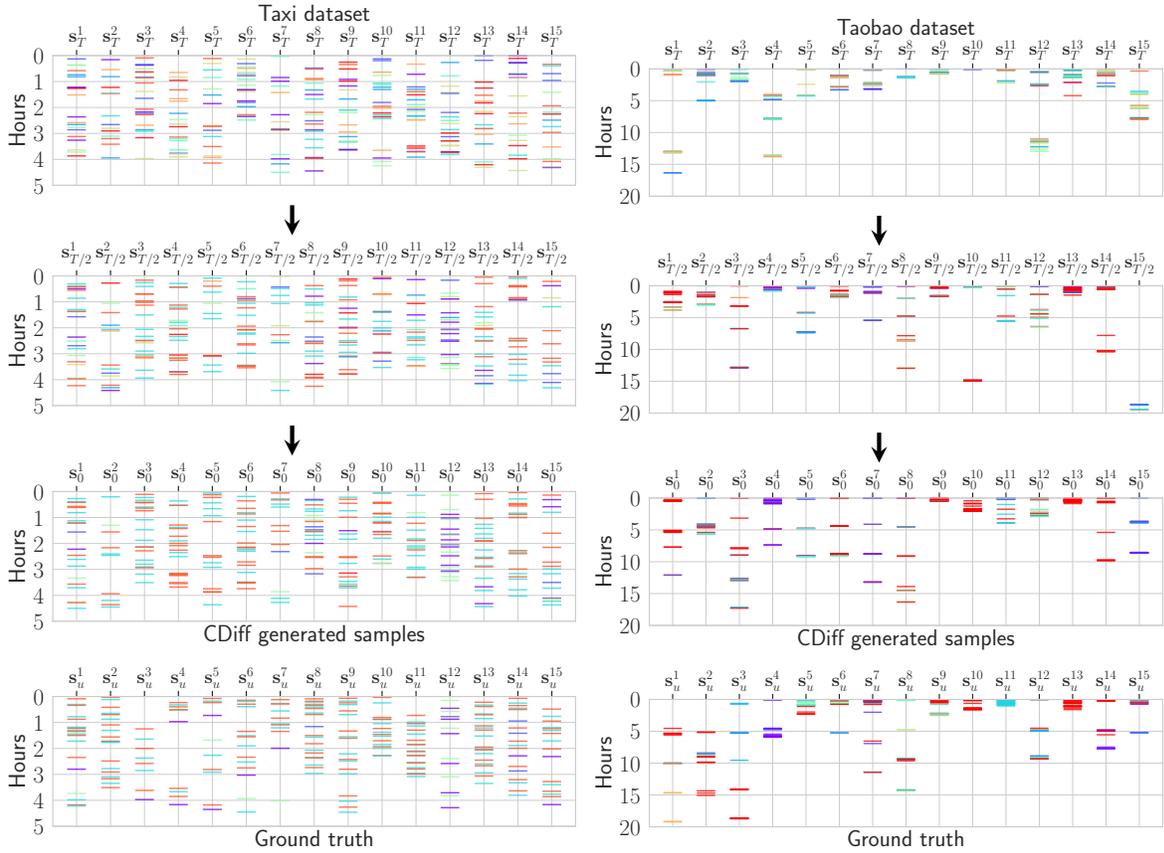

Figure \ref{fig:sample_chain_appendix} shows the reverse process of CDiff for Taxi dataset (on the left) and Taobao dataset (on the right). Upon inspection, it is evident that the recovered sequences bear a strong resemblance to their respective ground truth sequences, both in terms of inter-arrival time patterns and event classifications.

In the Taxi dataset, the original sequences prominently feature events colored in Cyan and Orange. This indicates a high frequency of these two event categories, a pattern which is consistently replicated in the sequences derived from CDiff.

Conversely, for the Taobao dataset, the ground truth predominantly showcases shorter inter-arrival times, signifying closely clustered events. However, there are also occasional extended inter-arrival times introducing gaps in the sequences. Notably, this dichotomy is accurately reflected in the reconstructed sequences.

\subsection{Tables of results with different evaluation metrics for different horizon}
\label{sec:complete_result}

Tables \ref{tab:complete_table_n20}, \ref{tab:complete_table_n10} and \ref{tab:complete_table_n5} show the results of all metrics across all models for all datasets with different prediction horizons. We test for significance using a paired Wilcoxon signed-rank test at the 5\% significance level.

\begin{table*}[h]
\small
\caption{Results for all metrics across 7 different datasets for \textbf{$N=20$ events forecasting} and \textbf{long interval forecasting}, bold case indicates the best, under line indicates the second best, * indicates stats. significance w.r.t. the method with the lowest value.}
\centering\resizebox{\linewidth}{!}{\begin{tabular}{lccccc|cccc} 
\\
&\multicolumn{9}{c}{\textbf{Synthetic dataset}}\\ \midrule
&\multicolumn{5}{c|}{\textbf{$N=20$ events forecasting}}     &\multicolumn{4}{c}{\textbf{Interval forecasting $t'$ long}}\\ 
& $\textbf{OTD}$ & $\textbf{RMSE}_{e}$ &$\textbf{RMSE}_{x^+}$ & \textbf{MAPE} &\textbf{sMAPE} & $\textbf{OTD}$ & $\textbf{RMSE}_{e}$&$\textbf{RMSE}_{|\bs^+|}$ & $\textbf{MAE}_{|\bs^+|}$\\ \midrule
\textbf{HYPRO} & $\text{\underline{20.609 $\pm$ 0.328}}$ & $\text{2.464 $\pm$ 0.039}$ & $\tts{\underline{0.104 $\pm$ 0.002}}$ & $\ttt{717.417 $\pm$ 56.443} $& $\tts{100.535 $\pm$ 0.084}$ & $\tts{\underline{ 20.224 $\pm$ 0.236}}$ & $\ttt{\underline{2.409 $\pm$ 0.082}}$ & \undl{1.608 $\pm$ 0.103} & \textbf{0.573 $\pm$ 0.049}\\
\textbf{Dual-TPP} & $\text{22.117 $\pm$ 0.368}^*$ & $\tts{2.506 $\pm$ 0.044}$ & $\tts{0.108 $\pm$ 0.001}$ & $\tts{724.681 $\pm$ 28.097}$ & $\tts{100.857 $\pm$ 0.624}$ & $\tts{21.521 $\pm$ 0.375}$ & $\tts{2.511 $\pm$ 0.050}$ & $\tts{2.297 $\pm$ 0.117}$ & $\tts{0.952 $\pm$ 0.057}$ \\ 
\textbf{Attnhp} & $\tts{21.843 $\pm$ 0.316}$ & $\tts{2.509 $\pm$ 0.051}$ & \undl{$\tts{0.104 $\pm$ 0.004}$} & \underline{682.086 $\pm$ 63.199} & $\tts{101.117 $\pm$ 0.295}$ & $\tts{21.153 $\pm$ 0.206}$ & $\tts{2.509 $\pm $ 0.048}$  & $\tts{2.806 $\pm$ 0.073}$ & $\tts{0.809 $\pm$ 0.033}$ \\ 
\textbf{NHP} & $\tts{21.541 $\pm$ 0.203}$ & $\tts{\underline{2.462 $\pm$ 0.018}}$ & $\tts{0.109 $\pm$ 0.001}$ & $\tts{786.866 $\pm$ 31.782}$ & $\tts{\underline{99.622 $\pm$ 0.426}} $& $\tts{20.541 $\pm$ 0.203}$ & $\tts{2.462 $\pm$ 0.021}$ & \tbf{1.411 $\pm$ 0.048 } & $\tts{\underline{0.588 $\pm$ 0.013}} $\\  
\textbf{LogNM} & $\tts{22.082 $\pm$ 0.225}$ & $\tts{2.932 $\pm$ 0.028}$ & $\tts{0.109 $\pm$ 0.005}$ & $\tts{815.764 $\pm$ 32.480}$ & $\tts{102.207 $\pm$ 0.472}$ & $\tts{21.713 $\pm$ 0.198}$ & $\tts{2.914 $\pm$ 0.019}$ & $\tts{1.982 $\pm$ 0.078}$ & $\ttt{0.741 $\pm$ 0.054}$\\  
\textbf{TCDDM} & $\ttt{21.270 $\pm$ 0.528}$ & $\tts{2.796 $\pm$ 0.027}$ & $\ttt{0.102 $\pm$ 0.002}$ & $\ttt{700.630 $\pm$ 40.377}$ & $\tts{100.237 $\pm$ 0.275}$ & $\ttt{20.912 $\pm$ 0.310}$ & $\ttt{2.735 $\pm$ 0.026}$ & $\tts{1.959 $\pm$ 0.03}$ & $\tts{0.816 $\pm$ 0.011}$\\
\textbf{Homog. Poisson} & $\tts{22.595 $\pm$ 0.198}$ & $\tts{2.946 $\pm$ 0.023}$ & $\tts{0.129 $\pm$ 0.001}$ & $\tts{1025.234 $\pm$ 139.141}$ & $\tts{101.973 $\pm$ 0.380}$ & $\tts{22.179 $\pm$ 0.298}$ & $\tts{2.918 $\pm$ 0.037}$ & $\tts{2.903 $\pm$ 0.065}$ & $\tts{0.991 $\pm$ 0.067}$\\
\midrule
\textbf{CDiff} & \textbf{19.788 $\pm$ 0.343}   & \textbf{{2.375 $\pm$ 0.021}} & \textbf{{0.098 $ \pm$ 0.02}}& \textbf{668.287 $\pm$ 51.873} &\textbf{98.933 $\pm$ 0.573} & \textbf{19.674 $\pm$ 0.125} & \textbf{2.370 $\pm$ 0.061} & {1.932 $\pm$ 0.094} & $\tts{0.812 $\pm$ 0.051}$ \\ \bottomrule \\

&\multicolumn{9}{c}{\textbf{Taxi dataset}}\\ \midrule
&\multicolumn{5}{c|}{\textbf{$N=20$ events forecasting}}     &\multicolumn{4}{c}{\textbf{Interval forecasting $t'$ long}}\\ 
& $\textbf{OTD}$ & $\textbf{RMSE}_{e}$ &$\textbf{RMSE}_{x^+}$ & \textbf{MAPE} &\textbf{sMAPE} & $\textbf{OTD}$ & $\textbf{RMSE}_{e}$&$\textbf{RMSE}_{|\bs^+|}$ & $\textbf{MAE}_{|\bs^+|}$\\ \midrule
\textbf{HYPRO} & \underline{21.653 $\pm$ 0.163} & $\text{\underline{1.231 $\pm$ 0.015}}^*$ & $\text{\underline{0.372 $\pm$ 0.004}}^*$ & $\ttt{\underline{252.761 $\pm$ 6.827}}$ &$\text{\underline{93.803 $\pm$ 0.454}}^*$ & \underline{19.632 $\pm$ 0.179 } & \underline{1.550 $\pm$ 0.026} & $\tts{\underline{4.326 $\pm$ 0.063}}$ & $\tts{\underline{2.781 $\pm$ 0.088}}$\\
\textbf{Dual-TPP}      & ${24.483 \pm 0.383}^*$ & $\text{1.353 $\pm$ 0.037}^*$ & $\text{0.402 $\pm$ 0.006}^*$ & $\tts{285.590 $\pm$ 8.088}$ & $\text{95.211 $\pm$ 0.187}^*$ & $\tts{20.952 $\pm$ 0.278}$ & $\tts{1.627 $\pm$ 0.033}$ & $\tts{4.995 $\pm$ 0.150}$ & $\tts{3.795 $\pm$ 0.107}$\\ 
\textbf{Attnhp}     &  ${24.762 \pm 0.217}^*$  &$\text{1.276  $\pm $ 0.015}^*$ & ${0.430 \pm 0.003}^*$  & $\tts{286.869 $\pm$ 9.973}$ & $\text{97.388 $\pm$  0.381}^*$ & $\tts{20.588 $\pm$ 0.208}$ & {1.590 $\pm$ 0.024} & $\tts{4.915 $\pm$ 0.116}$ & $\tts{3.509 $\pm$ 0.112}$\\ 
\textbf{NHP}      & $\text{25.114 $\pm$ 0.268}^*$ &$ {1.297 \pm 0.019}^*$ & ${0.399 \pm 0.040}^*$ & $\tts{281.306 $\pm$ 8.271}$ & $\text{96.459 $\pm$ 0.521}^*$ & $ \tts{21.134 $\pm$ 0.148}$ & $\tts{1.632 $\pm$ 0.030}$ & $\tts{4.883 $\pm$ 0.119}$ & $\tts{3.526 $\pm$ 0.135}$ \\  
\textbf{LogNM} & $\tts{24.053 $\pm$ 0.609}$ & $\tts{1.364 $\pm$ 0.032}$ & $\tts{0.384 $\pm$ 0.005}$ & $\tts{282.173 $\pm$ 4.532}$ & $\tts{95.719 $\pm$ 0.779}$ & $\ttt{20.422 $\pm$ 0.224}$ & $\ttt{1.603 $\pm$ 0.033}$ & $\tts{ 5.072 $\pm$ 0.066}$ & $\tts{3.796 $\pm$ 0.116}$\\
\textbf{TCDDM} & $\ttt{22.148 $\pm$ 0.529}$ & $\tts{1.309 $\pm$ 0.030}$ & $\ttt{0.382 $\pm$ 0.019}$ & $\ttt{259.944 $\pm$ 7.220}$ & $\ttt{90.596 $\pm$ 0.574}$ & $\ttt{20.191 $\pm$ 0.271}$ & $\tts{1.589 $\pm$ 0.064}$ & $\tts{ 4.530 $\pm$ 0.118}$ & $\ttt{2.953 $\pm$ 0.237}$\\
\textbf{Homog. Poisson} & $\tts{25.104 $\pm$ 0.083}$ & $\tts{1.391 $\pm$ 0.032}$ & $\tts{0.407 $\pm$ 0.002}$ & $\tts{280.065 $\pm$ 7.541}$ & $\tts{97.689 $\pm$ 0.613}$ & $\tts{21.880 $\pm$ 0.175}$ & $\tts{1.685 $\pm$ 0.019}$ & $\tts{5.117 $\pm$ 0.151}$ & $\tts{3.849 $\pm$ 0.105}$\\

\midrule
\textbf{CDiff}    & \textbf{21.013  $\pm$  0.158}   & \textbf{{1.131 $\pm$ 0.017}} & \textbf{{0.351 $ \pm$ 0.004}}& \textbf{243.2 $\pm$ 7.725} &\textbf{87.993 $\pm$ 0.178} & \textbf{19.028 $\pm$ 0.224} & \textbf{1.329 $\pm$ 0.029} & \textbf{3.690 $\pm$ 0.097} & \textbf{2.593 $\pm$ 0.124} \\ \bottomrule

\\
&\multicolumn{9}{c}{\textbf{Taobao dataset}}\\ \midrule
&\multicolumn{5}{c|}{\textbf{$N=20$ events forecasting}}     &\multicolumn{4}{c}{\textbf{Interval forecasting $t'$ long}}\\ 
& $\textbf{OTD}$ & $\textbf{RMSE}_{e}$ &$\textbf{RMSE}_{x^+}$ & \textbf{MAPE} &\textbf{sMAPE} & $\textbf{OTD}$ & $\textbf{RMSE}_{e}$&$\textbf{RMSE}_{|\bs^+|}$ & $\textbf{MAE}_{|\bs^+|}$\\ \midrule
\textbf{HYPRO} & \textbf{44.336 $\pm$0.127} & $\tts{\underline{2.710 $\pm$0.021}}$ & $\tts{\underline{0.594 $\pm$0.030}}$ & $\ttt{\underline{6397.66 $\pm$ 154.977}}$ & $\tts{134.922 $\pm$ 0.473}$ & $\tts{42.525 $\pm$ 0.151 }$ & \textbf{2.810 $\pm$ 0.028} & \textbf{4.022 $\pm$ 0.067} & $\tts{3.019 $\pm$ 0.017}$  \\
\textbf{Dual-TPP} & $\tts{47.324 $\pm$ 0.541}$ & $\tts{3.237 $\pm$ 0.049}$ & $\tts{0.871 $\pm$0.014}$ & $\tts{8325.564 $\pm$ 245.765}$ & $\tts{141.687 $\pm$ 0.431}$ & $\tts{\underline{38.530 $\pm$ 0.263}}$ & $\tts{4.439 $\pm$ 0.019}$ & $\tts{5.893 $\pm$ 0.088}$ & $\tts{3.832 $\pm$ 0.016}$ \\ 
\textbf{Attnhp} & $\tts{45.555 $\pm$ 0.345}$ & {2.737 $\pm$ 0.021} & $\tts{0.708 $\pm$ 0.011}$ & \textbf{6250.83 $\pm$ 265.440} & $\tts{\underline{134.582 $\pm$ 0.920}}$ & $\tts{43.624 $\pm$ 0.282}$ & {2.855 $\pm$ 0.020} & \underline{4.097 $\pm$ 0.016} & \textbf{2.892 $\pm$ 0.024} \\ 
\textbf{NHP} & $\tts{48.131 $\pm$ 0.297}$ & $\tts{3.355 $\pm$ 0.030}$ & $\tts{0.837 $\pm$ 0.009}$ & $\tts{7909.437 $\pm$ 149.274}$ & $\tts{137.644 $\pm$ 0.764}$ & \textbf{38.204 $\pm$ 0.302} & $\tts{3.515 $\pm$ 0.028}$ & $\tts{5.41 $\pm$ 0.081}$ & $\tts{3.998 $\pm$ 0.027}$ \\  
\textbf{LogNM} & $\tts{45.757 $\pm$ 0.287}$ & $\tts{3.193 $\pm$ 0.043}$ & $\tts{0.575 $\pm$ 0.012}$ & $\ttt{6558.437 $\pm$ 170.430}$ & $\ttt{127.436 $\pm$ 0.606}$ & $\ttt{39.769 $\pm$ 0.615}$ & $\ttt{3.085 $\pm$ 0.076}$ & $\tts{4.914 $\pm$ 0.137}$ & $\tts{3.814 $\pm$ 0.096}$\\
\textbf{TCDDM} & $\tts{45.563 $\pm$ 0.889}$ & $\ttt{2.850 $\pm$ 0.058}$ & $\ttt{0.569 $\pm$ 0.015}$ & $\ttt{6843.217 $\pm$ 278.296}$ & $\ttt{126.512 $\pm$ 0.491}$ & $\tts{42.441 $\pm$ 0.434}$ & $\ttt{2.940 $\pm$ 0.094}$ & $\ttt{4.231 $\pm$ 0.158}$ & $\ttt{2.883 $\pm$ 0.057}$\\
\textbf{Homog. Poisson} & $\tts{52.990 $\pm$ 0.234}$ & $\tts{3.288 $\pm$ 0.022}$ & $\tts{0.906 $\pm$ 0.012}$ & $\tts{35474.601 $\pm$ 3495.078}$ & $\tts{151.689 $\pm$ 0.615}$ & $\tts{41.476 $\pm$ 0.811}$ & $\tts{3.519 $\pm$ 0.036}$ & $\tts{6.567 $\pm$ 0.083}$ & $\tts{4.731 $\pm$ 0.075}$\\
\midrule
\textbf{CDiff} & \underline{44.621 $\pm$ 0.139} & \textbf{2.653 $\pm$ 0.011} & \textbf{0.551 $\pm$ 0.013} & $\ttt{6850.359 $\pm$ 165.400}$ & \textbf{125.685 $\pm$ 0.151} & $\tts{40.783 $\pm$ 0.059}$ & \underline{2.831 $\pm$ 0.009} & {4.103 $\pm$ 0.034} & \underline{2.947 $\pm$0.019} \\ \bottomrule

\\
&\multicolumn{9}{c}{\textbf{Stackoverflow dataset}}\\ \midrule
&\multicolumn{5}{c|}{\textbf{$N=20$ events forecasting}}     &\multicolumn{4}{c}{\textbf{Interval forecasting $t'$ long}}\\ 
& $\textbf{OTD}$ & $\textbf{RMSE}_{e}$ &$\textbf{RMSE}_{x^+}$ & \textbf{MAPE} &\textbf{sMAPE} & $\textbf{OTD}$ & $\textbf{RMSE}_{e}$&$\textbf{RMSE}_{|\bs^+|}$ & $\textbf{MAE}_{|\bs^+|}$\\ \midrule
\textbf{HYPRO} & {42.359 $\pm$ 0.170} & \underline{1.140 $\pm$ 0.014} & $\tts{1.554 $\pm$ 0.010}$ & $\tts{2013.055 $\pm$ 160.862}$ & $\tts{110.988 $\pm$ 0.559}$ & \underline{38.460 $\pm$ 0.204} & \textbf{1.294 $\pm$ 0.016} & $\tts{{2.672 $\pm$ 0.019}}$ & {1.496 $\pm$ 0.017} \\
\textbf{Dual-TPP} & \underline{41.752 $\pm$ 0.200} & \textbf{1.134 $\pm$ 0.019} & $\tts{1.514 $\pm$ 0.017}$ & $\tts{1729.83 $\pm$ 67.928}$ & $\tts{117.582 $\pm$ 0.420}$ & {38.474 $\pm$ 0.274} & \underline{1.364 $\pm$ 0.019} & $\tts{3.332 $\pm$ 0.088}$ & $\tts{1.753 $\pm$ 0.036}$ \\ 
\textbf{Attnhp} & $\tts{42.591 $\pm$ 0.408}$ & {1.145 $\pm$ 0.011} & {1.340 $\pm$ 0.006} & \tbf{1519.740 $\pm$ 52.216} & \ttt{108.542 $\pm$0.531} & $\tts{39.76 $\pm$ 0.373}$ & $\tts{1.385 $\pm$ 0.014}$ & $\tts{3.424 $\pm$ 0.023}$ & $\tts{1.813 $\pm$ 0.014}$ \\ 
\textbf{NHP} & $\tts{43.791 $\pm$ 0.147}$ & $\tts{1.244 $\pm$ 0.030}$ & $\tts{1.487 $\pm$ 0.004}$ & $\tts{1693.977 $\pm$ 113.300}$ & $\tts{116.952 $\pm$ 0.404}$ & $\tts{40.453 $\pm$ 0.188}$ & $\tts{1.447 $\pm$ 0.012}$ & $\tts{3.552 $\pm$ 0.051}$ & $\tts{1.793 $\pm$ 0.057}$ \\  
\textbf{LogNM} & $\tts{46.280 $\pm$ 0.892}$ & $\tts{1.447 $\pm$ 0.057}$ & $\tts{1.669 $\pm$ 0.005}$ & $\ttt{2133.278 $\pm$ 163.516}$ & $\tts{115.122 $\pm$ 0.627}$ & $\tts{42.594 $\pm$ 0.148}$ & $\ttt{1.507 $\pm$ 0.027}$ & $\tts{3.714 $\pm$ 0.078}$ & $\tts{1.864 $\pm$ 0.076}$\\
\textbf{TCDDM} & $\ttt{42.128 $\pm$ 0.591}$ & $\tts{1.467 $\pm$ 0.014}$ & \undl{1.315 $\pm$ 0.004} & $\ttt{1762.121 $\pm$ 64.437}$ & \undl{$\ttt{107.659 $\pm$ 0.934}$} & $\ttt{38.697 $\pm$ 0.718}$ & $\ttt{1.444 $\pm$ 0.019}$ & \undl{{2.623 $\pm$ 0.044}} & \undl{1.428 $\pm$ 0.070}\\
\textbf{Homog. Poisson} & $\tts{45.923 $\pm$ 0.286}$ & $\ttt{1.374 $\pm$ 0.022}$ & $\tts{1.359 $\pm$ 0.012}$ & $\tts{2762.4786 $\pm$ 196.091}$ & $\tts{116.447 $\pm$ 0.418}$ & $\tts{43.288 $\pm$ 0.503}$ & $\tts{1.539 $\pm$ 0.016}$ & $\tts{3.459 $\pm$ 0.039}$ & $\tts{1.778 $\pm$ 0.051}$\\
\midrule
\textbf{CDiff} & \textbf{41.245 $\pm$ 1.400} & {1.141 $\pm$ 0.007} & \textbf{1.199 $\pm$ 0.006} & \undl{1667.884 $\pm$ 32.220} & \textbf{106.175 $\pm$ 0.340} & \textbf{37.659 $\pm$ 0.334} & $\tts{1.421 $\pm$ 0.015}$ & \textbf{1.726 $\pm$ 0.043} & \textbf{1.239 $\pm$ 0.029}  \\ \bottomrule
\\

&\multicolumn{9}{c}{\textbf{Retweet dataset}}\\ \midrule
&\multicolumn{5}{c|}{\textbf{$N=20$ events forecasting}}     &\multicolumn{4}{c}{\textbf{Interval forecasting $t'$ long}}\\ 
& $\textbf{OTD}$ & $\textbf{RMSE}_{e}$ &$\textbf{RMSE}_{x^+}$ & \textbf{MAPE} &\textbf{sMAPE} & $\textbf{OTD}$ & $\textbf{RMSE}_{e}$&$\textbf{RMSE}_{|\bs^+|}$ & $\textbf{MAE}_{|\bs^+|}$\\ \midrule
\textbf{HYPRO} & $\tts{61.031 $\pm$ 0.092}$ & $\tts{2.623 $\pm$ 0.036}$ & $\tts{30.100 $\pm$ 0.413}$ & $\tts{19686.811 $\pm$ 966.339}$ & \undl{106.110 $\pm$ 1.505} & \underline{59.292 $\pm$ 0.197} & {3.011 $\pm$ 0.029} & $\ttt{3.109 $\pm$ 0.092}$ & {1.858 $\pm$ 0.067} \\
\textbf{Dual-TPP} & $\tts{61.095 $\pm$ 0.101}$ & $\tts{2.679 $\pm$ 0.026}$ & {28.914 $\pm$ 0.300} & $\tts{17619.400 $\pm$ 1003.001}$ & {106.900 $\pm$ 1.293} & \textbf{59.164 $\pm$ 0.069} & $\tts{2.981 $\pm$ 0.041}$ & $\tts{{2.548 $\pm$ 0.133}}$ & $\tts{1.608 $\pm$ 0.028}$ \\ 
\textbf{Attnhp} & \textbf{60.634 $\pm$ 0.097} & \underline{2.561 $\pm$ 0.054} & $\tts{28.812 $\pm$ 0.272}$ & \textbf{15396.198 $\pm$ 1058.618} & $\tts{107.234 $\pm$ 1.293}$ & {59.302 $\pm$ 0.160} & {2.832 $\pm$ 0.057} & {2.736 $\pm$ 0.119} & \ttt{1.554 $\pm$ 0.084} \\ 
\textbf{NHP} &  {60.953 $\pm$ 0.079} & $\tts{2.651 $\pm$ 0.045}$ & \underline{27.130 $\pm$ 0.224} & \underline{15824.614 $\pm$ 1039.258} & $\tts{107.075 $\pm$ 1.398}$ & {59.395 $\pm$ 0.098} & \ttt{2.780 $\pm$ 0.046} & $\tts{2.649 $\pm$ 0.104}$ & $\tts{1.650 $\pm$ 0.044}$ \\  
\textbf{LogNM} & $\tts{61.715 $\pm$ 0.152}$ & $\tts{2.776 $\pm$ 0.043}$ & $\ttt{27.582 $\pm$ 0.191}$ & $\ttt{17914.114 $\pm$ 919.022}$ & $\tts{106.711 $\pm$ 1.615}$ & $\ttt{59.223 $\pm$ 0.247}$ & $\ttt{2.815 $\pm$ 0.095}$ & $\ttt{2.873 $\pm$ 0.118}$ & $\tts{1.847 $\pm$ 0.095}$\\ 
\textbf{TCDDM} & $\ttt{60.501 $\pm$ 0.087}$ & $\ttt{2.387 $\pm$ 0.050}$ & $\ttt{27.303 $\pm$ 0.152}$ & $\ttt{16070.5290 $\pm$ 540.227}$  & $\textbf{106.048 $\pm$ 0.610}$ & $\ttt{59.934 $\pm$ 0.122}$ & \undl{2.762 $\pm$ 0.189} & $\textbf{2.131 $\pm$ 0.090}$ & \undl{1.129 $\pm$ 0.055}\\
\textbf{Homog. Poisson} & $\tts{61.224 $\pm$ 0.135}$ & $\tts{3.179 $\pm$ 0.066}$ & $\ttt{35.125 $\pm$ 0.083}$ & $\tts{16800.047 $\pm$ 1793.164}$ & $\tts{117.581 $\pm$ 0.500}$ & $\ttt{59.304 $\pm$ 0.194}$ & $\tts{2.920 $\pm$ 0.075}$ & $\tts{3.076 $\pm$ 0.041}$ & $\tts{1.901 $\pm$ 0.079}$\\
\midrule
\textbf{CDiff} & \underline{60.661 $\pm$ 0.101} & \textbf{2.293 $\pm$ 0.034} & \textbf{27.101 $\pm$ 0.113} & {16895.629 $\pm$ 741.331} & \ttt{106.184 $\pm$ 1.121} & {59.744 $\pm$ 0.574} & \textbf{2.661 $\pm$ 0.030} & \undl{2.132 $\pm$ 0.131} & \textbf{1.088 $\pm$ 0.031} \\ \bottomrule

\\

&\multicolumn{9}{c}{\textbf{Mooc dataset}}\\ \midrule
&\multicolumn{5}{c|}{\textbf{$N=20$ events forecasting}}     &\multicolumn{4}{c}{\textbf{Interval forecasting $t'$ long}}\\ 
& $\textbf{OTD}$ & $\textbf{RMSE}_{e}$ &$\textbf{RMSE}_{x^+}$ & \textbf{MAPE} &\textbf{sMAPE} & $\textbf{OTD}$ & $\textbf{RMSE}_{e}$&$\textbf{RMSE}_{|\bs^+|}$ & $\textbf{MAE}_{|\bs^+|}$\\ \midrule
\textbf{HYPRO} & \undl{48.621 $\pm$ 0.352} & $\ttt{1.169 $\pm$ 0.094}$ & \textbf{0.410 $\pm$ 0.005} & $\ttt{12592.704 $\pm$ 235.279}$ & \textbf{143.045 $\pm$ 7.992} & $\tts{42.985 $\pm$ 0.113}$ & $\textbf{1.037 $\pm$ 0.027}$ & $\ttt{5.769 $\pm$ 0.207}$ & $\ttt{2.777 $\pm$ 0.119}$\\

\textbf{Dual-TPP} & $\ttt{50.184 $\pm$ 1.127}$ & $\tts{1.312 $\pm$ 0.019}$ & $\tts{0.435 $\pm$ 0.006}$ & $\ttt{12511.299 $\pm$ 131.275}$ & $\tts{147.003 $\pm$ 2.908}$ & $\ttt{41.295 $\pm$ 0.074}$ & $\tts{1.272 $\pm$ 0.016}$ & $\tts{6.121 $\pm$ 0.159}$ & $\ttt{3.255 $\pm$ 0.051}$\\

\textbf{AttNHP} & $\tts{49.121 $\pm$ 0.720}$ & $\ttt{1.297 $\pm$ 0.049}$ & $\ttt{0.420 $\pm$ 0.009}$ & $\ttt{12838.668 $\pm$ 296.147}$ & $\ttt{147.756 $\pm$ 4.812}$ & $\tts{43.001 $\pm$ 0.111}$ & \undl{1.038 $\pm$ 0.025} & \undl{5.591 $\pm$ 0.083} & \undl{2.597 $\pm$ 0.076}\\

\textbf{NHP} & $\tts{51.277 $\pm$ 1.768}$ & $\tts{1.458 $\pm$ 0.063}$ & $\tts{0.442 $\pm$ 0.007}$ & $\ttt{13082.583 $\pm$ 352.970}$ & $\tts{148.913 $\pm$ 11.628}$ & $\textbf{40.933 $\pm$ 0.204}$ & $\tts{1.298 $\pm$ 0.016}$ & $\ttt{6.160 $\pm$ 0.080}$ & $\tts{3.337 $\pm$ 0.047}$\\

\textbf{LogNM} & $\tts{52.890 $\pm$ 1.151}$ & $\tts{1.428 $\pm$ 0.061}$ & $\tts{0.454 $\pm$ 0.008}$ & $\ttt{14868.891 $\pm$ 315.812}$ & $\tts{149.987 $\pm$ 16.581}$ & \undl{41.003 $\pm$ 0.127} & $\tts{1.307 $\pm$ 0.039}$ & $\tts{5.895 $\pm$ 0.057}$ & $\ttt{2.838 $\pm$ 0.063}$\\

\textbf{TCDDM} & $\tts{50.739 $\pm$ 0.765}$ & $\ttt{1.407 $\pm$ 0.112}$ & $\ttt{0.429 $\pm$ 0.015}$ & \undl{12409.522 $\pm$ 267.312} & \undl{145.745 $\pm$ 11.835} & $\ttt{42.662 $\pm$ 0.200}$ & $\tts{1.199 $\pm$ 0.057}$ & $\ttt{5.634 $\pm$ 0.094}$ & $\ttt{2.663 $\pm$ 0.091}$\\
\textbf{Homog. Poisson} & $\tts{58.568 $\pm$ 0.147}$ & \undl{1.161 $\pm$ 0.004} & $\tts{0.536 $\pm$ 0.002}$ & $\ttt{235478.446 $\pm$ 353.632}$ & $\tts{175.587 $\pm$ 10.333}$ & $\tts{43.442 $\pm$ 0.716}$ & $\tts{1.088 $\pm$ 0.037}$ & $\tts{6.943 $\pm$ 0.155}$ & $\tts{3.741 $\pm$ 0.112}$\\
\midrule
\textbf{CDiff} & $\textbf{47.214 $\pm$ 0.628}$   & \textbf{{1.095 $\pm$ 0.048}} & \undl{{0.411 $ \pm$ 0.009}} & \textbf{12243.367 $\pm$ 188.453} & $\tts{146.361 $\pm$ 14.837}$ & $\tts{42.118 $\pm$ 0.171}$ & $\ttt{1.041 $\pm$ 0.021}$ & $\textbf{5.584 $\pm$ 0.186}$ & $\textbf{2.566 $\pm$ 0.092}$\\ 

\\

&\multicolumn{9}{c}{\textbf{Amazon dataset}}\\ \midrule
&\multicolumn{5}{c|}{\textbf{$N=20$ events forecasting}}     &\multicolumn{4}{c}{\textbf{Interval forecasting $t'$ long}}\\ 
& $\textbf{OTD}$ & $\textbf{RMSE}_{e}$ &$\textbf{RMSE}_{x^+}$ & \textbf{MAPE} &\textbf{sMAPE} & $\textbf{OTD}$ & $\textbf{RMSE}_{e}$&$\textbf{RMSE}_{|\bs^+|}$ & $\textbf{MAE}_{|\bs^+|}$\\ \midrule
\textbf{HYPRO} & $\tts{\undl{38.613 $\pm$ 0.536}}$ & $\textbf{2.007 $\pm$ 0.054}$ & $\tts{0.477 $\pm$ 0.010}$ & $\ttt{1247.592 $\pm$ 96.544}$ & \undl{82.506 $\pm$ 0.840}  & \undl{38.229 $\pm$ 0.052} & $\textbf{1.995 $\pm$ 0.005}$ & \undl{0.986 $\pm$ 0.011} & $\textbf{0.414 $\pm$ 0.004}$\\

\textbf{Dual-TPP} & $\tts{42.646 $\pm$ 0.752}$ & $\ttt{2.562 $\pm$ 0.202}$ & $\tts{0.482 $\pm$ 0.012}$ & $\ttt{1414.225 $\pm$ 70.306}$ & $\ttt{86.453 $\pm$ 2.044}$ & $\tts{40.987 $\pm$ 0.490}$ & $\tts{2.410 $\pm$ 0.034}$ & $\tts{1.269 $\pm$ 0.011}$ & $\tts{0.617 $\pm$ 0.003}$\\

\textbf{AttNHP} & $\ttt{39.480 $\pm$ 0.326}$ & $\tts{2.166 $\pm$ 0.026}$ & \undl{0.476 $\pm$ 0.033} & $\ttt{1372.409 $\pm$ 53.202}$ & $\tts{84.323 $\pm$ 1.815}$ & $\ttt{39.870 $\pm$ 0.641}$ & $\ttt{2.042 $\pm$ 0.031}$ & $\ttt{0.998 $\pm$ 0.008}$ & $\ttt{0.417 $\pm$ 0.005}$\\

\textbf{NHP} & $\tts{42.571 $\pm$ 0.293}$ & $\ttt{2.561 $\pm$ 0.060}$ & $\tts{0.519 $\pm$ 0.023}$ & $\tts{1426.601 $\pm$ 16.437}$ & $\tts{92.053 $\pm$ 1.553}$ & $\tts{41.110 $\pm$ 0.272}$ & $\tts{2.447 $\pm$ 0.053}$ & $\tts{1.278 $\pm$ 0.005}$ & $\tts{0.603 $\pm$ 0.005}$\\

\textbf{LNM} & $\tts{43.820 $\pm$ 0.232}$ & $\tts{3.050 $\pm$ 0.286}$ & $\tts{0.481 $\pm$ 0.145}$ & $\tts{1523.064 $\pm$ 312.396}$ & $\tts{90.910 $\pm$ 1.611}$ & $\ttt{41.953 $\pm$ 0.395}$ & $\tts{2.872 $\pm$ 0.015}$ & $\tts{ 1.268 $\pm$ 0.007}$ & $\tts{0.614 $\pm$ 0.009}$ \\  

\textbf{TCDDM} & $\tts{42.245 $\pm$ 0.174}$ & $\tts{2.998 $\pm$ 0.115}$ & \undl{0.476 $\pm$ 0.111} & $\textbf{1086.146 $\pm$ 94.188}$ & $\ttt{83.826 $\pm$ 1.508}$ & $\ttt{40.432 $\pm$ 0.307}$ & $\ttt{2.797 $\pm$ 0.048}$ & $\tts{0.996 $\pm$ 0.004}$ & $\tts{0.429 $\pm$ 0.003}$\\
\textbf{Homog. Poisson} & $\tts{43.940 $\pm$ 0.360}$ & $\tts{4.870 $\pm$ 0.019}$ & $\tts{0.691 $\pm$ 0.004}$ & $\tts{1775.151 $\pm$ 37.202}$ & $\tts{112.392 $\pm$ 0.464}$ & $\tts{42.713 $\pm$ 0.474}$ & $\tts{3.526 $\pm$ 0.037}$ & $\tts{1.524 $\pm$ 0.009}$ & $\tts{0.934 $\pm$ 0.006}$\\
\midrule
\textbf{CDiff}    & \textbf{37.728  $\pm$  0.199}   & \undl{{2.091 $\pm$ 0.163}} & \textbf{{0.464 $ \pm$ 0.086}}& \undl{1189.691 $\pm$ 71.215} &\textbf{81.987 $\pm$ 1.905} & $\textbf{37.068 $\pm$ 0.038}$ & \undl{2.058 $\pm$ 0.009} & \textbf{0.961 $\pm$ 0.018} & \undl{0.416 $\pm$ 0.006} \\ 

\end{tabular}}

\label{tab:complete_table_n20}
\end{table*}

\begin{table*}[h]
\small
\caption{Results for all metrics across 7 different datasets for \textbf{$N=10$ events forecasting} and \textbf{medium interval forecasting}, bold case indicates the best, under line indicates the second best,  * indicates stats. significance w.r.t. the method with the lowest value}

\centering\resizebox{\linewidth}{!}{\begin{tabular}{lccccc|cccc} 

\\
&\multicolumn{9}{c}{\textbf{Synthetic dataset}}\\ \midrule
&\multicolumn{5}{c|}{\textbf{$N=10$ events forecasting}}     &\multicolumn{4}{c}{\textbf{Interval forecasting $t'$ medium}}\\ 
& $\textbf{OTD}$ & $\textbf{RMSE}_{e}$ &$\textbf{RMSE}_{x^+}$ & \textbf{MAPE} &\textbf{sMAPE} & $\textbf{OTD}$ & $\textbf{RMSE}_{e}$&$\textbf{RMSE}_{|\bs^+|}$ & $\textbf{MAE}_{|\bs^+|}$\\ \midrule
\textbf{HYPRO} & \textbf{12.962 $\pm$ 0.128} & \textbf{1.747 $\pm$ 0.041} & {0.104 $\pm$ 0.006 } & {612.354 $\pm$ 21.017} & {99.473 $\pm$ 0.767} & \textbf{13.263 $\pm$ 0.213 } & \tbf{1.721 $\pm$ 0.011} & \tbf{1.404 $\pm$ 0.023} & \tbf{0.561 $\pm$ 0.029} \\
\textbf{Dual-TPP} & $\tts{14.141 $\pm$ 0.125}$ & $\tts{1.965 $\pm$ 0.053}$ & $\tts{0.108 $\pm$ 0.008}$ & $\tts{713.157 $\pm$ 30.615}$ & $\tts{99.688 $\pm$ 0.672}$ & $\tts{13.919 $\pm$0.271}$ & $\tts{1.777 $\pm$ 0.019}$ & $\tts{2.109 $\pm$ 0.048}$ & $\tts{0.667 $\pm$ 0.033}$ \\ 
\textbf{Attnhp} & {13.916 $\pm$ 0.110} & $\tts{1.851 $\pm$ 0.039}$ & \underline{0.103 $\pm$ 0.009} & \underline{587.161 $\pm$ 41.113} & $\tts{100.041 $\pm$ 0.551}$ & $\ttt{13.654 $\pm$ 0.163}$ & $\tts{1.799 $\pm$ 0.018}$ & $\tts{1.517 $\pm$ 0.045}$ & $\tts{0.741 $\pm$ 0.019}$ \\ 
\textbf{NHP} & {13.588 $\pm$ 0.313} & $\tts{1.801 $\pm$ 0.016}$ & $\tts{0.107 $\pm$ 0.005}$ & $\tts{712.673 $\pm$ 66.121}$ & $\tts{{99.343 $\pm$ 0.721}}$ & $\tts{13.551 $\pm$ 0.197}$ & {1.801 $\pm$ 0.030} & \undl{1.408 $\pm$ 0.051} & {0.590 $\pm$ 0.027} \\  

\textbf{LogNM} & $\tts{13.969 $\pm$ 0.266}$ & $\tts{1.915 $\pm$ 0.029}$ & $\ttt{0.105 $\pm$ 0.004}$ & $\tts{667.876 $\pm$ 58.456}$ & $\ttt{99.552 $\pm$ 0.901}$ & $\tts{13.784 $\pm$ 0.192}$ & $\ttt{1.779 $\pm$ 0.029}$ & $\tts{1.493 $\pm$ 0.043}$ & \undl{0.571 $\pm$ 0.033}\\

\textbf{TCDDM} & $\ttt{\undl{13.503 $\pm$ 0.160}}$ & $\ttt{1.863 $\pm$ 0.037}$ & $\ttt{0.105 $\pm$ 0.001}$ & $\ttt{632.431 $\pm$ 42.223}$ & \undl{99.267 $\pm$ 0.576} & $\tts{13.559 $\pm$ 0.177}$ & $\ttt{1.761 $\pm$ 0.032}$ & $\tts{1.485 $\pm$ 0.025}$ & $\tts{0.631 $\pm$ 0.011}$\\

\textbf{Homog. Poisson} & $\tts{15.532 $\pm$ 0.197}$ & $\tts{2.057 $\pm$ 0.018}$ & $\tts{0.143 $\pm$ 0.005}$ & $\tts{1014.814 $\pm$ 72.140}$ & $\tts{101.156 $\pm$ 0.601}$ & $\tts{15.240 $\pm$ 0.232}$ & $\tts{1.923 $\pm$ 0.087}$ & $\tts{1.740 $\pm$ 0.049}$ & $\tts{1.041 $\pm$ 0.019}$\\
\midrule
\textbf{CDiff} & {13.792 $\pm$ 0.251} & \underline{1.786 $\pm$ 0.019} & \textbf{0.096 $\pm$ 0.005} & \textbf{419.982 $\pm$ 52.083} & \textbf{99.063 $\pm$ 0.523} & \underline{13.371 $\pm$ 0.572} & $\tts{\undl{1.773 $\pm$ 0.017}}$ & $\tts{1.473 $\pm$ 0.035}$ & $\tts{0.632 $\pm$ 0.015}$ \\ \bottomrule \\
    
&\multicolumn{9}{c}{\textbf{Taxi dataset}}\\ \midrule
&\multicolumn{5}{c|}{\textbf{$N=10$ events forecasting}}     &\multicolumn{4}{c}{\textbf{Interval forecasting $t'$ medium}}\\ 
& $\textbf{OTD}$ & $\textbf{RMSE}_{e}$ &$\textbf{RMSE}_{x^+}$ & \textbf{MAPE} &\textbf{sMAPE} & $\textbf{OTD}$ & $\textbf{RMSE}_{e}$&$\textbf{RMSE}_{|\bs^+|}$ & $\textbf{MAE}_{|\bs^+|}$\\ \midrule
\textbf{HYPRO} & \undl{11.875 $\pm$ 0.172} & \tbf{0.764 $\pm$ 0.008} & \undl{0.363 $\pm$ 0.002} & {261.896 $\pm$ 33.712} & {89.524 $\pm$ 0.552} & \undl{10.184 $\pm$ 0.191} & \tbf{0.906 $\pm$ 0.019} & \undl{2.976 $\pm$ 0.093} & \undl{2.216 $\pm$ 0.061} \\
\textbf{Dual-TPP} & $\tts{13.058 $\pm$ 0.220}$ & $\tts{0.966 $\pm$ 0.011}$ & $\tts{0.395 $\pm$ 0.003}$ & $\tts{268.407 $\pm$ 41.313}$ & $\tts{\undl{90.812 $\pm$ 0.497}}$ & $\tts{11.031 $\pm$ 0.227}$ & $\tts{1.044 $\pm$ 0.027}$ & $\tts{3.478 $\pm$ 0.147}$ & $\tts{2.547 $\pm$ 0.127}$ \\ 
\textbf{Attnhp} & {12.542 $\pm$ 0.336} & {0.823 $\pm$ 0.007} & $\tts{0.376 $\pm$ 0.003}$ & \undl{253.040 $\pm$ 37.710} & {92.812 $\pm$ 0.129} & $\tts{10.339 $\pm$ 0.194}$ & {0.929 $\pm$ 0.031} & $\tts{3.249 $\pm$ 0.099}$ & $\tts{2.341 $\pm$ 0.147}$ \\ 
\textbf{NHP} & $\tts{13.377 $\pm$ 0.184}$ & $\tts{0.922 $\pm$ 0.009}$ & $\tts{0.397 $\pm$ 0.005}$ & $\tts{269.204 $\pm$ 28.418}$ & $\tts{92.182 $\pm$ 0.384}$ & $\tts{11.115 $\pm$ 0.209}$ & $\tts{1.044 $\pm$ 0.017}$ & $\tts{3.523 $\pm$ 0.102}$ & $\tts{2.548 $\pm$ 0.121}$ \\  

\textbf{LogNM} & $\tts{12.765 $\pm$ 0.106}$ & $\tts{1.004 $\pm$ 0.013}$ & $\tts{0.383 $\pm$ 0.015}$ & $\ttt{263.311 $\pm$ 26.418}$ & $\tts{93.120 $\pm$ 0.526}$ & $\tts{10.527 $\pm$ 0.140}$ & $\tts{0.958 $\pm$ 0.033}$ & $\tts{3.398 $\pm$ 0.158}$ & $\tts{2.431 $\pm$ 0.106}$\\

\textbf{TCDDM} & $\ttt{11.885 $\pm$ 0.149}$ & $\tts{1.121 $\pm$ 0.072}$ & $\tts{0.385 $\pm$ 0.009}$ & $\ttt{254.312 $\pm$ 33.659}$ & $\ttt{90.703 $\pm$ 0.356}$ & $\tts{10.209 $\pm$ 0.337}$ & $\tts{0.998 $\pm$ 0.035}$ & $\tts{3.441 $\pm$ 0.201}$ & $\ttt{2.339 $\pm$ 0.154}$\\

\textbf{Homog. Poisson} & $\tts{14.209 $\pm$ 0.097}$ & $\tts{1.402 $\pm$ 0.033}$ & $\tts{0.397 $\pm$ 0.004}$ & $\tts{279.410 $\pm$ 19.417}$ & $\tts{96.350$\pm$ 0.513}$ & $\tts{11.059 $\pm$ 0.172 }$ & $\tts{1.112 $\pm$ 0.0315}$ & $\tts{4.065 $\pm$ 0.197}$ & $\tts{2.994 $\pm$ 0.251}$\\
\midrule
\textbf{CDiff} & \tbf{11.004 $\pm$ 0.191} & \undl{0.785 $\pm$ 0.007} & \tbf{0.350 $\pm$ 0.002} & \tbf{236.572 $\pm$ 35.459} & \tbf{90.721 $\pm$ 0.291} & \tbf{9.335 $\pm$ 0.211} & \undl{0.926 $\pm$ 0.023} & \tbf{2.972 $\pm$ 0.111} & \tbf{2.117 $\pm$ 0.090} \\ \bottomrule

\\
&\multicolumn{9}{c}{\textbf{Taobao dataset}}\\ \midrule
&\multicolumn{5}{c|}{\textbf{$N=10$ events forecasting}}     &\multicolumn{4}{c}{\textbf{Interval forecasting $t'$ medium}}\\ 
& $\textbf{OTD}$ & $\textbf{RMSE}_{e}$ &$\textbf{RMSE}_{x^+}$ & \textbf{MAPE} &\textbf{sMAPE} & $\textbf{OTD}$ & $\textbf{RMSE}_{e}$&$\textbf{RMSE}_{|\bs^+|}$ & $\textbf{MAE}_{|\bs^+|}$\\ \midrule
\textbf{HYPRO} & $\tts{21.547 $\pm$ 0.138}$ & $\tts{1.527 $\pm$ 0.035}$ & $\ttt{\undl{0.591 $\pm$ 0.019}}$ & \tbf{5968.317 $\pm$ 240.664} & {133.147 $\pm$ 0.341} & $\tts{20.101 $\pm$ 0.127}$ & $\tts{1.671 $\pm$ 0.012}$ & $\ttt{2.403 $\pm$ 0.042}$ & $\ttt{1.391 $\pm$ 0.023}$ \\
\textbf{Dual-TPP} & $\tts{23.691 $\pm$ 0.203}$ & $\tts{2.674 $\pm$ 0.032}$ & $\tts{0.873 $\pm$ 0.010}$ & $\tts{8413.261 $\pm$ 222.427}$ & $\tts{139.271 $\pm$ 0.348}$ & \textbf{18.817 $\pm$ 0.215} & $\tts{1.738 $\pm$ 0.010}$ & $\tts{4.207 $\pm$ 0.076}$ & $\tts{2.352 $\pm$ 0.021}$ \\ 
\textbf{Attnhp} & {21.683 $\pm$ 0.215} & $\tts{\undl{1.514 $\pm$ 0.015}}$ & $\tts{0.608 $\pm$ 0.011}$ & \undl{6034.771 $\pm$ 170.267} & {135.271 $\pm$ 0.395} & $\tts{20.653 $\pm$ 0.162}$ & \tbf{1.342 $\pm$ 0.009} & \tbf{2.221 $\pm$ 0.045} & \undl{1.297 $\pm$ 0.011} \\ 
\textbf{NHP} & $\tts{24.068 $\pm$ 0.331}$ & $\tts{2.769 $\pm$ 0.033}$ & $\tts{0.855 $\pm$ 0.013}$ & $\tts{7734.518 $\pm$ 276.670}$ & $\tts{137.693 $\pm$ 0.225}$ & $\tts{\undl{18.991 $\pm$ 0.278}} $ & $\tts{1.862 $\pm$ 0.014}$ & $\tts{3.995 $\pm$ 0.077}$ & $\tts{2.437 $\pm$ 0.017}$ \\  

\textbf{LogNM} & $\tts{23.195 $\pm$ 0.039}$ & $\tts{2.429 $\pm$ 0.045}$ & $\tts{0.602 $\pm$ 0.037}$ & $\ttt{6719.015 $\pm$ 163.868}$ & \undl{127.411 $\pm$ 0.573} & $\tts{19.383 $\pm$ 0.402}$ & $\tts{ 1.826 $\pm$ 0.005}$ & $\tts{3.634 $\pm$ 0.058}$ & $\tts{1.745 $\pm$ 0.014}$\\

\textbf{TCDDM} & $\textbf{21.012 $\pm$ 0.520}$ & $\ttt{2.598 $\pm$ 0.047}$ & $\tts{0.610 $\pm$ 0.022}$ & $\ttt{6630.487 $\pm$ 259.540}$ & $\ttt{132.7112 $\pm$ 0.774}$ & $\tts{20.032 $\pm$ 0.691}$ & $\tts{1.558 $\pm$ 0.015}$ & $\tts{2.951 $\pm$ 0.069}$ & $\ttt{1.649 $\pm$ 0.0183}$\\

\textbf{Homog. Poisson} & $\tts{27.353 $\pm$ 0.426}$ & $\tts{2.772 $\pm$ 0.016}$ & $\tts{0.887 $\pm$ 0.014}$ & $\tts{19301.747 $\pm$ 349.301}$ & $\tts{155.236 $\pm$ 0.729}$ & $\tts{19.251 $\pm$ 0.221}$ & $\tts{1.920 $\pm$ 0.009}$ & $\tts{5.001 $\pm$0.022 }$ & $\tts{3.209 $\pm$ 0.011}$\\

\midrule
\textbf{CDiff} & \undl{21.221 $\pm$ 0.176} & \tbf{1.416 $\pm$ 0.024} & \tbf{0.535 $\pm$ 0.016} & {6718.144 $\pm$ 161.416} & \tbf{126.824 $\pm$ 0.366} & $\tts{19.677 $\pm$ 0.103}$ & \undl{1.438 $\pm$ 0.012} & \undl{2.307 $\pm$ 0.059} & \tbf{1.160 $\pm$ 0.019} \\ \bottomrule

\\
&\multicolumn{9}{c}{\textbf{Stackoverflow dataset}}\\ \midrule
&\multicolumn{5}{c|}{\textbf{$N=10$ events forecasting}}     &\multicolumn{4}{c}{\textbf{Interval forecasting $t'$ medium}}\\ 
& $\textbf{OTD}$ & $\textbf{RMSE}_{e}$ &$\textbf{RMSE}_{x^+}$ & \textbf{MAPE} &\textbf{sMAPE} & $\textbf{OTD}$ & $\textbf{RMSE}_{e}$&$\textbf{RMSE}_{|\bs^+|}$ & $\textbf{MAE}_{|\bs^+|}$\\ \midrule
\textbf{HYPRO} & \undl{21.062 $\pm$ 0.372} & \undl{0.921 $\pm$ 0.019} & {1.235 $\pm$ 0.006} & $\tts{1925.362 $\pm$ 149.208}$ & $\tts{107.566 $\pm$ 0.218} $& \undl{18.523 $\pm$ 0.301} & \undl{0.907 $\pm$ 0.013} & \undl{2.327 $\pm$ 0.040} & {1.339 $\pm$ 0.033} \\
\textbf{Dual-TPP} & $\tts{21.229 $\pm$ 0.394}$ & $\ttt{0.936 $\pm$ 0.013}$ & $\tts{\undl{1.223 $\pm$ 0.010}}$ & $\tts{1845.469 $\pm$ 103.450}$ & $\tts{107.274 $\pm$ 0.200}$ &$ \tts{19.155 $\pm$ 0.116}$ & $\tts{0.923 $\pm$ 0.011}$ & $\tts{2.344 $\pm$ 0.053}$ & $\tts{1.478 $\pm$ 0.038}$ \\ 
\textbf{Attnhp} & $\tts{22.019 $\pm$ 0.220}$ & $\ttt{0.978 $\pm$ 0.023}$ & $\tts{1.225 $\pm$ 0.007}$ & \tbf{1571.807 $\pm$ 99.921 }  & \tbf{100.137 $\pm$ 0.167} & $\tts{19.487 $\pm$ 0.130}$ & $\tts{0.973 $\pm$ 0.013}$ & $\tts{2.415 $\pm$ 0.026}$ & $\tts{1.455 $\pm$ 0.025}$ \\ 
\textbf{NHP} &  $\tts{21.655 $\pm$ 0.314}$ & $\tts{0.970 $\pm$ 0.014}$ & $\tts{1.266 $\pm$ 0.003}$ & \undl{1698.947 $\pm$ 123.208} & $\tts{108.867 $\pm$ 0.361}$ & $\tts{19.314 $\pm$ 0.098}$ & $\tts{0.959 $\pm$ 0.017}$ & $\tts{2.481 $\pm$ 0.035}$ & $\tts{1.419 $\pm$ 0.031}$ \\  

\textbf{LogNM} & $\tts{22.339 $\pm$ 0.322}$ & $\ttt{0.970 $\pm$ 0.011}$ & $\ttt{1.251 $\pm$ 0.005}$ & $\tts{1841.119 $\pm$ 71.077}$ & $\ttt{105.674 $\pm$ 0.337}$ & $\ttt{19.303 $\pm$ 0.137}$ & $\ttt{0.955 $\pm$ 0.014}$ & $\tts{2.751 $\pm$ 0.028}$ & $\tts{1.487 $\pm$ 0.046}$\\

\textbf{TCDDM} & $\tts{22.042 $\pm$ 0.193}$ & $\ttt{1.205 $\pm$ 0.014}$ & $\tts{1.228 $\pm$ 0.010}$ & $\tts{1772.325 $\pm$ 221.358}$ & $\tts{108.1113 $\pm$ 0.112}$ & $\ttt{18.920 $\pm$ 0.125}$ & $\ttt{0.930 $\pm$ 0.015}$ & $\ttt{2.472 $\pm$ 0.033}$ & \undl{1.293 $\pm$ 0.050}\\

\textbf{Homog. Poisson} & $\tts{23.115 $\pm$ 0.318}$ & $\ttt{1.012 $\pm$ 0.027}$ & $\tts{1.327 $\pm$ 0.004}$ & $\tts{2105.433 $\pm$ 88.409}$ & $\tts{108.322 $\pm$ 0.315}$ & $\tts{22.714 $\pm$ 0.300}$ & $\tts{0.973 $\pm$ 0.023}$ & $\tts{2.889 $\pm$ 0.020}$ & $\tts{1.597 $\pm$ 0.021}$ \\

\midrule
\textbf{CDiff} & \tbf{20.191 $\pm$ 0.455} & \tbf{0.916 $\pm$ 0.010} & \tbf{1.180 $\pm$ 0.003} & {1880.59 $\pm$ 78.283} & $\tts{\undl{102.367 $\pm$ 0.267}}$ & \tbf{18.268 $\pm$ 0.167} & \tbf{0.883 $\pm$ 0.009} & \tbf{2.107 $\pm$ 0.031} & \tbf{1.219 $\pm$ 0.023} \\ \bottomrule
\\
&\multicolumn{9}{c}{\textbf{Retweet dataset}}\\ \midrule
&\multicolumn{5}{c|}{\textbf{$N=10$ events forecasting}}     &\multicolumn{4}{c}{\textbf{Interval forecasting $t'$ medium}}\\ 
& $\textbf{OTD}$ & $\textbf{RMSE}_{e}$ &$\textbf{RMSE}_{x^+}$ & \textbf{MAPE} &\textbf{sMAPE} & $\textbf{OTD}$ & $\textbf{RMSE}_{e}$&$\textbf{RMSE}_{|\bs^+|}$ & $\textbf{MAE}_{|\bs^+|}$\\ \midrule
\textbf{HYPRO} & $\tts{31.743 $\pm$ 0.068}$ & $\tts{1.927 $\pm$ 0.027}$ & $\tts{33.683 $\pm$ 0.245}$ & $\tts{17696.498 $\pm$ 986.684}$ & \tbf{105.073 $\pm$ 0.958} & \undl{27.411 $\pm$ 0.190} & $\tts{2.013 $\pm$ 0.032}$ & $\tts{2.741 $\pm$ 0.108}$ & $\tts{1.971 $\pm$ 0.031}$\\
\textbf{Dual-TPP} & $\tts{31.652 $\pm$ 0.075}$ & $\tts{1.963 $\pm$ 0.038}$ & $\tts{28.104 $\pm$ 0.486}$ & $\tts{17553.619 $\pm$ 731.120}$ & $\tts{106.721 $\pm$ 0.774}$ & $\tts{28.357 $\pm$ 0.176}$ & $\tts{1.991 $\pm$ 0.050}$ & {1.963 $\pm$ 0.094} & {1.615 $\pm$ 0.037}\\ 
\textbf{Attnhp} & \tbf{30.337 $\pm$ 0.065} & $\tts{1.823 $\pm$ 0.031}$ & \tbf{26.310 $\pm$ 0.333} & \tbf{14377.241$\pm$ 1319.797} & {106.021 $\pm$ 1.011} & \textbf{26.787 $\pm$ 0.114} & \tbf{1.961 $\pm$ 0.029} & $\tts{1.981 $\pm$ 0.115}$ & $\tts{1.597 $\pm$ 0.058}$\\ 
\textbf{NHP} & \undl{30.817 $\pm$ 0.090} & \tbf{1.713 $\pm$ 0.024} & $\tts{27.010 $\pm$ 0.429}$ & $\tts{\undl{15214.175$\pm$ 695.184}}$ & $\tts{107.053 $\pm$ 1.390}$ & $\tts{27.617 $\pm$ 0.099}$ & $\tts{1.997 $\pm$ 0.047}$ & $\tts{\undl{1.959 $\pm$ 0.124}}$ & $\tts{{1.562 $\pm$ 0.080}}$\\   

\textbf{LogNM} & $\tts{31.974 $\pm$ 0.032}$ & $\tts{1.942 $\pm$ 0.062}$ & $\ttt{28.825 $\pm$ 0.221}$ & $\tts{17339.802 $\pm$ 765.475}$ & $\ttt{106.014 $\pm$ 0.633}$ & $\tts{27.283 $\pm$ 0.078}$ & $\tts{1.995 $\pm$ 0.026}$ & $\ttt{2.327 $\pm$ 0.126}$ & $\ttt{1.649 $\pm$ 0.069}$\\

\textbf{TCDDM} & $\ttt{32.006 $\pm$ 0.074}$ & $\ttt{1.789 $\pm$ 0.094}$ & $\ttt{29.124 $\pm$ 0.405}$ & $\ttt{18874.939 $\pm$828.544}$ & $\ttt{106.738 $\pm$ 0.791}$ & $\ttt{27.993 $\pm$ 0.230}$ & $\tts{2.035 $\pm$ 0.047}$ & $\ttt{1.997 $\pm$ 0.215}$ & \undl{1.337 $\pm$ 0.080}\\

\textbf{Homog. Poisson} & $\ttt{30.885 $\pm$ 0.017}$ & $\tts{1.987 $\pm$ 0.036}$ & $\tts{33.241 $\pm$ 0.512}$ & $\tts{17892.301 $\pm$ 355.213}$ & $\tts{114.286 $\pm$ 0.753}$ & $\ttt{26.950 $\pm$ 0.306}$ & $\tts{1.987 $\pm$ 0.026}$ & $\tts{2.774 $\pm$ 0.118}$ & $\tts{2.023 $\pm$ 0.0355}$\\

\midrule
\textbf{CDiff} &  $\tts{31.237 $\pm$ 0.078}$ & \undl{1.745 $\pm$ 0.036} & \undl{26.429 $\pm$ 0.201} & {15636.184 $\pm$ 713.516} & \undl{105.767 $\pm$ 0.771} & {27.739 $\pm$ 0.105} & \undl{1.973 $\pm$ 0.036} & \tbf{1.907 $\pm$ 0.111} & \tbf{1.299 $\pm$ 0.043} \\ \bottomrule

\\

&\multicolumn{9}{c}{\textbf{Mooc dataset}}\\ \midrule
&\multicolumn{5}{c|}{\textbf{$N=10$ events forecasting}}     &\multicolumn{4}{c}{\textbf{Interval forecasting $t'$ medium}}\\ 
& $\textbf{OTD}$ & $\textbf{RMSE}_{e}$ &$\textbf{RMSE}_{x^+}$ & \textbf{MAPE} &\textbf{sMAPE} & $\textbf{OTD}$ & $\textbf{RMSE}_{e}$&$\textbf{RMSE}_{|\bs^+|}$ & $\textbf{MAE}_{|\bs^+|}$\\ \midrule
\textbf{HYPRO} & $\ttt{25.861 $\pm$ 0.352}$ & \undl{1.032 $\pm$ 0.073} & $\textbf{0.391 $\pm$ 0.002}$ & $\textbf{11931.797 $\pm$ 254.663}$ & $\textbf{142.041 $\pm$ 5.730}$ & $\tts{22.640 $\pm$ 0.171}$ & $\textbf{0.921 $\pm$ 0.063}$ & $\textbf{4.956 $\pm$ 0.277}$ & $\textbf{2.214 $\pm$ 0.057}$\\

\textbf{Dual-TPP} & $\tts{28.785 $\pm$ 0.384}$ & $\tts{1.087 $\pm$ 0.012}$ & $\tts{0.421 $\pm$ 0.006}$ & $\ttt{12721.909 $\pm$ 126.31}$ & $\tts{146.841 $\pm$ 4.188}$ & $\tts{22.359 $\pm$ 0.083}$ & $\ttt{1.028 $\pm$ 0.009}$ & $\tts{5.573 $\pm$ 0.173}$ & $\tts{2.931 $\pm$ 0.029}$\\

\textbf{AttNHP} & $\tts{26.765 $\pm$ 0.221}$ & $\ttt{1.054 $\pm$ 0.009}$ & $\ttt{0.421 $\pm$ 0.011}$ & $\tts{13138.381 $\pm$ 372.632}$ & $\tts{144.641 $\pm$ 3.093}$ & $\tts{23.185 $\pm$ 0.071}$ & $\ttt{0.958 $\pm$ 0.044}$ & \undl{5.105 $\pm$ 0.040} & $\ttt{2.491$\pm$ 0.050}$\\

\textbf{NHP} & $\tts{27.371 $\pm$ 0.632}$ & $\ttt{1.134 $\pm$ 0.064}$ & $\tts{0.429 $\pm$ 0.007}$ & $\tts{13275.513 $\pm$ 262.612}$ & $\tts{143.526$\pm$ 9.509}$ & $\textbf{21.275 $\pm$ 0.051}$ & $\tts{1.038 $\pm$ 0.026}$ & $\tts{5.349 $\pm$ 0.077}$ & $\tts{3.163 $\pm$ 0.043}$\\

\textbf{LogNM} & $\tts{29.497 $\pm$ 0.325}$ & $\tts{1.120 $\pm$ 0.037}$ & $\tts{0.433 $\pm$ 0.013}$ & $\ttt{12692.049 $\pm$ 255.629}$ & $\tts{144.093 $\pm$ 5.077}$ & \undl{21.727 $\pm$ 0.183} & $\tts{1.121 $\pm$ 0.018}$ & $\ttt{5.297 $\pm$ 0.029}$ & $\tts{3.099 $\pm$ 0.060}$\\

\textbf{TCDDM} & $\textbf{24.515 $\pm$ 0.339}$ & $\tts{1.218 $\pm$ 0.065}$ & $\ttt{0.425 $\pm$ 0.019}$ & \undl{11958.023 $\pm$ 267.593} & \undl{143.293 $\pm$ 12.089} & $\tts{23.020 $\pm$ 0.145}$ & $\tts{1.126 $\pm$ 0.052}$ & $\tts{5.224 $\pm$ 0.075}$ & \ttt{2.476 $\pm$ 0.049}\\
\textbf{Homog. Poisson} & $\tts{33.349 $\pm$ 0.143}$ & $\tts{1.269 $\pm$ 0.006}$ & $\tts{0.443 $\pm$ 0.016}$ & $\tts{232853.735 $\pm$ 71.130}$ & $\tts{168.305 $\pm$ 3.126}$ & $\ttt{21.950 $\pm$ 0.043}$ & $\ttt{1.183 $\pm$ 0.017}$ & $\tts{6.036 $\pm$ 0.261}$ & $\tts{3.330 $\pm$ 0.026  }$\\
\midrule
\textbf{CDiff} & \undl{24.544 $\pm$ 0.305}   & $\textbf{{0.944 $\pm$ 0.032}}$ & \undl{{0.404 $ \pm$ 0.003}} & $\ttt{12052.014 $\pm$ 213.141}$ & $\tts{144.313 $\pm$ 8.726}$ & $\tts{22.768 $\pm$ 0.125}$ & \undl{0.935 $\pm$ 0.074} & $\tts{5.120 $\pm$ 0.116}$ & \undl{2.439 $\pm$ 0.034}\\ 

\\

&\multicolumn{9}{c}{\textbf{Amazon dataset}}\\ \midrule
&\multicolumn{5}{c|}{\textbf{$N=10$ events forecasting}}     &\multicolumn{4}{c}{\textbf{Interval forecasting $t'$ medium}}\\ 
& $\textbf{OTD}$ & $\textbf{RMSE}_{e}$ &$\textbf{RMSE}_{x^+}$ & \textbf{MAPE} &\textbf{sMAPE} & $\textbf{OTD}$ & $\textbf{RMSE}_{e}$&$\textbf{RMSE}_{|\bs^+|}$ & $\textbf{MAE}_{|\bs^+|}$\\ \midrule
\textbf{HYPRO} & $\ttt{24.956 $\pm$ 0.663}$ & \undl{1.765 $\pm$ 0.039} & $\textbf{0.442 $\pm$ 0.015}$ & \undl{1211.590 $\pm$ 62.458} & $\ttt{83.401 $\pm$ 1.033}$  & $\tts{24.096 $\pm$ 0.043}$ & \undl{1.678 $\pm$ 0.024} & \undl{0.987 $\pm$ 0.009} & $\textbf{0.408 $\pm$ 0.010}$\\

\textbf{Dual-TPP} & $\tts{25.929 $\pm$ 0.280}$ & $\tts{2.098 $\pm$ 0.101}$ & $\tts{0.475 $\pm$ 0.008}$ & $\tts{1376.448 $\pm$ 104.345}$ & $\ttt{82.352 $\pm$ 1.285}$ & $\tts{23.688 $\pm$ 0.411}$ & $\tts{2.208 $\pm$ 0.094}$ & $\tts{1.162 $\pm$ 0.031}$ & $\tts{0.612 $\pm$ 0.009}$\\

\textbf{AttNHP} & $\textbf{24.116 $\pm$ 0.807}$ & $\textbf{1.741 $\pm$ 0.039}$ & $\ttt{0.454 $\pm$ 0.014}$ & $\ttt{1323.165 $\pm$ 62.289}$ & $\ttt{84.323 $\pm$ 1.815}$ & $\tts{24.278 $\pm$ 0.218}$ & $\ttt{1.693$\pm$ 0.067}$ & $\ttt{0.998 $\pm$ 0.005}$ & $\ttt{0.431 $\pm$ 0.010}$\\

\textbf{NHP} & $\tts{25.730 $\pm$ 0.497}$ & $\tts{1.843 $\pm$ 0.053}$ & $\tts{0.491 $\pm$ 0.048}$ & $\tts{1426.601 $\pm$ 16.437}$ & $\tts{89.135 $\pm$ 1.092}$ & $\textbf{22.506 $\pm$ 0.141}$ & $\tts{1.884 $\pm$ 0.092}$ & $\ttt{1.218 $\pm$ 0.006}$ & $\tts{0.566 $\pm$ 0.010}$\\

\textbf{LNM} & $\tts{26.632 $\pm$ 0.519}$ & $\tts{1.955 $\pm$ 0.112}$ & $\tts{0.464 $\pm$ 0.066}$ & $\tts{1555.852 $\pm$ 33.930}$ & $\tts{89.305 $\pm$ 1.288}$ & \undl{23.049 $\pm$ 0.412} & $\tts{2.658 $\pm$ 0.030}$ & $\tts{ 1.117 $\pm$ 0.009}$ & $\tts{0.513 $\pm$ 0.008}$ \\  

\textbf{TCDDM} & $\tts{25.091 $\pm$ 0.227}$ & $\ttt{1.778 $\pm$ 0.090}$ & \undl{0.448 $\pm$ 0.082} & $\ttt{1274.340 $\pm$ 92.095}$ & $\textbf{82.105 $\pm$ 1.564}$ & $\tts{24.007 $\pm$ 0.109}$ & $\tts{2.103 $\pm$ 0.043}$ & $\textbf{0.980 $\pm$ 0.004}$ & $\ttt{0.430 $\pm$ 0.011}$\\
\textbf{Homog. Poisson} & $\tts{28.945 $\pm$ 0.441}$ & $\tts{3.076 $\pm$ 0.021}$ & $\tts{0.700 $\pm$ 0.009}$ & $\tts{2103.582 $\pm$ 38.491}$ & $\tts{109.143 $\pm$ 0.304}$ & $\ttt{23.745 $\pm$ 0.738}$ & $\tts{1.988 $\pm$ 0.057}$ & $\tts{1.423 $\pm$ 0.005}$ & $\tts{0.847 $\pm$ 0.003}$\\
\midrule
\textbf{CDiff}    & \undl{24.230  $\pm$  0.287}   & \ttt{{1.766$\pm$ 0.079}} & \ttt{{0.450 $ \pm$ 0.049}}& \textbf{1146.530 $\pm$ 43.595} &\undl{82.124 $\pm$ 2.094} & $\tts{23.994 $\pm$ 0.113}$ & $\textbf{1.503 $\pm$ 0.034}$ & $\ttt{1.005 $\pm$ 0.010}$ & \undl{0.409 $\pm$ 0.005} \\

\end{tabular}}

\label{tab:complete_table_n10}
\end{table*}

\begin{table*}[h]
\small
\caption{Results for all metrics across 7 different datasets for \textbf{$N=5$ events forecasting} and \textbf{small interval forecasting}, bold case indicates the best, under line indicates the second best,  * indicates stats. significance w.r.t. the method with the lowest value}

\centering\resizebox{\linewidth}{!}{\begin{tabular}{lccccc|cccc} 

\\
&\multicolumn{9}{c}{\textbf{Synthetic dataset}}\\ \midrule
&\multicolumn{5}{c|}{\textbf{$N=5$ events forecasting}}     &\multicolumn{4}{c}{\textbf{Interval forecasting $t'$ small}}\\ 
& $\textbf{OTD}$ & $\textbf{RMSE}_{e}$ &$\textbf{RMSE}_{x^+}$ & \textbf{MAPE} &\textbf{sMAPE} & $\textbf{OTD}$ & $\textbf{RMSE}_{e}$&$\textbf{RMSE}_{|\bs^+|}$ & $\textbf{MAE}_{|\bs^+|}$\\ \midrule
\textbf{HYPRO} &  $\ttt{8.706 $\pm$ 0.138}$ & {1.216 $\pm$ 0.023} & {0.091 $\pm$ 0.003} & $\tts{510.171 $\pm$ 23.802}$ & {98.857 $\pm$ 0.185} & {8.230 $\pm$ 0.210} & $\tts{1.184 $\pm$ 0.051}$ & $\tts{1.281 $\pm$ 0.079}$ & $\tts{0.724 $\pm$ 0.048}$ \\
\textbf{Dual-TPP} &  $\tts{8.644 $\pm$ 0.102}$ & $\tts{1.280 $\pm$ 0.011}$ & $\tts{0.093 $\pm$ 0.001}$ & $\tts{453.129 $\pm$ 27.592}$ & \undl{98.683 $\pm$ 0.351} & $\tts{8.248 $\pm$ 0.235}$ & {1.177 $\pm$ 0.047} & $\tts{1.161 $\pm$ 0.093}$ & $\tts{0.560 $\pm$ 0.018}$ \\
\textbf{Attnhp}     &  {8.687 $\pm$ 0.149} & $\ttt{1.225 $\pm$ 0.031}$ & \undl{0.089 $\pm$ 0.003} & \tbf{415.593 $\pm$ 24.153} & {100.762 $\pm$ 0.020} & {8.342 $\pm$ 0.078} & $\tts{1.192 $\pm$ 0.040}$ & \undl{1.131 $\pm$ 0.059} & \undl{0.528 $\pm$ 0.034 } \\
\textbf{NHP}      &  \undl{8.565 $\pm$ 0.098} & \undl{1.207 $\pm$ 0.017} & $\tts{0.094 $\pm$ 0.002}$ & \undl{431.286 $\pm$ 30.272} & $\tts{100.861 $\pm$ 0.183}$ & \undl{8.128 $\pm$ 0.274} & \tbf{1.171 $\pm$ 0.053} & $\tts{1.217 $\pm$ 0.073}$ & $\tts{0.608 $\pm$0.023}$ \\ 

\textbf{LogNM} & $\tts{10.093 $\pm$ 0.145}$ & $\tts{1.390 $\pm$ 0.019}$ & $\tts{0.093 $\pm$ 0.005}$ & $\tts{482.341 $\pm$ 29.601}$ & $\tts{101.984 $\pm$ 0.147}$ & $\ttt{8.449 $\pm$ 0.093}$ & $\tts{1.244 $\pm$ 0.101}$ & $\tts{1.239 $\pm$ 0.028}$ & $\ttt{0.552 $\pm$ 0.011}$\\

\textbf{TCDDM} & $\tts{8.881 $\pm$ 0.112}$ & $\tts{1.295 $\pm$ 0.008}$ & $\ttt{0.095 $\pm$ 0.001}$ & $\ttt{472.54 $\pm$ 33.634}$ & $\ttt{99.008 $\pm$ 0.251}$ & $\ttt{8.593 $\pm$ 0.185}$ & $\ttt{1.227 $\pm$ 0.061}$ & $\tts{1.221 $\pm$ 0.058}$ & \undl{0.524 $\pm$ 0.023}\\

\textbf{Homog. Poisson} & $\tts{10.23 $\pm$ 0.135}$ & $\tts{1.268 $\pm$ 0.015}$ & $\tts{0.101 $\pm$ 0.005}$ & $\tts{486.35 $\pm$ 20.561}$ & $\tts{101.357 $\pm$ 0.301}$ & $\tts{10.587 $\pm$ 0.227}$ & $\tts{1.265 $\pm$ 0.114}$ & $\tts{1.475 $\pm$ 0.062}$ & $\tts{0.679 $\pm$ 0.027}$\\

\midrule
\textbf{CDiff}    &  \tbf{8.459 $\pm$ 0.167} & \tbf{1.196 $\pm$ 0.015} & \tbf{0.088 $\pm$0.002} & $\ttt{473.506 $\pm$ 15.600}$ & \tbf{98.011 $\pm$ 0.197} & \tbf{8.095 $\pm$ 0.176} & \undl{1.175 $\pm$ 0.059} & \tbf{1.068 $\pm$ 0.035} & \tbf{0.517 $\pm$ 0.039} \\ \bottomrule \\
    
&\multicolumn{9}{c}{\textbf{Taxi dataset}}\\ \midrule
&\multicolumn{5}{c|}{\textbf{$N=5$ events forecasting}}     &\multicolumn{4}{c}{\textbf{Interval forecasting $t'$ small}}\\ 
& $\textbf{OTD}$ & $\textbf{RMSE}_{e}$ &$\textbf{RMSE}_{x^+}$ & \textbf{MAPE} &\textbf{sMAPE} & $\textbf{OTD}$ & $\textbf{RMSE}_{e}$&$\textbf{RMSE}_{|\bs^+|}$ & $\textbf{MAE}_{|\bs^+|}$\\ \midrule
\textbf{HYPRO} & \undl{5.952 $\pm$ 0.126} & \tbf{0.500 $\pm$ 0.011} & \undl{0.322 $\pm$ 0.004} & \tbf{221.745 $\pm$ 5.084} & \tbf{85.994 $\pm$ 0.227 } & \tbf{4.780 $\pm$ 0.214} & \tbf{0.518 $\pm$ 0.010} & \undl{1.893 $\pm$ 0.052} & {1.405 $\pm$ 0.108} \\
\textbf{Dual-TPP} & $\tts{7.534 $\pm$ 0.111}$ & $\tts{0.636 $\pm$ 0.009}$ & {0.340 $\pm$ 0.003} & $\tts{252.822 $\pm$ 3.853}$ & {89.727 $\pm$0.320} &$ \tts{6.225 $\pm$ 0.117}$ & $\tts{0.647 $\pm$ 0.029}$ & $\tts{1.910 $\pm$  0.043}$ & $\tts{1.417 $\pm$ 0.081}$ \\
\textbf{Attnhp} &  $\ttt{6.441 $\pm$ 0.090}$ & {0.682 $\pm$ 0.010} & {0.347 $\pm$0.002} & $\tts{259.480 $\pm$ 4.819}$ & \undl{89.070 $\pm$ 0.152} & {6.201 $\pm$ 0.111} & {0.642 $\pm$ 0.024} & $\tts{1.923 $\pm$ 0.062}$ & \tbf{1.362 $\pm$ 0.095} \\ 
\textbf{NHP}      &  $\tts{7.405 $\pm$ 0.122}$ & $\tts{0.641 $\pm$ 0.013}$ & $\tts{0.351 $\pm$ 0.008}$ & $\tts{231.504 $\pm$ 6.054}$ & $\tts{91.625 $\pm$ 0.177}$ & $\tts{6.244 $\pm$ 0.172}$ & $\tts{0.653 $\pm$ 0.019}$ & $\tts{1.927 $\pm$ 0.038}$ & $\tts{1.387 $\pm$ 0.117}$ \\  

\textbf{LogNM} & $\tts{7.209 $\pm$ 0.184}$ & $\ttt{0.608 $\pm$ 0.008}$ & $\ttt{0.335 $\pm$ 0.003}$ & $\tts{255.600 $\pm$ 4.601}$ & $\ttt{90.512 $\pm$ 0.169}$ & $\tts{6.664 $\pm$ 0.143}$ & $\tts{0.721 $\pm$ 0.013}$ & $\tts{1.897 $\pm$ 0.044}$ & $\ttt{1.401 $\pm$ 0.079}$\\

\textbf{TCDDM} & \textbf{5.877 $\pm$ 0.095} & $\tts{0.648 $\pm$ 0.015}$ & $\ttt{0.327 $\pm$ 0.005}$ & $\ttt{246.121 $\pm$ 5.512}$ & $\ttt{88.051 $\pm$ 0.240}$ & $\ttt{5.792 $\pm$ 0.110}$ & $\ttt{0.683 $\pm$ 0.024}$ & $\tts{1.910 $\pm$ 0.037}$ & $\ttt{1.395 $\pm$ 0.100}$\\

\textbf{Homog. Poisson} & $\tts{6.905 $\pm$ 0.094}$ & $\tts{0.692 $\pm$ 0.007}$ & $\tts{0.393 $\pm$ 0.006}$ & $\tts{272.51 $\pm$ 3.049}$ & $\tts{94.501 $\pm$ 0.192}$ & $\tts{6.520 $\pm$ 0.133}$ & $\tts{0.797 $\pm$ 0.019}$ & $\tts{2.057 $\pm$ 0.012}$ & $\tts{1.584 $\pm$ 0.078}$\\

\midrule
\textbf{CDiff}    &  \ttt{5.966 $\pm$ 0.083} & \undl{0.547 $\pm$ 0.007} & \tbf{0.318 $\pm$ 0.003} & \undl{223.073 $\pm$ 6.221} & {89.535 $\pm$ 0.294} & \undl{5.128 $\pm$ 0.148} & \undl{0.603 $\pm$ 0.025} & \tbf{1.889 $\pm$ 0.019} & \undl{1.363 $\pm$ 0.074}  \\ \bottomrule

\\
&\multicolumn{9}{c}{\textbf{Taobao dataset}}\\ \midrule
&\multicolumn{5}{c|}{\textbf{$N=5$ events forecasting}}     &\multicolumn{4}{c}{\textbf{Interval forecasting $t'$ small}}\\ 
& $\textbf{OTD}$ & $\textbf{RMSE}_{e}$ &$\textbf{RMSE}_{x^+}$ & \textbf{MAPE} &\textbf{sMAPE} & $\textbf{OTD}$ & $\textbf{RMSE}_{e}$&$\textbf{RMSE}_{|\bs^+|}$ & $\textbf{MAE}_{|\bs^+|}$\\ \midrule
\textbf{HYPRO} & {11.317 $\pm$ 0.111} & \undl{0.817$\pm$ 0.037 } & $\tts{0.573 $\pm$ 0.011}$ & \undl{4652.619 $\pm$189.940} & {133.837 $\pm$ 0.524} & $\tts{11.546 $\pm$ 0.124}$ & {0.866$\pm$ 0.016} & \undl{1.402$\pm$ 0.062} & \undl{0.654$\pm$ 0.011 } \\
\textbf{Dual-TPP} &$\tts{13.280 $\pm$ 0.092}$ & $\tts{1.877$\pm$0.014}$ & $\tts{0.691$\pm$ 0.007}$ & $\tts{6828.105 $\pm$ 235.303}$ & $\tts{134.437$\pm$ 0.458}$ & $\tts{9.779$\pm$ 0.194}$ & $\tts{1.655$\pm$ 0.028}$ & $\tts{3.474$\pm$ 0.037}$ & $\tts{1.966$\pm$0.018}$  \\ 
\textbf{Attnhp}& \ttt{11.223 $\pm$ 0.145} & {0.873$\pm$ 0.023} & \undl{0.550$\pm$ 0.014} & \tbf{4231.499$\pm$ 155.699} & \ttt{132.266$\pm$ 0.532} & $\tts{11.498$\pm$ 0.175}$ & \tbf{0.858$\pm$ 0.020} & \tbf{1.312$\pm$ 0.034} & \tbf{0.566$\pm$ 0.024}  \\ 
\textbf{NHP} & $\tts{11.973 $\pm$ 0.176}$ & $\tts{1.910 $\pm$ 0.031}$ & $\tts{0.712$\pm$ 0.017}$ & $\tts{5961.627$\pm$ 183.108}$ & $\tts{134.693$\pm$ 0.369}$ & \tbf{8.748$\pm$ 0.294} & $\tts{1.718$\pm$ 0.035}$ & $\tts{3.297$\pm$ 0.051}$ & $\tts{2.001$\pm$ 0.015}$  \\ 

\textbf{LogNM} & $\tts{\undl{11.052 $\pm$ 0.108}}$ & $\tts{1.941 $\pm$ 0.049}$ & $\ttt{0.601 $\pm$ 0.017}$ & $\ttt{5006.301 $\pm$ 287.390}$ & \undl{126.32 $\pm$ 0.591} & $\tts{10.395 $\pm$ 0.201}$ & $\tts{1.304 $\pm$ 0.040}$ & $\tts{1.932 $\pm$ 0.027}$ & $\ttt{0.994 $\pm$ 0.008}$\\

\textbf{TCDDM} & $\ttt{11.609$\pm$ 0.184}$ & $\tts{1.690 $\pm$ 0.023}$ & $\ttt{0.675 $\pm$ 0.009}$ & $\tts{5042.501 $\pm$ 324.55}$ & $\tts{129.009$\pm$ 0.923}$ & $\tts{11.203 $\pm$ 0.192}$ & $\tts{1.209 $\pm$ 0.068}$ & $\ttt{2.003 $\pm$ 0.033}$ & $\ttt{1.024 $\pm$ 0.020}$\\

\textbf{Homog. Poisson} & $\tts{13.510 $\pm$ 0.203}$ & $\tts{1.392 $\pm$ 0.034}$ & $\ttt{1.093 $\pm$ 0.047}$ & $\tts{5039.401 $\pm$ 442.580}$ & $\tts{143.105 $\pm$ 0.699}$ & $\ttt{9.300 $\pm$ 0.225}$ & $\tts{1.527 $\pm$ 0.079}$ & $\tts{3.342 $\pm$ 0.042}$ & $\tts{2.401 $\pm$ 0.0028}$\\

\midrule
\textbf{CDiff}    &  \tbf{10.147 $\pm$ 0.140} & \tbf{0.730 $\pm$ 0.019} & \tbf{0.519 $\pm$ 0.008 } & {4736.039 $\pm$ 114.586} & \tbf{124.339 $\pm$ 0.322} & \undl{9.122 $\pm$ 0.179} & \undl{0.861 $\pm$ 0.022} & {1.628$\pm$ 0.033} & {0.730 $\pm$ 0.013} \\ \bottomrule

\\
&\multicolumn{9}{c}{\textbf{Stackoverflow dataset}}\\ \midrule
&\multicolumn{5}{c|}{\textbf{$N=5$ events forecasting}}     &\multicolumn{4}{c}{\textbf{Interval forecasting $t'$ small}}\\ 
& $\textbf{OTD}$ & $\textbf{RMSE}_{e}$ &$\textbf{RMSE}_{x^+}$ & \textbf{MAPE} &\textbf{sMAPE} & $\textbf{OTD}$ & $\textbf{RMSE}_{e}$&$\textbf{RMSE}_{|\bs^+|}$ & $\textbf{MAE}_{|\bs^+|}$\\ \midrule
\textbf{HYPRO} & \undl{11.590$\pm$ 0.186}  & {0.586 $\pm$ 0.019} & 1.227$\pm$ 0.018 & $\ttt{1413.759$\pm$ 79.723}$ & 109.014$\pm$ 0.422 & \undl{9.677$\pm$ 0.117} & \tbf{0.530 $\pm$ 0.021} & $\tts{1.689 $\pm$ 0.017}$ & 1.007$\pm$ 0.030\\
\textbf{Dual-TPP} & $\tts{11.719 $\pm$ 0.109}$ & $\tts{0.591 $\pm$ 0.026}$ &  $\tts{1.296 $\pm$ 0.010} $&  \undl{1319.909 $\pm$ 121.366} &  106.697 $\pm$ 0.381 &  $\tts{9.963 $\pm$ 0.230}$ &  0.563 $\pm$ 0.023 &  1.572 $\pm$ 0.036 & \undl{0.987 $\pm$ 0.042}\\ 
\textbf{Attnhp} & {11.595 $\pm$ 0.197} & \undl{0.575 $\pm$ 0.009} & \undl{1.188$\pm$ 0.014 } & 1418.384$\pm$ 48.412 & \undl{105.799$\pm$ 0.516} & 9.787$\pm$ 0.321 & 0.552$\pm$ 0.018 & $\tts{\undl{1.559 $\pm$ 0.031}}$ & \tbf{0.963 $\pm$ 0.025}\\ 
\textbf{NHP} & $\tts{11.807$\pm$ 0.155}$ & $\tts{0.596 $\pm$ 0.015}$ & $\tts{1.261$\pm$0.013}$ & \tbf{1292.252$\pm$ 133.873} &  $\tts{108.074$\pm$ 0.661}$ &  $\tts{10.809$\pm$ 0.182}$ & $\tts{0.570$\pm$ 0.026}$ &  $\tts{1.716$\pm$ 0.037}$ &  $\tts{1.033$\pm$ 0.027}$\\  

\textbf{LogNM} & $\tts{13.124 $\pm$ 0.174}$ & $\tts{0.702 $\pm$ 0.008}$ & $\ttt{1.182 $\pm$ 0.039}$ & $\ttt{1335.23 $\pm$ 145.031}$ & $\ttt{108.409 $\pm$ 0.692}$ & $\tts{11.015 $\pm$ 0.191}$ & $\tts{0.629 $\pm$ 0.093}$ & $\tts{1.664 $\pm$ 0.042}$ & $\tts{1.032 $\pm$ 0.018}$\\

\textbf{TCDDM} & $\ttt{11.41 $\pm$ 0.129}$ & $\tts{0.630 $\pm$ 0.015}$ & $\ttt{1.201 $\pm$ 0.028}$ & $\ttt{1412.195 $\pm$ 135.312}$ & $\ttt{107.893 $\pm$ 0.942}$ & $\tts{10.23 $\pm$ 0.096}$ & $\tts{0.611 $\pm$ 0.024}$ & $\textbf{1.532 $\pm$ 0.06}$ & $\tts{1.021 $\pm$ 0.010}$\\

\textbf{Homog. Poisson} & $\tts{15.493 $\pm$ 0.144}$ & $\tts{0.693 $\pm$ 0.013}$ & $\tts{1.336 $\pm$ 0.059}$ & $\ttt{2034.235 $\pm$ 125.314}$ & $\tts{108.900 $\pm$ 0.074}$ & $\tts{13.12 $\pm$ 0.073}$ & $\tts{0.921 $\pm$ 0.045}$ & $\tts{ 1.886$\pm$ 0.008}$ & $\tts{1.120 $\pm$ 0.039}$\\

\midrule
\textbf{CDiff} & \tbf{10.735 $\pm$ 0.183} & \tbf{0.571 $\pm$ 0.012 }& \tbf{1.153 $\pm$ 0.011} & 1386.314 $\pm$ 57.750 & \tbf{100.586 $\pm$ 0.299} & \tbf{8.849  $\pm$ 0.187} & \undl{0.545 $\pm$ 0.015} & $\tts{1.564 $\pm$ 0.029}$ & 0.991$\pm$ 0.035\\ \bottomrule

\\
&\multicolumn{9}{c}{\textbf{Retweet dataset}}\\ \midrule
&\multicolumn{5}{c|}{\textbf{$N=5$ events forecasting}}     &\multicolumn{4}{c}{\textbf{Interval forecasting $t'$ small}}\\ 
& $\textbf{OTD}$ & $\textbf{RMSE}_{e}$ &$\textbf{RMSE}_{x^+}$ & \textbf{MAPE} &\textbf{sMAPE} & $\textbf{OTD}$ & $\textbf{RMSE}_{e}$&$\textbf{RMSE}_{|\bs^+|}$ & $\textbf{MAE}_{|\bs^+|}$\\ \midrule
\textbf{HYPRO} & 16.145$\pm$ 0.096& 1.105$\pm$ 0.026& \undl{27.236$\pm$ 0.259}& 22428.809 $\pm$ 780.393& \undl{103.052$\pm$ 1.206} & \tbf{13.199 $\pm$ 0.089} & 1.201$\pm$  0.053 & $\tts{1.602$\pm$ 0.096}$ & $\tts{1.103$\pm$ 0.075}$ \\
\textbf{Dual-TPP} & 16.050 $\pm$ 0.085& $\tts{1.077$\pm$ 0.027}$&  $\tts{31.493$\pm$0.162}$ &  \tbf{15403.772$\pm$ 831.413}&  \tbf{101.322$\pm$ 1.127} &  \undl{13.809$\pm$ 0.048} &  $\tts{1.197 $\pm$ 0.025}$ &  $\tts{1.478$\pm$0.082}$ &  $\tts{0.980 $\pm$ 0.038}$ \\ 
\textbf{Attnhp} & 16.124 $\pm$ 0.089& \undl{1.058 $\pm$ 0.029}&  29.247  $\pm$ 0.145& \undl{18377.481 $\pm$ 878.880}&  105.93  $\pm$ 1.380& $\tts{14.120  $\pm$ 0.127}$& \undl{1.144 $\pm$ 0.034} &  \undl{1.315 $\pm$ 0.070} &  \ttt{0.862  $\pm$ 0.051}\\ 
\textbf{NHP} & \ttt{15.945  $\pm$ 0.094}& $\tts{1.113  $\pm$ 0.040}$& $\tts{32.367  $\pm$ 0.104}$& $\tts{22611.646  $\pm$ 797.268}$ & $\tts{107.022 $\pm$ 1.077}$&  $\tts{14.201 $\pm$ 0.119}$ &  $\tts{1.161 $\pm$ 0.023}$ &  $\tts{1.369 $\pm$ 0.102}$&  $\tts{0.894 $\pm$ 0.025}$\\  

\textbf{LogNM} & $\ttt{16.043 $\pm$ 0.222}$ & $\tts{1.313 $\pm$ 0.011}$ & $\ttt{30.853 $\pm$ 0.119}$ & $\ttt{23084.93$\pm$ 784.430}$ & $\ttt{106.941 $\pm$ 2.031}$ & $\ttt{13.937 $\pm$ 0.239}$ & $\tts{1.208 $\pm$ 0.029}$ & $\ttt{1.590 $\pm$ 0.113}$ & $\ttt{0.874 $\pm$ 0.068}$\\

\textbf{TCDDM} & \undl{15.874 $\pm$ 0.053} & $\tts{1.194 $\pm$ 0.021}$ & $\ttt{28.530 $\pm$ 0.110}$ & $\ttt{19093.229 $\pm$ 880.932}$ & $\ttt{105.570 $\pm$ 0.94}$ & $\tts{14.771 $\pm$ 0.298}$ & $\tts{1.340 $\pm$ 0.030}$ & $\ttt{1.275 $\pm$ 0.084}$ & \undl{0.798 $\pm$ 0.028}\\

\textbf{Homog. Poisson} & $\tts{19.432 $\pm$ 0.033}$ & $\tts{1.405 $\pm$ 0.008}$ & $\tts{30.543 $\pm$ 0.083}$ & $\tts{28094.854 $\pm$ 684.501}$ & $\tts{108.591 $\pm$ 1.049}$ & $\tts{15.039 $\pm$ 0.591}$ & $\tts{1.347 $\pm$ 0.094}$ & $\tts{1.898 $\pm$ 0.020}$ & $\tts{1.091 $\pm$ 0.044}$\\

\midrule
\textbf{CDiff} & \tbf{15.858 $\pm$ 0.080} & \tbf{1.023$\pm$ 0.036} & \tbf{26.078$\pm$ 0.175}& 21778.765$\pm$ 689.206 & 106.62$\pm$ 1.008 & $\tts{14.073 $\pm$ 0.065}$ & \tbf{1.127 $\pm$ 0.029} & \tbf{1.123 $\pm$0.099} & \tbf{0.782 $\pm$ 0.063}\\ \bottomrule

\\

&\multicolumn{9}{c}{\textbf{Mooc dataset}}\\ \midrule
&\multicolumn{5}{c|}{\textbf{$N=5$ events forecasting}}     &\multicolumn{4}{c}{\textbf{Interval forecasting $t'$ small}}\\ 
& $\textbf{OTD}$ & $\textbf{RMSE}_{e}$ &$\textbf{RMSE}_{x^+}$ & \textbf{MAPE} &\textbf{sMAPE} & $\textbf{OTD}$ & $\textbf{RMSE}_{e}$&$\textbf{RMSE}_{|\bs^+|}$ & $\textbf{MAE}_{|\bs^+|}$\\ \midrule
\textbf{HYPRO} & $\ttt{11.718 $\pm$ 0.240}$ & \undl{0.811 $\pm$ 0.045} & \undl{0.308 $\pm$ 0.015} & $\ttt{12949.391 $\pm$ 441.590}$ & $\textbf{142.735 $\pm$ 17.901}$ & $\ttt{10.657 $\pm$ 0.092}$ & \undl{0.692 $\pm$ 0.018} & $\ttt{1.772 $\pm$ 0.034}$ & $\textbf{0.890 $\pm$ 0.056}$\\

\textbf{Dual-TPP} & $\tts{14.503 $\pm$ 0.334}$ & $\ttt{0.950 $\pm$ 0.027}$ & $\tts{0.412 $\pm$ 0.006}$ & $\ttt{14492.350 $\pm$ 294.754}$ & $\tts{146.100 $\pm$ 31.051}$ & $\tts{9.443 $\pm$ 0.130}$ & $\tts{0.845 $\pm$ 0.021}$ & $\tts{2.107 $\pm$ 0.045}$ & $\tts{1.335 $\pm$ 0.030}$\\

\textbf{AttNHP} & $\ttt{12.007 $\pm$ 0.214}$ & $\tts{0.854 $\pm$ 0.013}$ & $\textbf{0.297 $\pm$ 0.009}$ & $\textbf{11049.592 $\pm$ 509.283}$ & $\ttt{144.901 $\pm$ 24.093}$ & $\tts{10.201 $\pm$ 0.097}$ & $\ttt{0.775 $\pm$ 0.018}$ & $\ttt{1.891 $\pm$ 0.029}$ & $\ttt{1.084 $\pm$ 0.044}$\\

\textbf{NHP} & $\ttt{13.790 $\pm$ 0.327}$ & $\tts{0.983 $\pm$ 0.042}$ & $\ttt{0.394 $\pm$ 0.009}$ & $\ttt{14092.491 $\pm$ 301.340}$ & $\tts{143.534 $\pm$ 15.324}$ & $\tts{9.795 $\pm$ 0.204}$ & $\ttt{0.811 $\pm$ 0.021}$ & $\ttt{1.960 $\pm$ 0.047}$ & $\ttt{1.320 $\pm$ 0.084}$\\

\textbf{LogNM} & $\ttt{12.667 $\pm$ 0.255}$ & $\ttt{0.818 $\pm$ 0.034}$ & $\tts{0.407 $\pm$ 0.016}$ & $\ttt{15030.593 $\pm$ 503.492}$ & $\tts{143.010 $\pm$ 25.029}$ & \textbf{8.763 $\pm$ 0.113} & $\ttt{0.796 $\pm$ 0.033}$ & $\ttt{1.833 $\pm$ 0.042}$ & $\tts{1.230 $\pm$ 0.051}$\\

\textbf{TCDDM} & \undl{10.491 $\pm$ 0.134} & $\ttt{0.825 $\pm$ 0.019}$ & $\ttt{0.355 $\pm$ 0.007}$ & $\ttt{13059.245 $\pm$ 109.501}$ & \undl{142.941 $\pm$ 20.302} & $\ttt{9.506 $\pm$ 0.107}$ & $\ttt{0.753 $\pm$ 0.023}$ & $\textbf{1.634 $\pm$ 0.094}$ & $\ttt{1.046 $\pm$ 0.064}$\\

\textbf{Homog. Poisson} & $\tts{15.203 $\pm$ 0.075}$ & $\tts{1.007 $\pm$ 0.004}$ & $\tts{0.582 $\pm$ 0.009}$ & $\tts{18930.407 $\pm$ 404.338}$ & $\tts{175.587 $\pm$ 10.333}$ & \undl{9.104 $\pm$ 0.058} & $\tts{0.924$\pm$ 0.018}$ & $\tts{2.296 $\pm$ 0.106}$ & $\tts{1.203 $\pm$ 0.049}$\\
\midrule

\textbf{CDiff} & $\textbf{10.019 $\pm$ 0.429}$   & $\textbf{{0.792 $\pm$ 0.028}}$ & $\ttt{{0.310 $ \pm$ 0.014}}$ & \undl{11304.592 $\pm$ 100.049} & $\tts{144.551 $\pm$ 25.537}$ & $\ttt{9.259 $\pm$ 0.212}$ & $\textbf{0.686 $\pm$ 0.021}$ & \undl{1.733 $\pm$ 0.104} & \undl{0.923 $\pm$ 0.078}\\ 

\\

&\multicolumn{9}{c}{\textbf{Amazon dataset}}\\ \midrule
&\multicolumn{5}{c|}{\textbf{$N=5$ events forecasting}}     &\multicolumn{4}{c}{\textbf{Interval forecasting $t'$ small}}\\ 
& $\textbf{OTD}$ & $\textbf{RMSE}_{e}$ &$\textbf{RMSE}_{x^+}$ & \textbf{MAPE} &\textbf{sMAPE} & $\textbf{OTD}$ & $\textbf{RMSE}_{e}$&$\textbf{RMSE}_{|\bs^+|}$ & $\textbf{MAE}_{|\bs^+|}$\\ \midrule
\textbf{HYPRO} & $\ttt{9.552 $\pm$ 0.172}$ & $\ttt{1.397 $\pm$ 0.033}$ & $\ttt{0.433 $\pm$ 0.008}$ & \undl{1280.563 $\pm$ 45.347} & $\ttt{82.847 $\pm$ 0.748}$  & $\ttt{8.927 $\pm$ 0.052}$ & $\ttt{1.284 $\pm$ 0.010}$ & $\ttt{0.805 $\pm$ 0.004}$ & $\ttt{0.391 $\pm$ 0.008}$\\

\textbf{Dual-TPP} & $\tts{11.309 $\pm$ 0.093}$ & $\tts{1.742 $\pm$ 0.302}$ & $\tts{0.476 $\pm$ 0.010}$ & $\ttt{1420.118 $\pm$ 52.129}$ & $\tts{86.633 $\pm$ 0.573}$ & $\textbf{8.201 $\pm$ 0.490}$ & $\tts{1.408 $\pm$ 0.042}$ & $\tts{1.007 $\pm$ 0.011}$ & $\tts{0.517 $\pm$ 0.003}$\\

\textbf{AttNHP} & $\textbf{9.430 $\pm$ 0.131}$ & $\textbf{1.117 $\pm$ 0.049}$ & \undl{0.427 $\pm$ 0.033} & $\ttt{1335.591 $\pm$ 55.930}$ & $\ttt{83.121 $\pm$ 0.415}$ & $\tts{9.072 $\pm$ 0.059}$ & \undl{1.053 $\pm$ 0.041} & $\textbf{0.763 $\pm$ 0.015}$ & $\textbf{0.378 $\pm$ 0.007}$\\

\textbf{NHP} & $\tts{11.273 $\pm$ 0.198}$ & $\ttt{1.431 $\pm$ 0.024}$ & $\tts{0.501 $\pm$ 0.009}$ & $\ttt{1456.240 $\pm$ 35.557}$ & $\tts{90.591 $\pm$ 0.667}$ & $\tts{9.113 $\pm$ 0.135}$ & $\ttt{1.288 $\pm$ 0.018}$ & $\tts{0.978 $\pm$ 0.012}$ & $\tts{0.493 $\pm$ 0.012}$\\

\textbf{LNM} & $\tts{10.230 $\pm$ 0.224}$ & $\tts{1.663 $\pm$ 0.168}$ & $\ttt{0.447 $\pm$ 0.015}$ & $\tts{1447.203 $\pm$ 112.480}$ & $\ttt{88.900 $\pm$ 0.610}$ & $\ttt{9.042 $\pm$ 0.395}$ & $\tts{1.572 $\pm$ 0.031}$ & $\tts{ 0.874 $\pm$ 0.007}$ & $\ttt{0.487 $\pm$ 0.009}$ \\  

\textbf{TCDDM} & $\tts{10.557 $\pm$ 0.331}$ & $\ttt{1.409 $\pm$ 0.203}$ & $\ttt{0.460 $\pm$ 0.032}$ & $\ttt{1392.380 $\pm$ 84.213}$ & \undl{82.401 $\pm$ 0.810} & $\tts{10.003 $\pm$ 0.120}$ & $\ttt{1.338 $\pm$ 0.014}$ & $\ttt{0.793 $\pm$ 0.012}$ & $\ttt{0.420 $\pm$ 0.005}$\\

\textbf{Homog. Poisson} & $\tts{12.502 $\pm$ 0.155}$ & $\tts{2.130 $\pm$ 0.028}$ & $\tts{0.573 $\pm$ 0.007}$ & $\tts{1839.291 $\pm$ 54.200}$ & $\tts{105.831 $\pm$ 0.901}$ & \undl{8.923 $\pm$ 0.091} & $\tts{2.010 $\pm$ 0.014}$ & $\tts{1.042 $\pm$ 0.010}$ & $\tts{0.744 $\pm$ 0.011}$\\
\midrule

\textbf{CDiff}    & \undl{9.478  $\pm$  0.081}   & \undl{{1.326 $\pm$ 0.082}} & \textbf{{0.424 $ \pm$ 0.018}}& \textbf{1039.338 $\pm$ 43.030} &\textbf{81.287 $\pm$ 0.994} & $\tts{9.093 $\pm$ 0.049}$ & $\textbf{1.024 $\pm$ 0.016}$ & \undl{0.784 $\pm$ 0.009} & \undl{0.390 $\pm$ 0.007} \\

\bottomrule

\end{tabular}}

\label{tab:complete_table_n5}
\end{table*}

\end{document}